\documentclass[conf]{new-aiaa}
\usepackage[utf8]{inputenc}

\usepackage{varioref}% smart page, figure, table, and equation referencing
\usepackage{graphicx}
\usepackage{amsmath}
\usepackage[version=4]{mhchem}
\usepackage{siunitx}
\usepackage{longtable,tabularx}
\newcolumntype{P}[1]{>{\centering\arraybackslash}p{#1}}

\usepackage[nolist,printonlyused,nohyperlinks]{acronym}
\usepackage{todonotes}
\usepackage{booktabs}
\usepackage{comment}
\usepackage[format=hang]{caption}
\usepackage{subfigure}% subcaptions for subfigures
\usepackage{subfigmat}% matrices of similar subfigures, aka small mulitples
\usepackage{changepage} %margins for centered figures
\usepackage{fancyhdr}

\usepackage[ruled,vlined]{algorithm2e}

\title{Multi-GPU Approach for Training of Graph ML Models on large CFD Meshes}

\fancypagestyle{firstpage}
{
	\fancyhead[L]{}    
	\fancyhead[C]{\textcolor{red}{\large\textbf{AIAA SciTech Forum, January 23-27 2023}\\ \normalsize Reprinted by permission of the American Institute of Aeronautics and Astronautics, Inc..\\
		Original work available at \url{https://arc.aiaa.org/doi/10.2514/6.2023-1203}}}    
	\fancyhead[R]{}
}

\author{Sebastian Strönisch\footnote{Research Assistant, Center for Information Services and High Performance Computing (ZIH), sebastian.stroenisch@tu-dresden.de}}
\author{Maximilian Sander\footnote{Research Assistant, Center for Information Services and High Performance Computing (ZIH), maximilian.sander1@tu-dresden.de}}
\author{Andreas Knüpfer\footnote{Deputy Director, Center for Information Services and High Performance Computing (ZIH), andreas.knuepfer@tu-dresden.de}}
\affil{ZIH, Technische Universität Dresden, 01062 Dresden, Germany}
\author{Marcus Meyer\footnote{Specialist Aerothermal Methods, marcus.meyer@rolls-royce.com}}
\affil{Rolls-Royce Deutschland, 15827 Blankenfelde-Mahlow, Germany}

\begin{document}

\thispagestyle{firstpage}
\vspace*{-25pt}
\maketitle

% https://www.aiaa.org/SciTech/presentations-papers/technical-presenter-resources

\begin{abstract}
Mesh-based numerical solvers are an important part in many design tool chains.
However, accurate simulations like computational fluid dynamics are time and resource consuming which is why surrogate models are employed to speed-up the solution process.
Machine Learning based surrogate models on the other hand are fast in predicting approximate solutions but often lack accuracy.
Thus, the development of the predictor in a predictor-corrector approach is the focus here, where the surrogate model predicts a flow field and the numerical solver corrects it.
This paper scales a state-of-the-art surrogate model from the domain of graph-based machine learning to industry-relevant mesh sizes of a numerical flow simulation. 
The approach partitions and distributes the flow domain to multiple GPUs and provides halo exchange between these partitions during training. 
The utilized graph neural network operates directly on the numerical mesh and is able to preserve complex geometries as well as all other properties of the mesh.
The proposed surrogate model is evaluated with an application on a three dimensional turbomachinery setup and compared to a traditionally trained distributed model. 
The results show that the traditional approach produces superior predictions and outperforms the proposed surrogate model. %\todo[color=red]{der neue Ansatz kann zeitliche variabilität leider genauso wenig}
Possible explanations, improvements and future directions are outlined.
\end{abstract}

%\section{Nomenclature}
\begin{acronym}[URANS] %longest acronym
 	\acro{amp}[AMP]{Automatic Mixed Precision}
	\acro{cfd}[CFD]{Computational Fluid Dynamics}
	\acro{ddp}[DDP]{distributed data parallel}	
	\acro{dnn}[DNN]{Deep Neural Network}
	\acrodefplural{dnn}[DNNs]{Deep Neural Networks}
	\acro{gnn}[GNN]{Graph Neural Network}
	\acrodefplural{gnn}[GNNs]{Graph Neural Networks}
	\acro{gcn}[GCN]{graph convolutional network}
	\acro{gpu}[GPU]{Graphic Processing Unit}
	\acrodefplural{gpu}[GPUs]{Graphic Processing Units}
	\acro{ml}[ML]{Machine Learning}
	\acro{mlp}[MLP]{multi layer perceptron}
	\acro{mgn}[MGN]{\textsc{{M}esh{G}raph{N}ets}}
	\acro{dl}[DL]{Deep Learning}
	\acro{nn}[NN]{neural network}
	\acro{cnn}[CNN]{convolutional neural network}
	\acro{rans}[RANS]{Reynolds-Average Navier-Stokes}
	\acro{les}[LES]{Large Eddy Simulation}
	%\acro{pyg}[PyG]{PyTorch Geometric} %% NO PyG in torchDDP
	\acro{2d}[2D]{two-dimensional}
	\acro{3d}[3D]{three-dimensional}
	\acro{aoa}[AoA]{angle of attack}
	\acro{knn}[$k$-NN]{$k$-nearest neighbor}
	\acro{lstm}[LSTM]{Long Short-Term Memory}
	\acro{mse}[MSE]{mean square error}
	\acro{pinn}[PINN]{physics-informed neural network}
	\acro{oom}[OOM]{out-of-memory}
	\acro{rmse}[RMSE]{root mean square error}
	\acro{urans}[URANS]{unsteady Reynolds-Average Navier-Stokes}
	\acro{l}[$l$]{latent vector size}
	\acro{K}[$K$]{number of message passing steps}
	\acro{t}[$t$]{time step}
	
	\acrodefplural{t}[$t$]{time steps}
\end{acronym}
%printed nomenclature
%{\renewcommand\arraystretch{1.0}
%	\noindent\begin{longtable*}{@{}l @{\quad=\quad} l@{}}
%		$t$  & latent vector size \\
%		$N$ &    number of message passing steps \\
%		$t$& time step \\
%\end{longtable*}}
%\todo[inline,color=magenta]{Nomenclature \textbf{tbc}}

%% https://www.aiaa.org/publications/books/Publication-Policies#how-can-i-share-my-research-guidelines-for-authors

\section{Introduction}
The use of \ac{cfd} in many scientific disciplines for parameter studies, design optimization studies and uncertainty quantification is increasing significantly over the recent years. Since hundreds or even thousands of CFD calculations are needed for those types of studies even small improvements in solver convergence time can have a serious impact on the overall time of the design study, its computational resources and its costs. 
Data-driven approaches like \ac{ml} may help to address this issue: 
The main objective of this paper is a contribution towards the acceleration of the convergence process of \ac{cfd} simulations of industrially relevant size by applying a \ac{gnn} \cite{kipf2017semi} architecture to generate a tailored ML-based initial flow field prediction.

In recent years, graph-based \ac{ml} algorithms that preserve even complex geometries like airfoils were successfully applied for fluid flow field prediction \cite{Aulich2019,Belbute-Peres2020,thuerey2020deepFlowPred,Kashefi2021,pfaff2021learning,stroenischPASC22}. 
However, graph-based \ac{ml} approaches face the challenge of high \ac{gpu} memory consumption which limits the numerical mesh size of the CFD setup for which they can be utilized.
Therefore, scaling approaches like the one presented here that use multiple nodes with multiple \acp{gpu} need to be studied to enable the usage of graph-based \ac{ml} for relevant mesh sizes.
Recent advances in large graph \ac{ml} techniques mainly focus on graph embeddings \cite{hamilton2017inductive,pbg}.
To the best of the authors knowledge, there exists no multi-GPU training approach for regression problems like flow field prediction on large graphs using message-passing based model architectures.

The key contribution of this paper is the implementation of a graph-based multi-GPU learning approach and its application for the training and prediction of three-dimensional flow fields of CFD setups with up to one million cells while preserving the full numerical mesh. 
Furthermore, a comparison to a traditional distributed training approach reveals several limitations that need further investigation in the future.

Each flow domain is represented as a graph where the grid vertices are viewed as points and their connections as edges. 
This way graph convolution~\cite{kipf2017semi} can be utilized directly on the numerical mesh.
Flow values as well as coordinates are stored on each point as node features. 
The model can operate on graphs of arbitrary size but is constrained by the hardware that it is trained on \cite{stroenischPASC22}.
Here, up to 8 Nvidia A100~\cite{choquette_nvidia_2021} are used to train the model, whose main principles are outlined in chapter \ref{ch:model}.
The distributed learning approach is applied on the state-of-the-art \ac{mgn} from \cite{pfaff2021learning}.
Details on the proposed approach are given in the subsequent chapter \ref{ch:method}.

There exist two datasets on which the distributed learning approach is tested.
%The Navier-Stokes equations govern fluid flow and there exist a multitude of numerical methods to solve them for different spatial or temporal resolution of the flow, see \cite{Versteeg, Wendt} for further details.
The first dataset consists of the cylinderflow dataset from \cite{pfaff2021learning} and is used for validation of the approach. 
The results are outlined in section \ref{subsec:CYL}.
The second dataset is created from \ac{urans} simulations of geometry variations of a representative state-of-the-art turbine stator (see \cite{SchwarzeURANS2006} for more details on URANS).
The performance of the distributed \ac{mgn} will be evaluated according to its generalization capabilities on unseen geometry.
Results and discussions are given in section \ref{subsec:H01S}.

Finally, section \ref{subsec:perf} analyzes the performance aspect of the proposed approach and presents the communication overhead for multi-node training before deficiencies and future directions are discussed in section \ref{ch:conclusion}.

\section{Model}\label{ch:model}
The \ac{gnn} used in this work is trained to model the temporal progression of a flow field around an object.
Both datasets at hand consist of temporal trajectories of different geometries that are simulated with constant boundary conditions for a fixed number of \acp{t}.
The simulation mesh at time $t \in \mathcal{T}$ is denoted $\mathcal{M}^\text{t} = (\mathcal{N},\mathcal{E})$ with a set of nodes $\mathcal{N}$ that are connected by a set of edges $\mathcal{E}$.
Each node $i \in \mathcal{N}$ has an associated set of coordinates $\mathbf{x}_\text{i}$ and dynamical quantities $\mathbf{q}_\text{i,t}$ that are to be modeled.
Summarized, the learning task is to model the evolution of time-dependent dynamical quantities over a fixed mesh $\mathcal{M} = \mathcal{M}^\text{t}, t\in \mathcal{T}$ and sample $\mathbf{q}_\text{i,t}$ at the mesh nodes. 

The \ac{mgn} model that is used here was developed by \cite{pfaff2021learning} and can be utilized for all kinds of graph-based prediction tasks.
Its core architecture consists of an encoder, a series of processing blocks and a decoder. 
The simulation mesh and the node features of a time step $t$ are encoded into a graph $\mathcal{G}_\text{t} = (\mathcal{N}_\text{t},\mathcal{E})$. 
Analogous to \cite{pfaff2021learning} positional features $\mathbf{x}_\text{i}$ are provided as relative edge features to achieve spatial equivariance. 
Edge features $\mathtt{e}_\text{ij} \in \mathcal{E}$ of an edge between nodes $i$ and $j$ hold the relative displacement vector $\mathbf{x}_\text{ij} = \mathbf{x}_\text{i} - \mathbf{x}_\text{j}$ as well as its norm $|\mathbf{x}_\text{ij}|$.
The input node features $\mathbf{f}_\text{i,t} \in \mathcal{N}_\text{t}$ are concatenated from dynamical quantities $\mathbf{q}_\text{i,t}$ and a one-hot vector $\mathbf{n}_\text{i}$ indicating the node type, such as e.g. inflow, outflow or wall.
Node features $\mathbf{f}_\text{i,t}$ and Edge Features $\mathtt{e}_\text{ij}$ are encoded separately into a $l$-sized latent vector by the encoder block. \acused{l}
The encoded node and edge features are then passed to the processor which consists of $K$ processing blocks. 
A processing block is also called message passing block as features are exchanged between neighboring nodes.
Each block is modeled as an independent \ac{mlp} which is applied to the output of the previous block. 
After all \ac{K} are executed, a decoder transforms the node features in latent space to the desired output features $\mathbf{p}_\text{i}$.

A message-passing-like scheme called neighborhood aggregation is used in the processor block so that each node gathers and scatters node features from itself to its connected neighbors in the graph.
This can be viewed as a generalization of the convolutional operator to irregular domains.
For \ac{cfd} data this scheme is executed multiple times so that information flow across a large part of the domain is ensured.
In \cite{pfaff2021learning} prediction capabilities are demonstrated on simulation data from a flag, a deforming plate and two-dimensional flows around objects.
The \ac{cfd} datasets used by those authors consisted of samples with mesh sizes of around five thousand cells which were easily trained on a single \ac{gpu}. 

The learning task regarding CFD that is addressed with this architecture can be described by
\begin{align}
	f\left( \mathcal{E}, \mathcal{N}_t \right) \rightarrow \Delta \mathbf{q}_\text{t} 
\end{align}
so that the predicted flow field of the next time step $\mathbf{\tilde{q}}_\text{t+1} = f\left( \mathcal{E}, \mathcal{N}_t \right) + \mathbf{q}_\text{t} = \mathbf{p}_\text{i,t} + \mathbf{q}_\text{t}$.
The mesh is constant for a given trajectory so that $\mathcal{G}_\text{t+1} = (\mathcal{N}_\text{t+1},\mathcal{E})$. 

In this work, a custom PyTorch~\cite{pytorch2019} \ac{mgn} version is deployed on three-dimensional samples with roughly one million cells using the distribution approach presented in section \ref{ch:method}.

\section{Methodology} \label{ch:method}

Two approaches are considered and compared in the present paper for application of a \ac{dnn} to industry-relevant mesh sizes. 
The first and trivial one is training a neural network (NN) on a small part of the flow domain and combining predictions of all parts of the domain for the full flow field. 
The downside of this approach is the constraint information flow which is disrupted between the different parts. For evaluating a complete flow field, multiple predictions have to be stitched together. 
On the other side, already established scaling approaches like the one by Horovod \cite{sergeev2018horovod} can be used to increase the number of samples per training batch through utilizing multiple \ac{gpu}s. 

The second approach which is developed here combines classic high performance computing strategies with the machine learning approach at hand. 
The flow domain is partitioned using a current version of Metis \cite{metis97} and then distributed on multiple GPUs that train separate models in parallel. 
The parameters of \ac{ml} optimizer and model are kept separate but use synchronized gradients utilizing the \ac{ddp} framework of PyTorch \cite{pytorch2019,DDPDesignNote}.
This way the optimizers are expected to be synchronized as well.
However, this approach keeps separate copies of model and optimizer parameter on each \ac{gpu} which increases the per-GPU memory usage significantly \cite{deepspeedZERO20}.

\begin{figure}[t!]
	\begin{subfigmatrix}{2}
		\subfigure[\label{f:partitioned}]{\includegraphics[width=10.2cm,keepaspectratio]{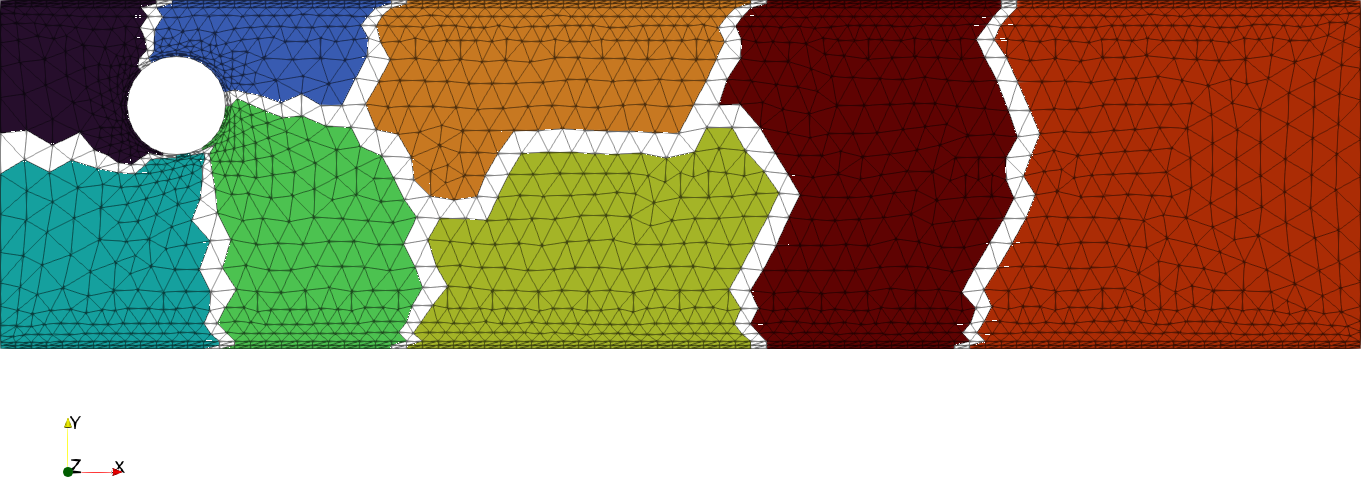}}
		\hfill
		\subfigure[\label{f:halosGreen}]{\includegraphics[width=5cm,keepaspectratio]{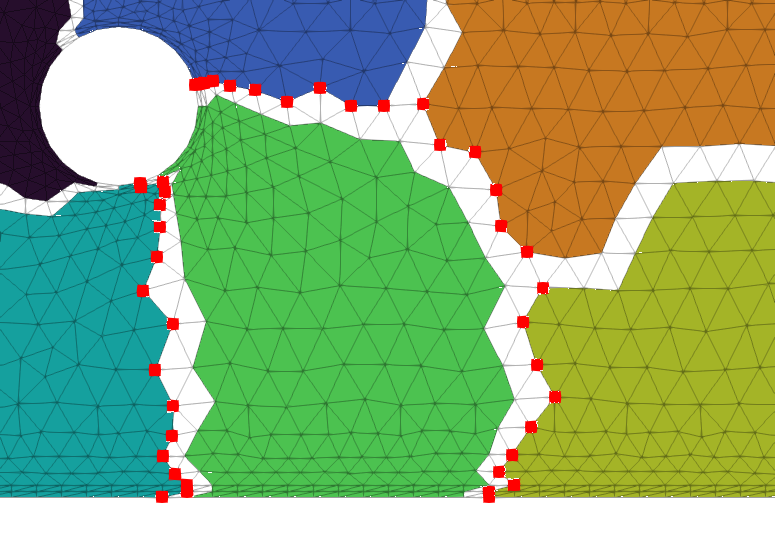}}   
		\hfill
	\end{subfigmatrix}
	\caption{(a) Partitioned cylinderflow sample from \cite{pfaff2021learning} and (b) halo (red dots) identification of bright green partition}
\end{figure}
In order to ensure information flow across the whole domain, each partition additionally shares all nodes that are directly connected to other partitions. 
These cells are typically called halo cells \cite{Halos}.
In this work the halo cell layer is of size one.
For demonstration purposes a partitioned sample from the cylinder dataset from \cite{pfaff2021learning} is shown in figure~\vref{f:partitioned}.
Furthermore the halo layer for one arbitrary partition (here bright green) is marked by red dots in figure~\ref{f:halosGreen}.

\begin{figure}[b!]
	\centering
	\includegraphics[width=16cm]{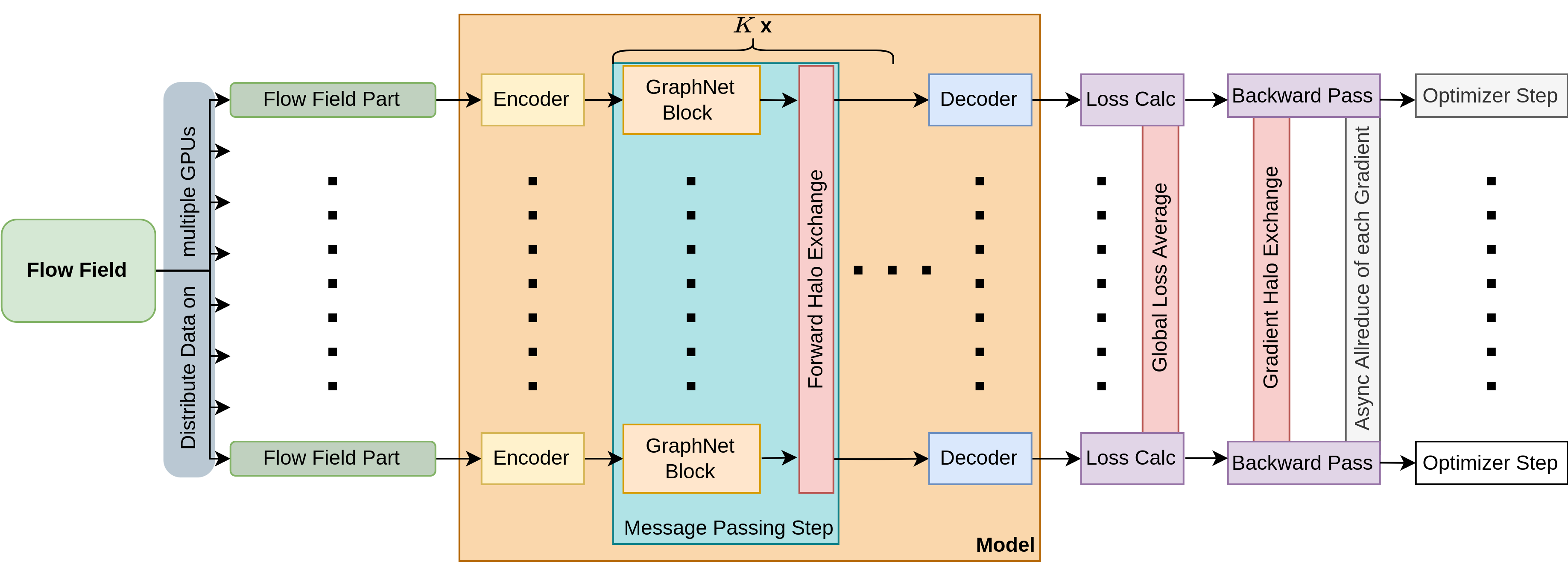}
	\caption{Multi GPU \textsc{{M}esh{G}raph{N}ets} (MGN) model architecture with $K$ message-passing steps derived from \cite{pfaff2021learning}}
	\label{fig:MGN_multiGPU}
\end{figure}

Figure \ref{fig:MGN_multiGPU} visualizes one optimizer step with a single backward step including the communication between all parallel model replicas.
Distributed model training already uses communication for gradient synchronization over all parameters with an asynchronous \textit{allreduce} after the loss backward-pass.
In addition, a global mean squared error loss calculation is used as well as a forward halo exchange and a backward gradient exchange between halos is introduced.
The additional communication is colored red in Fig. \ref{fig:MGN_multiGPU}.
Information exchange between halos happens between adjacent message passing steps.

This counter-intuitive approach forces synchronization between GPUs for each sample and each message passing step. 
The time that each device idles before all \acp{gpu} resume is called imbalance and section \ref{subsec:perf} visualizes it's impact on overall runtime.
Against expectations the communication overhead for a distributed two-dimensional sample on eight Nvidia A100 was only around 18\% compared to the single GPU application of the same dataset. 
This comparison is only possible in small cases, where a sample can fit in a single \ac{gpu} memory. 
The assumption is that the latency of the communication dominates for small samples, but for large graphs the computational part should predominate significantly.

For a detailed performance analysis Score-P~\cite{Knuepfer2012} in conjunction with Vampir~\cite{vampir} is used. 
Score-P is a tool used to record event traces and originates in the domain of High Performance Computing. 
It supports but is not limited to C/C++, Fortran, CUDA and MPI.
Recently, a detailed analysis of Python code with the focus on low overhead was made available~\cite{Gocht2021}.

\section{Results}
In this chapter the proposed distributed training method is first applied on a two-dimensional cylinderflow dataset. This pre-study is used to validate the approach and compare the accuracy to the single \ac{gpu} implementation.
Afterwards, a turbomachine flow field is used to test the method on meshes with relevant mesh sizes.
The proposed method is compared to a traditional training approach and the capabilities of the superior model are further examined.
Finally, an extensive performance study is presented, which analyses the introduced communication overhead as well as the device imbalance due to the synchronized halo exchange.
\subsection{Validation on Cylinderflow}\label{subsec:CYL}
The cylinderflow dataset from \cite{pfaff2021learning} is used to validate the proposed distributed learning approach.
Hence, a \ac{mgn} is trained on a single \ac{gpu} and a second model is trained using the proposed approach.
The dataset is compiled from geometric variation of a cylinder in \ac{2d} incompressible crossflow. 
Variations include diameter of the cylinder and position in the fluid domain. 
The dataset consists of 1000 training, 100 test and 100 validation trajectory of which each features 600 frames of a developing flow around the cylinder.
As the cylinder dataset consists of samples with only 1885 points in average, the full \ac{mgn} with 15 message passing steps and a latent vector size of 128 could be trained.

The input node features $\mathbf{f}_\text{i,t}$ are compiled from velocity in $x$ direction $u$, velocity in $y$ direction $v$ and the one-hot node type vector.
Output node features $\mathbf{p}_\text{i,t}$ are $\Delta u$ and $\Delta v$ as well as pressure $p$ of the next timestep. 
This choice of target features was made according to \cite{pfaff2021learning}.

In a first attempt to reproduce the results, the demo code that is provided by \cite{pfaff2021learning} and implemented in TensorFlow~\cite{TensorFlow} was adapted to fit the learning task of the cylinderflow. 
In the following, this version is denoted \textsc{MGN-demo}.
Table~\ref{tab:CompCYL} shows that its \ac{rmse} on the validation set is very close to the original work.
Furthermore, it even outperformed the results published in the paper for all extrapolation step widths.
Here, and throughout this paper the \ac{rmse} is used to be comparable to the original work \cite{pfaff2021learning}. A more sophisticated and \ac{cfd} specific error measure is part of future improvements.
In this work, the \ac{ml} framework of choice is PyTorch \cite{pytorch2019}, which is why a torch version of \ac{mgn} was implemented. 
The results are slightly worse (cf. \textsc{MGN-torch} in Table \ref{tab:CompCYL}) than the original paper but comparable. 
Other than the obvious difference in the framework that was used, a root cause investigation did not gave any hints on why the results are different.

\begin{table}[b!]
	\begin{center}
		\caption{Comparison of mean and standard error of \ac{rmse} over the cylinderflow validation set}
		\label{tab:CompCYL} 
		\begin{tabularx}{1\textwidth}{lcP{1.5cm}P{1.5cm}P{1.5cm}P{1.5cm}P{1.7cm}P{1.7cm}}
			\toprule
			\textbf{Model} &\textbf{\#~devices} &\textbf{\#~samples per batch per device}& \textbf{\#~batches presented} $\times 10^{6}$ & \textbf{\#~steps optimizer} $\times 10^{6}$ & \textbf{RMSE 1-step} $\times 10^{-3}$ & \textbf{RMSE 50-step} $\times 10^{-3}$ & \textbf{RMSE 600-steps} $\times 10^{-3}$\\
			\midrule
			{\protect\NoHyper\citeauthor{pfaff2021learning}\protect\endNoHyper} \cite{pfaff2021learning} &1&2&$10$&$10$&$2.34\pm0.12$&$6.30\pm0.70$&$40.88\pm7.20$\\
			\textsc{MGN-demo}                    &1&2&$10$&$10$&$1.90\pm0.06$&$5.69\pm0.35$& $36.65\pm2.79$\\
			\textsc{MGN-torch}                   &1&2&$10$&$10$&$3.40\pm0.12$&$11.61\pm0.66$&$40.02\pm3.38$\\
			\textsc{MGN-gradAcc}			   &1&1&$20$&$10$&$4.39\pm0.14$&$11.83\pm0.58$&$36.51\pm3.21$\\
 			\textsc{MGN-Halo} 		               &8&1&$20$&$10$&$4.23\pm0.11$&$15.30\pm0.75$&$58.95\pm3.58$\\
		\end{tabularx}	
	    
	\end{center}
	\renewcommand*\footnoterule{}

\end{table}
Now, another batching strategy called gradient accumulation is employed because large numerical meshes with up to and over $10^6$ points already exhaust device memory when partitioned onto eight \acp{gpu}.
So only a single sample per batch can be processed and the gradients are accumulated over two consecutive samples.
In order to have a fair comparison and explore the impacts of this batching strategy on the problem at hand, a single \ac{gpu} version with gradient accumulation called \textsc{MGN-gradAcc} was trained additionally.
Surprisingly, it showed a significant deterioration of the \ac{rmse} on the validation set.

Finally, a model called \textsc{MGN-Halo} employing the proposed learning approach was trained on the cylinder dataset using eight \acp{gpu}. 
The \ac{rmse} for the single step is slightly worse and for the 50-step rollout slightly better than for the comparable single GPU version \textsc{MGN-gradAcc}.
The overall validation of the approach is therefor regarded successful, even though the \ac{rmse} for the 600-step rollout deviates significantly (cf. Table \ref{tab:CompCYL})).

\begin{figure}[t!]
	\newcommand\x{0.325} %	0.33
	\newcommand\cropb{13} %
	% trim={<left> <lower> <right> <upper>}
	\begin{subfigmatrix}{3}
		
		\subfigure{\includegraphics[trim={0 \cropb cm 0 1cm},clip,width=\x\linewidth] {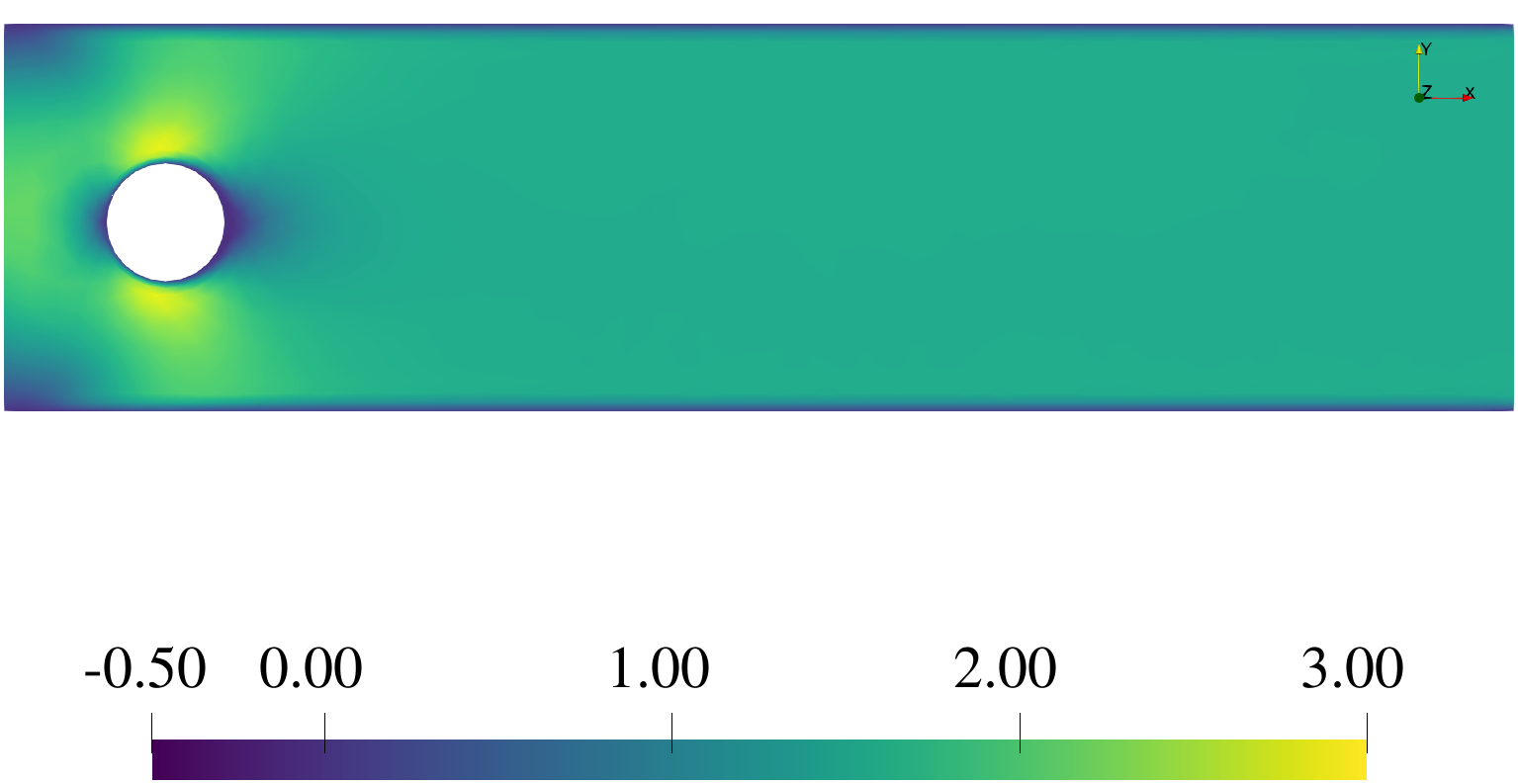}}
		\put(-0.345\linewidth,3){\rotatebox{90}{1-step}}   
		\hfill
		\subfigure{\includegraphics[trim={0 \cropb cm 0 1cm},clip,width=\x\linewidth] {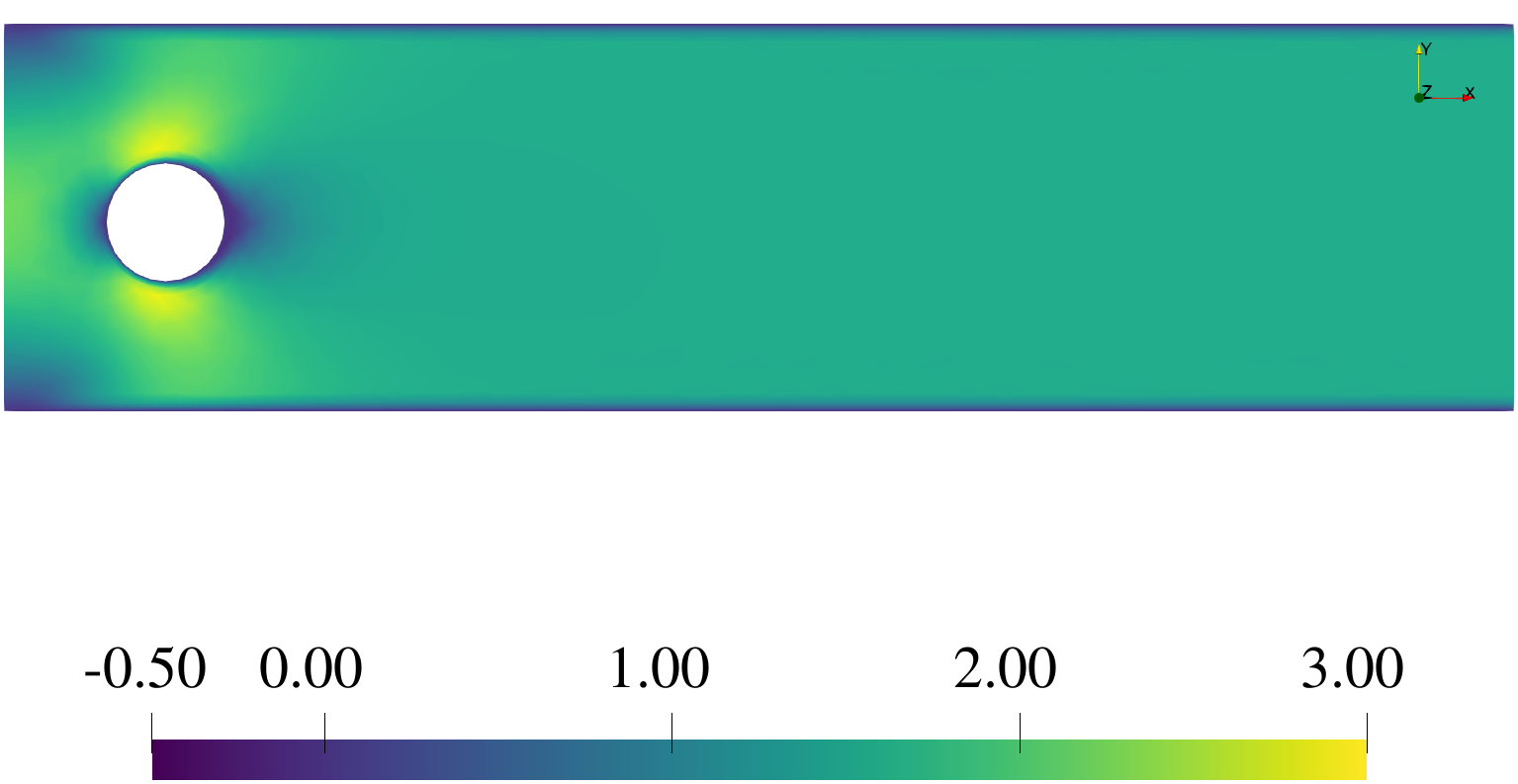}}   
		\hfill
		\subfigure{\includegraphics[trim={0 \cropb cm 0 1cm},clip,width=\x\linewidth] {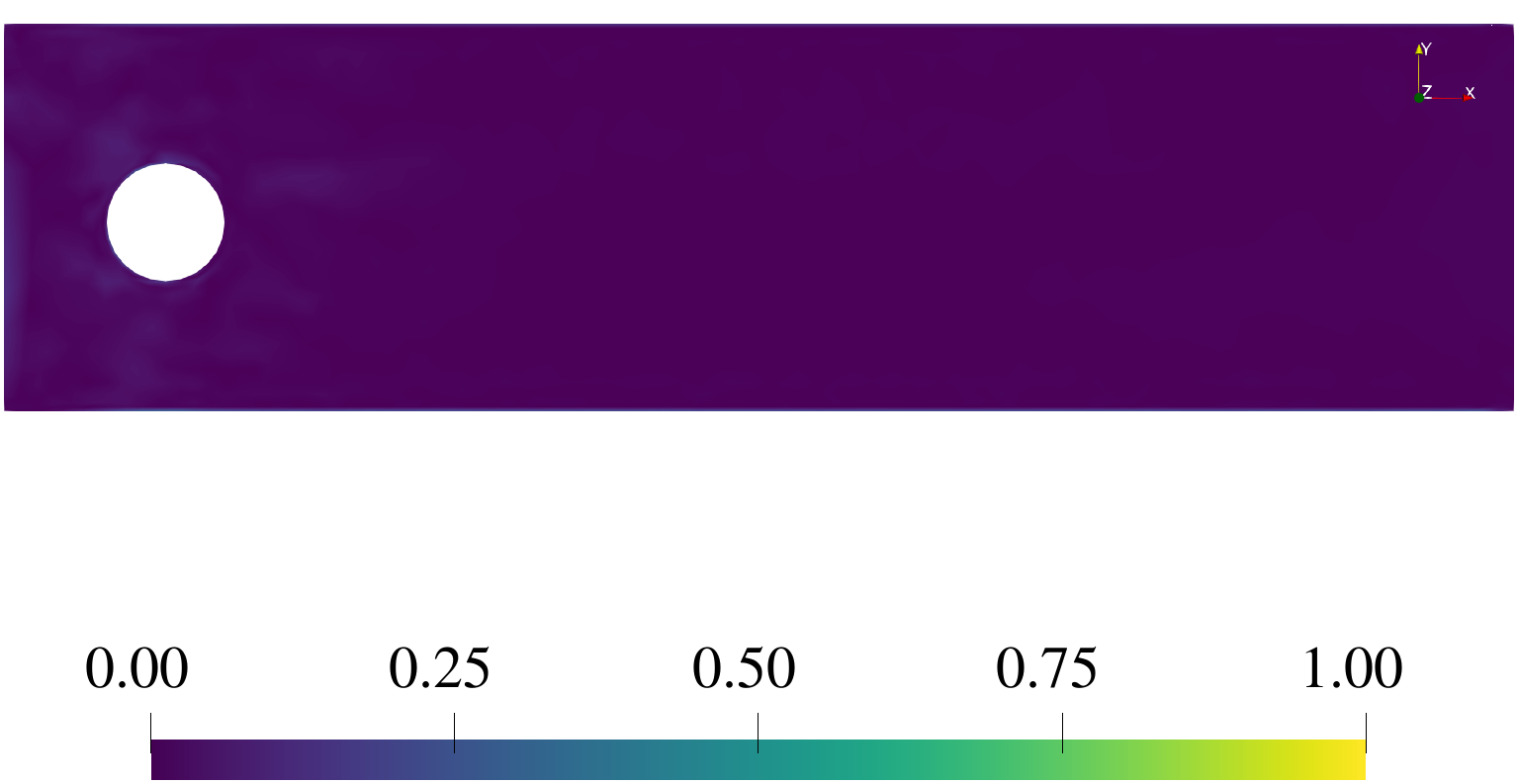}}   
		%% New Line
		\subfigure{\includegraphics[trim={0 \cropb cm 0 1cm},clip,width=\x\linewidth]{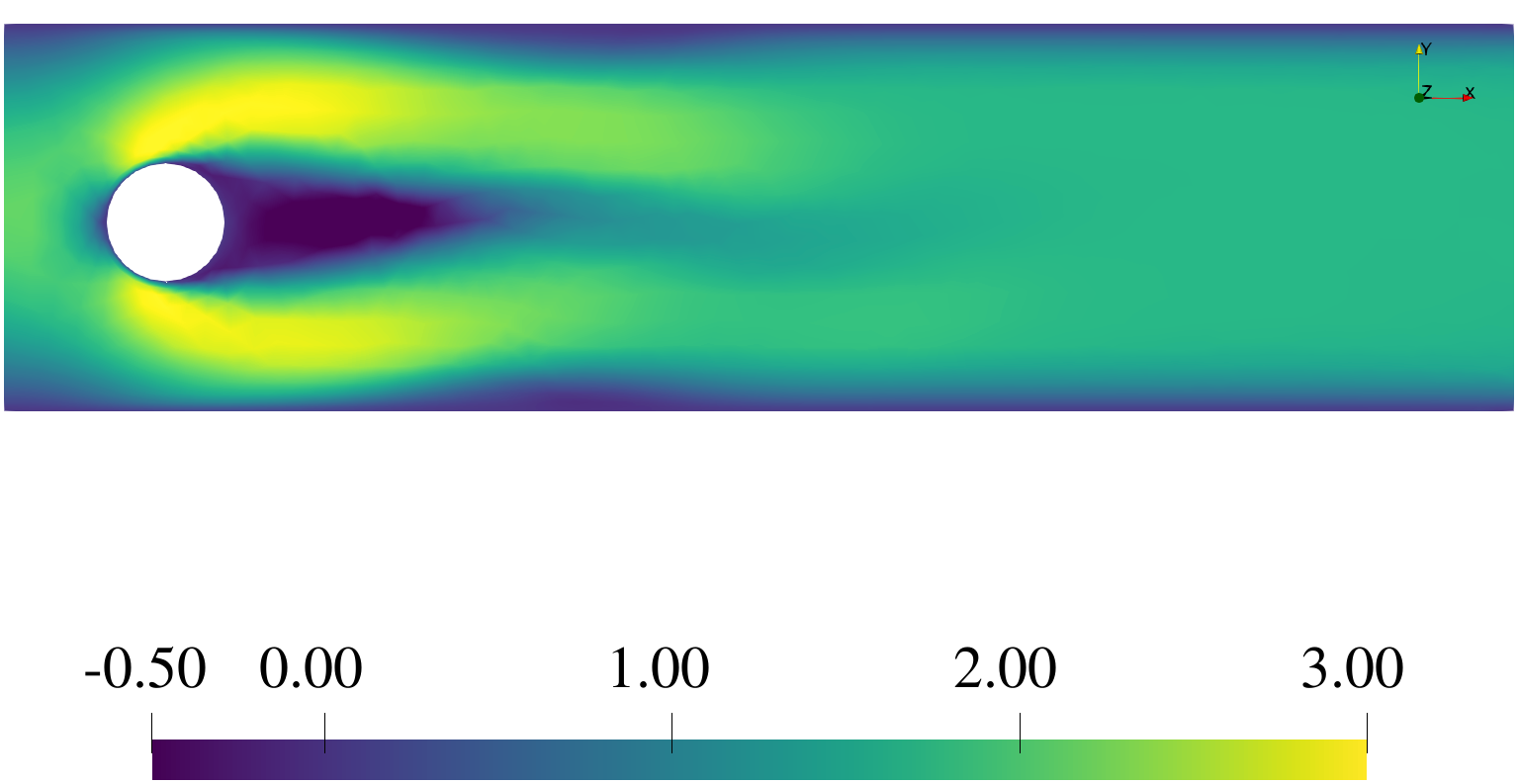}} 
		\put(-0.345\linewidth,3){\rotatebox{90}{50-step}} 
		\hfill
		\subfigure{\includegraphics[trim={0 \cropb cm 0 1cm},clip,width=\x\linewidth]{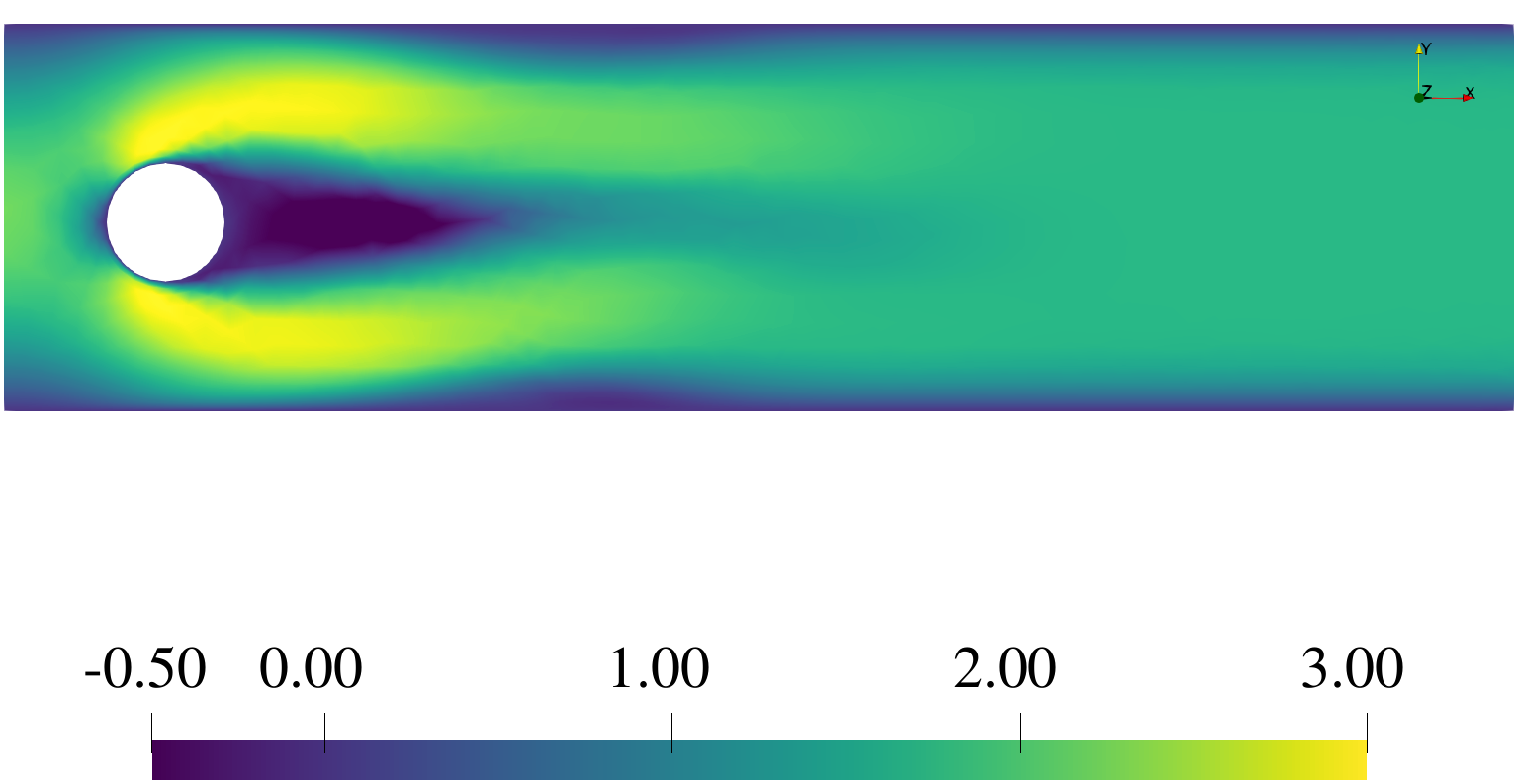}} 
		\hfill
		\subfigure{\includegraphics[trim={0 \cropb cm 0 1cm},clip,width=\x\linewidth] {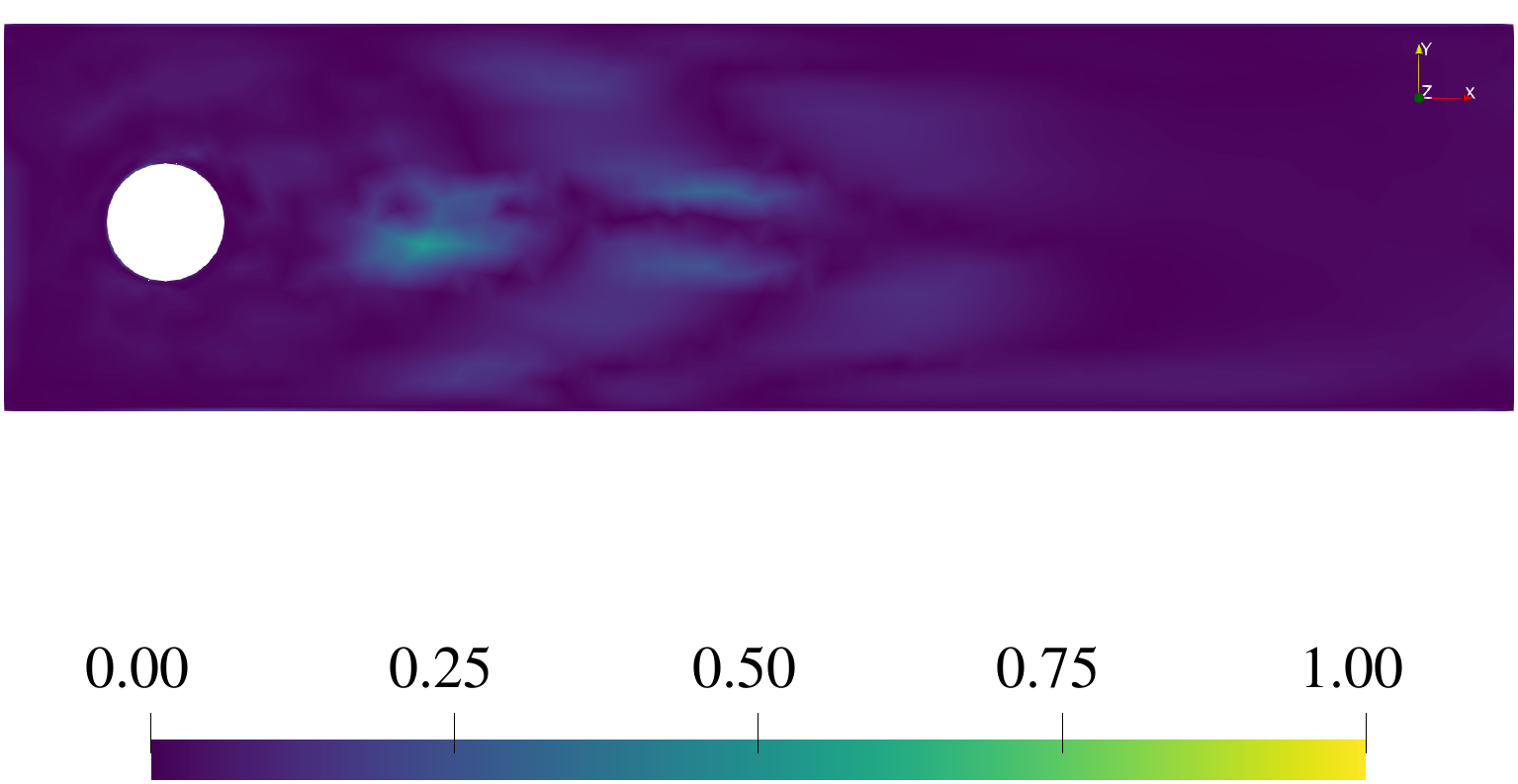}}   
		%% New Line
		\subfigure{\includegraphics[trim={0 \cropb cm 0 1cm},clip,width=\x\linewidth]{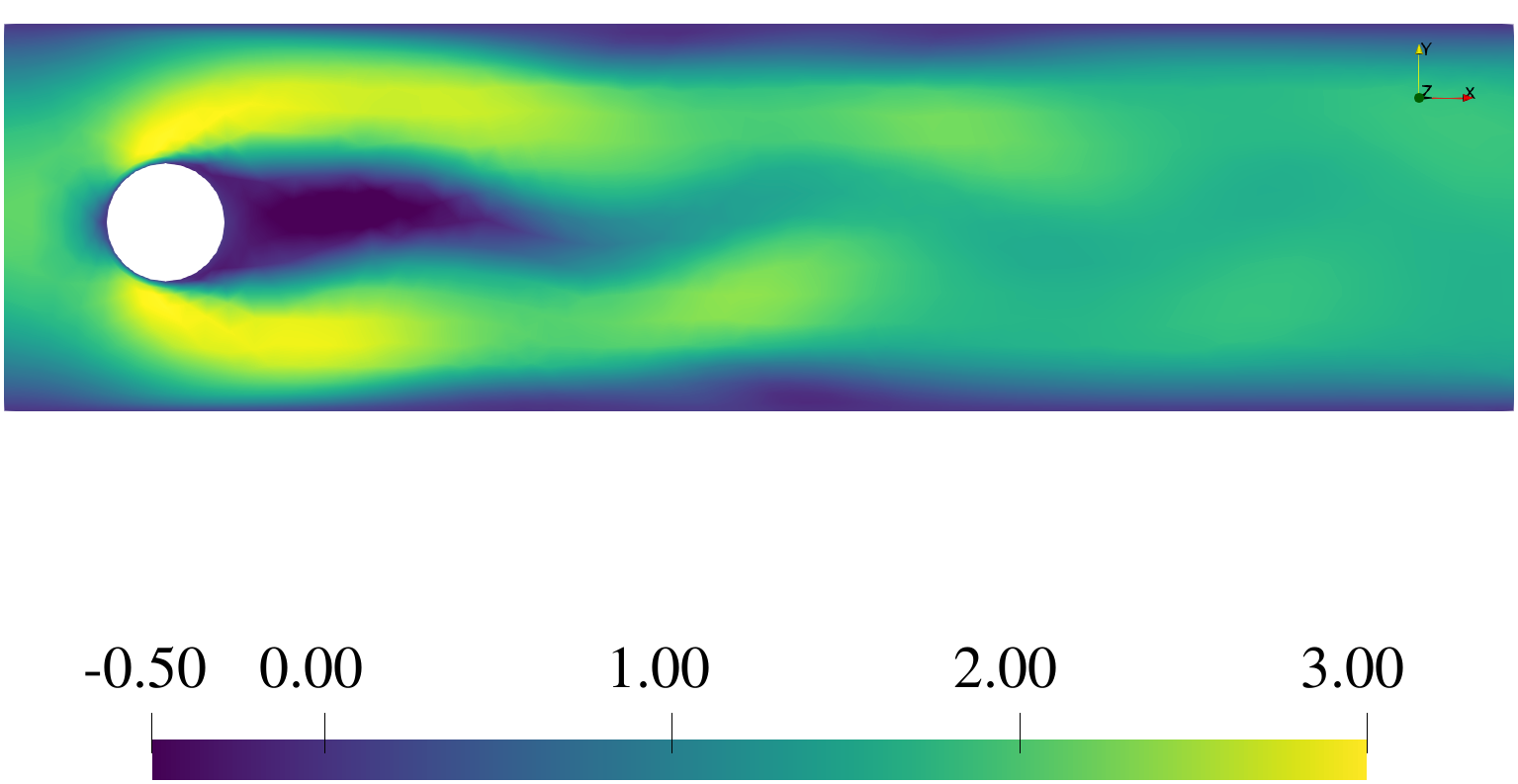}} 
		\put(-0.345\linewidth,3){\rotatebox{90}{100-step}} 
		\hfill
		\subfigure{\includegraphics[trim={0 \cropb cm 0 1cm},clip,width=\x\linewidth]{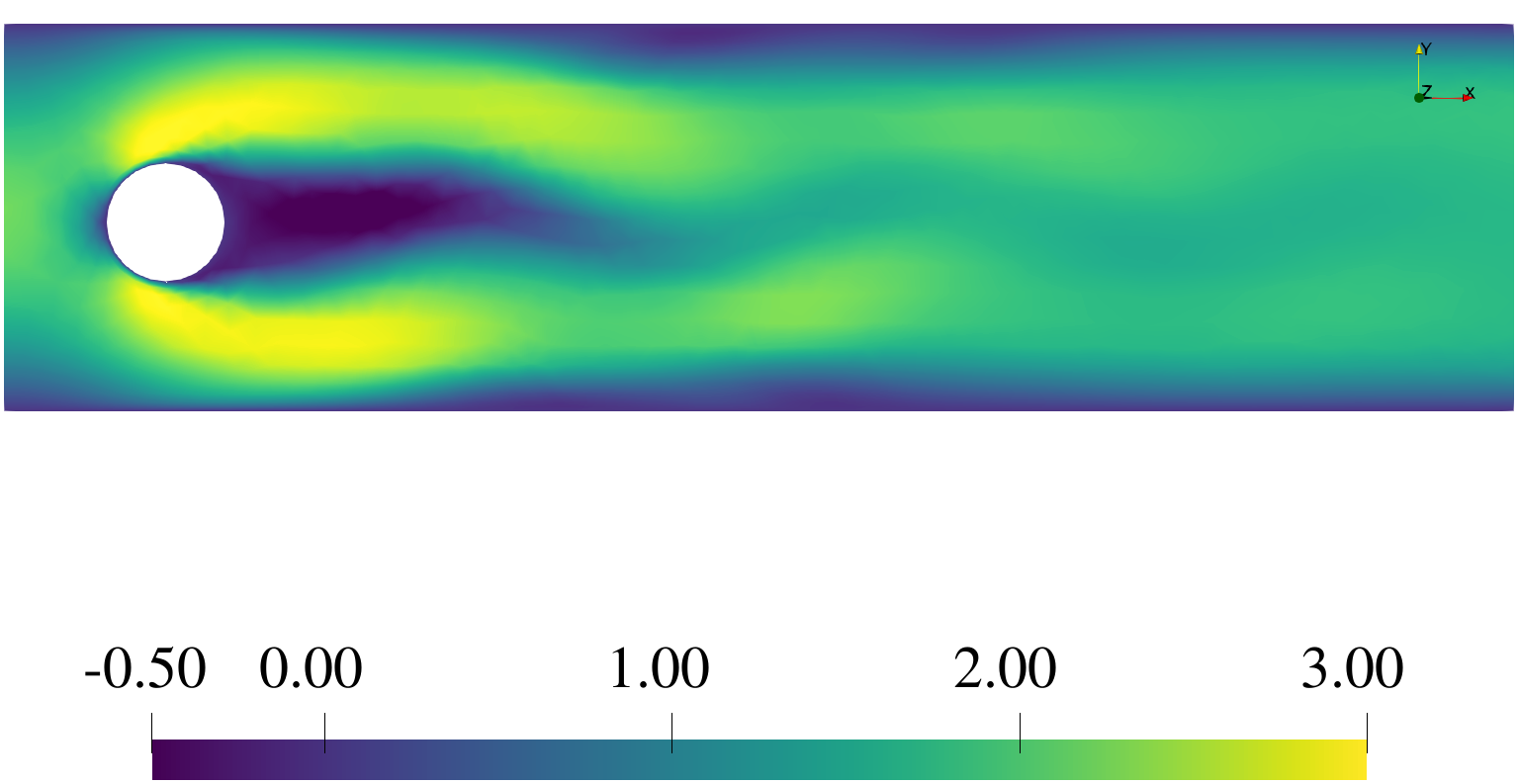}} 
		\hfill
		\subfigure{\includegraphics[trim={0 \cropb cm 0 1cm},clip,width=\x\linewidth]{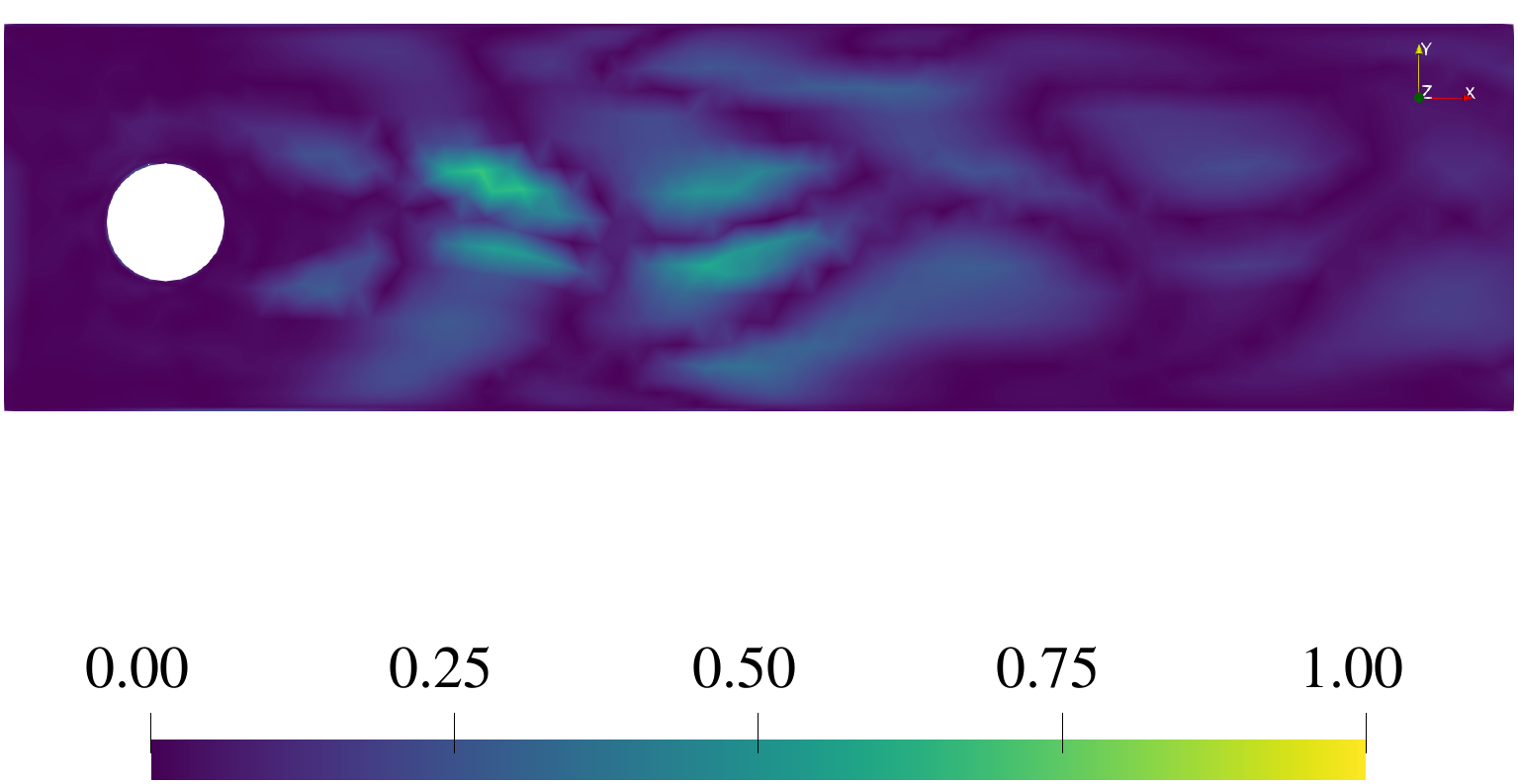}} 
		%% New Line
		\subfigure{\includegraphics[trim={0 \cropb cm 0 1cm},clip,width=\x\linewidth]{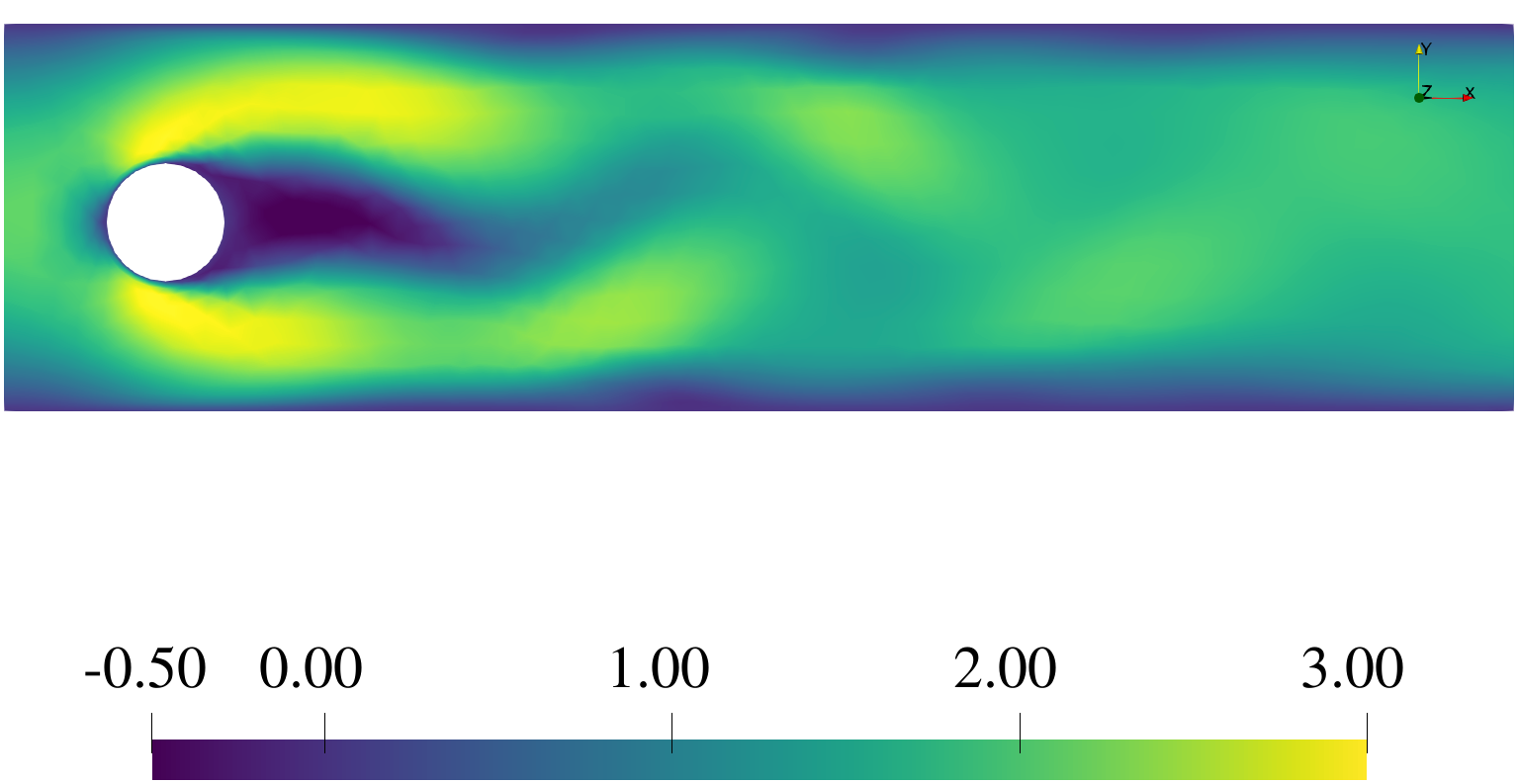}} 
		\put(-0.345\linewidth,3){\rotatebox{90}{150-step}} 
		\hfill
		\subfigure{\includegraphics[trim={0 \cropb cm 0 1cm},clip,width=\x\linewidth]{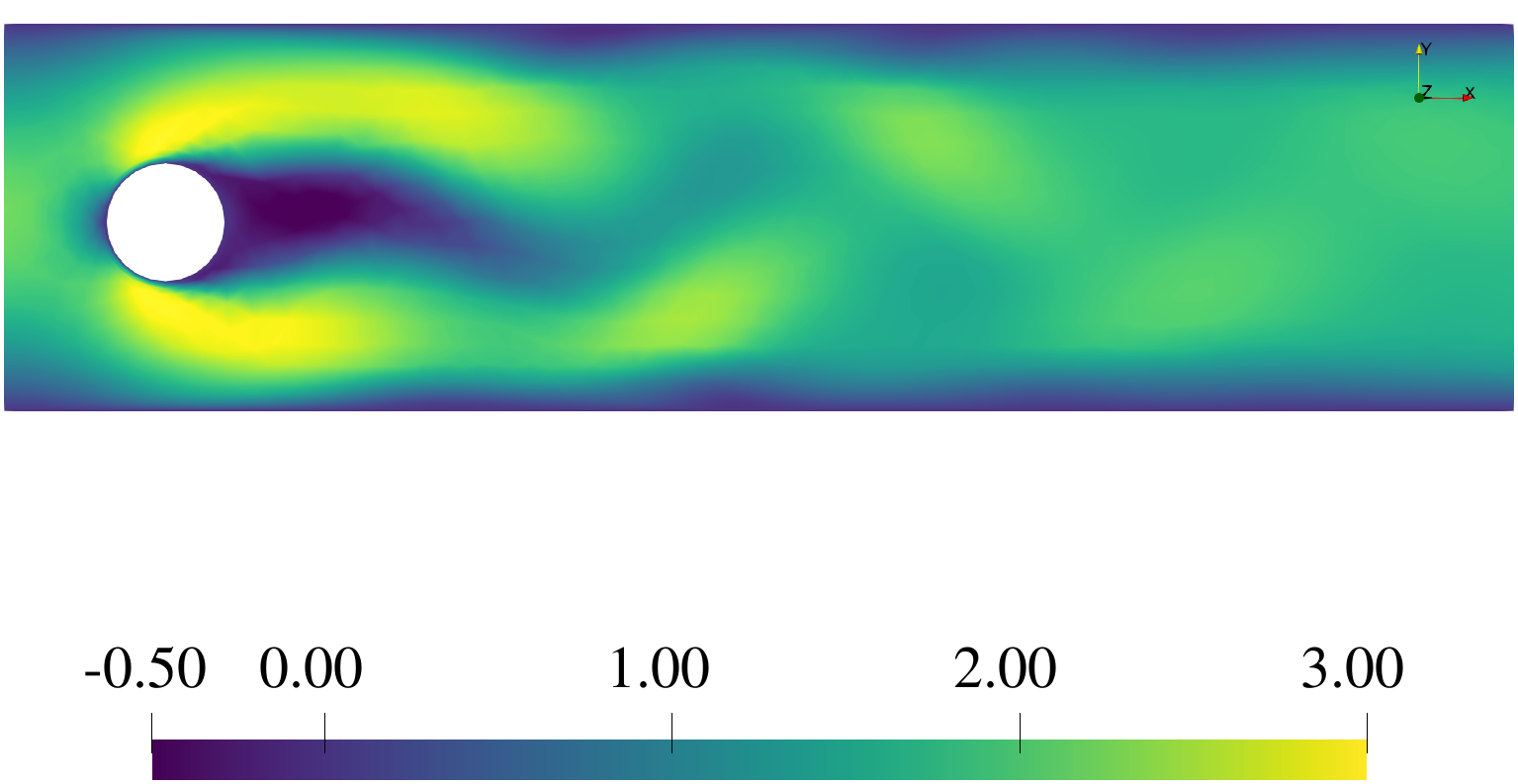}} 
		\hfill
		\subfigure{\includegraphics[trim={0 \cropb cm 0 1cm},clip,width=\x\linewidth]{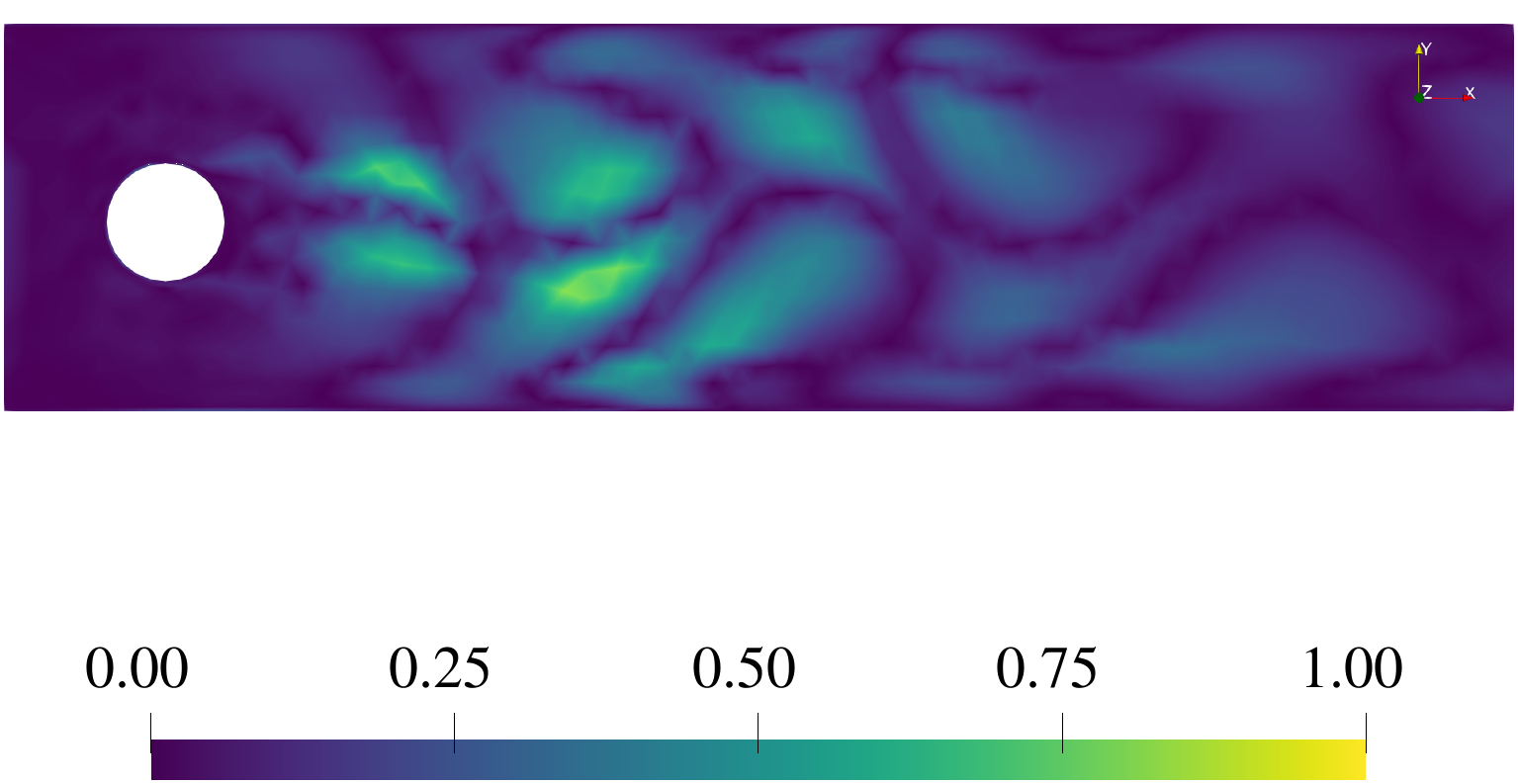}}
		%% New Line
		\subfigure{\includegraphics[trim={0 \cropb cm 0 1cm},clip,width=\x\linewidth]{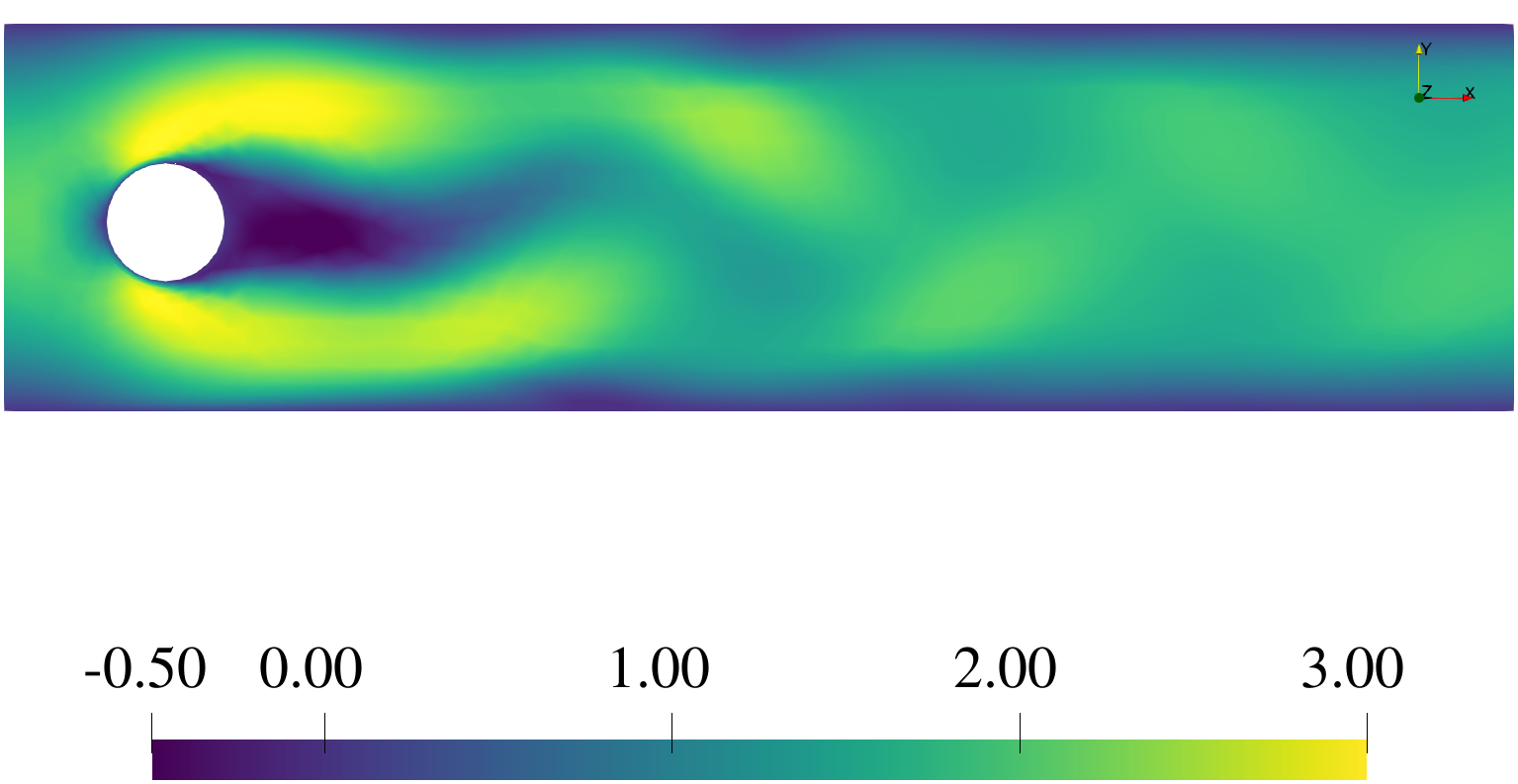}} 
		\put(-0.345\linewidth,3){\rotatebox{90}{200-step}} 
		\hfill
		\subfigure{\includegraphics[trim={0 \cropb cm 0 1cm},clip,width=\x\linewidth]{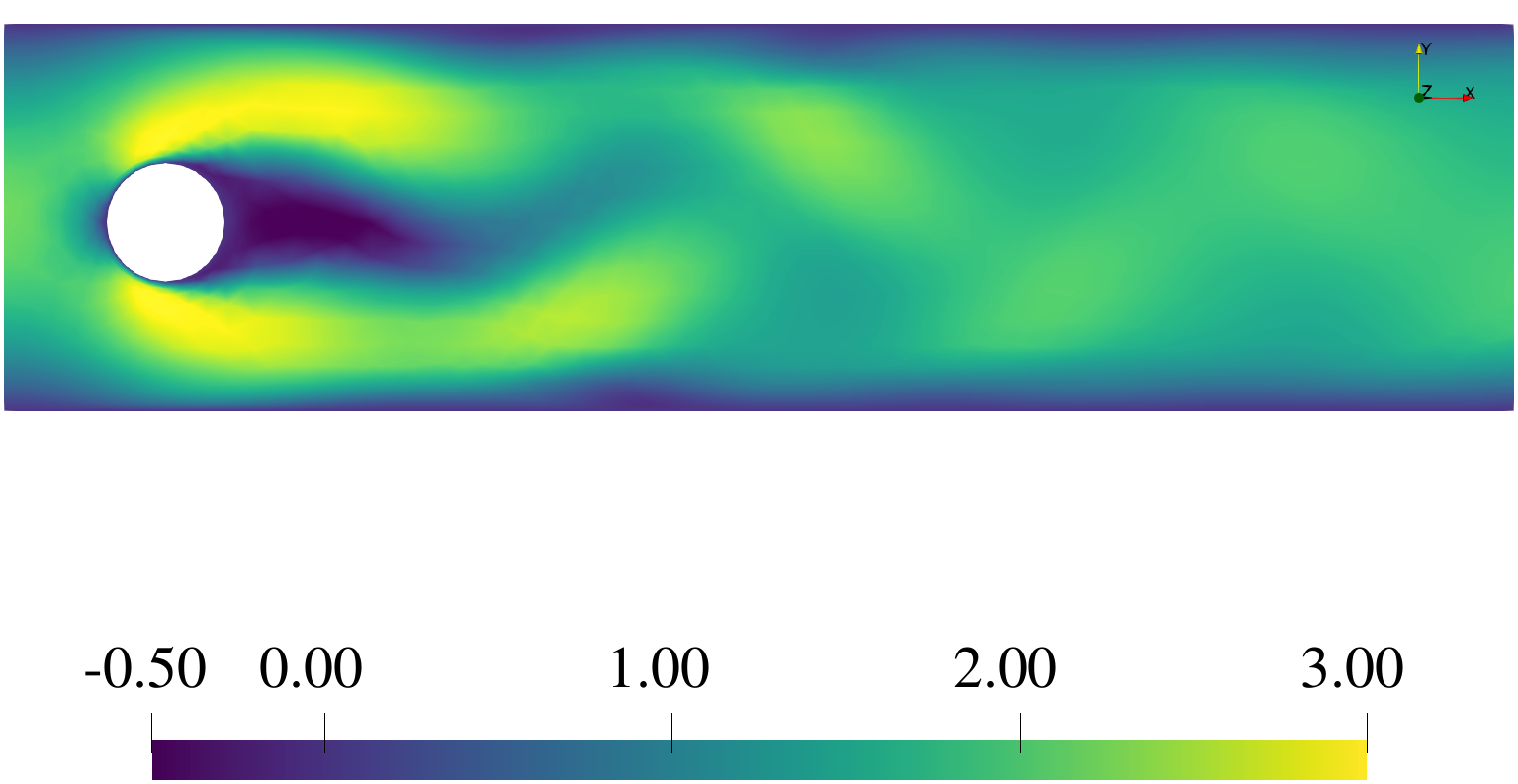}} 
		\hfill
		\subfigure{\includegraphics[trim={0 \cropb cm 0 1cm},clip,width=\x\linewidth]{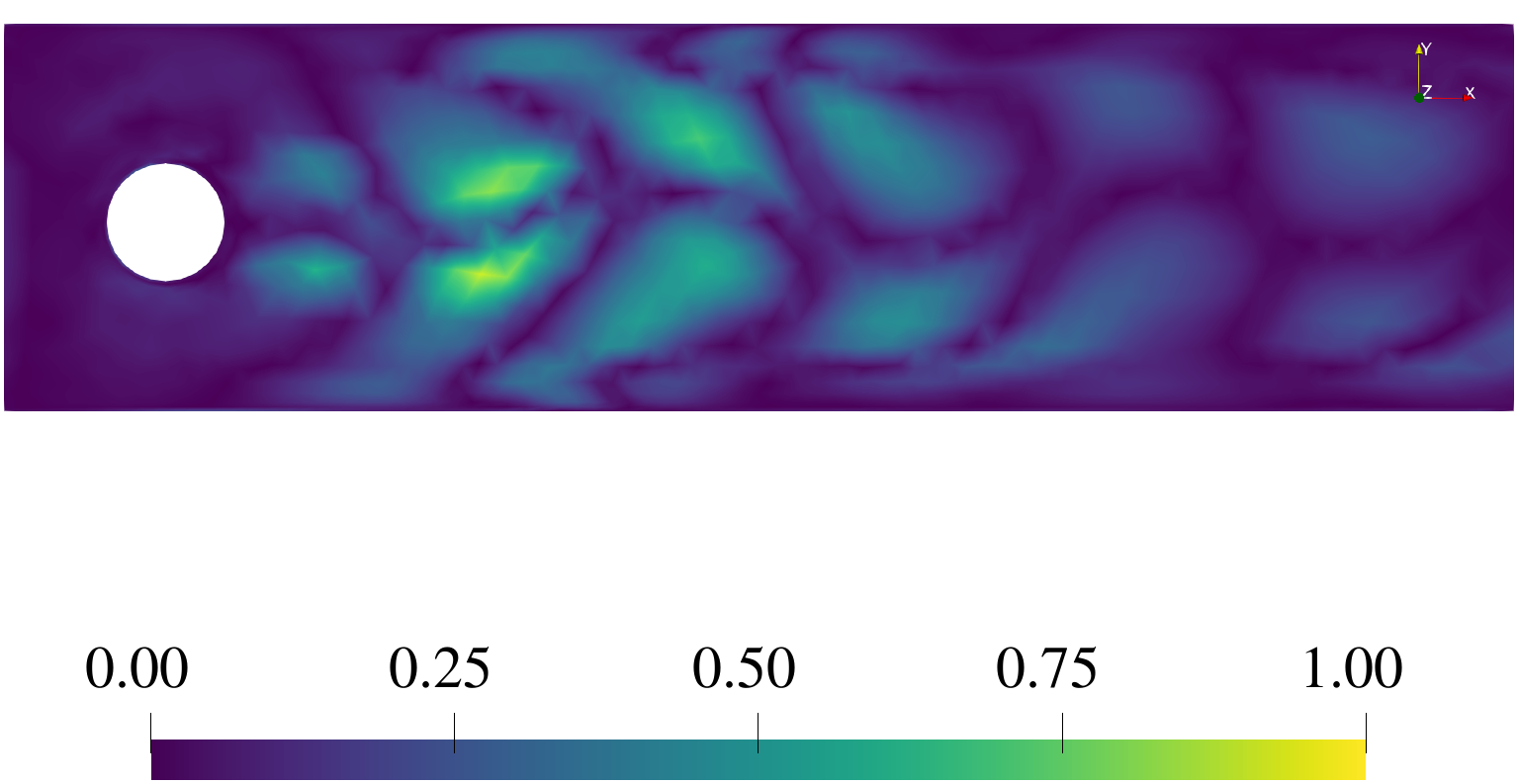}}  
		%  
		%% New Line (trim for color legend)
		%
		\subfigure{\includegraphics[trim={0 0 0 22cm},clip,width=0.33\linewidth]{./figures/traj_cyl/pred_u_200.png}}
		\put(-0.29\linewidth,22){Predicted Velocity $\tilde{u}$, \SI{}{\metre\per\second}}   
		\hfill
		\subfigure{\includegraphics[trim={0 0 0 22cm},clip,width=0.33\linewidth]{./figures/traj_cyl/gt_u_200.png}}
		\put(-0.29\linewidth,22){Target Velocity $u$, \SI{}{\metre\per\second}}      
		\hfill 
		\subfigure{\includegraphics[trim={0 0 0 22cm},clip,width=0.33\linewidth]{./figures/traj_cyl/abs_del_u_1.png}}   
		\put(-0.29\linewidth,22){Absolute Error $| \tilde{u} - u |$, \SI{}{\metre\per\second}}   
	\end{subfigmatrix}
	\caption{Side-by-side comparison of predicted velocity $u$ (left), the corresponding target velocity (mid) and their absolute difference (right) for rollouts of cylinderflow testset sample with highest \ac{rmse}.}
	\label{fig:compCYL}
\end{figure}
In order to give an idea of what a rollout looks like, Fig.~\ref{fig:compCYL} presents 5 snapshots of a 200-step rollout. The sample depicted was chosen as its predicted rollout has the highest \ac{rmse} of the testset.
Each row pictures from left to right the predicted flow field, the corresponding target and the absolute error between them.
One can observe that the model successfully predicts (left column) a flow field that regards the object placed at some random position in the left half of the domain and also learned that for certain combinations of diameter and position vortex shedding occurs.
But it is noticeable that the prediction of the model increasingly misaligns with its target over multiple rollout steps.
%However, it is noticeable that the model fails to predict the correct vortex shedding frequency, which results in an increasing misalignment between target and predicted flow field over multiple rollout steps.

This pre-study showed that the proposed approach produces comparable results with one sample distributed over eight \acp{gpu} to a similar single \ac{gpu} implementation.
However, the significant difference in the 600-step rollout errors might be an indication that the halo exchange is not passing all required information.

\subsection{Application to 3D Turbine}\label{subsec:H01S}
\begin{figure}[t!]
	\begin{adjustwidth}{1.5cm}{1.5cm}
		% trim={<left> <lower> <right> <upper>}
		\begin{subfigmatrix}{3}
			%
			% blocks
			%
			\subfigure[\label{fig:H01S_Isoview}]{\includegraphics[trim={16cm 3cm 10cm 0},clip, height=15\baselineskip ,keepaspectratio]{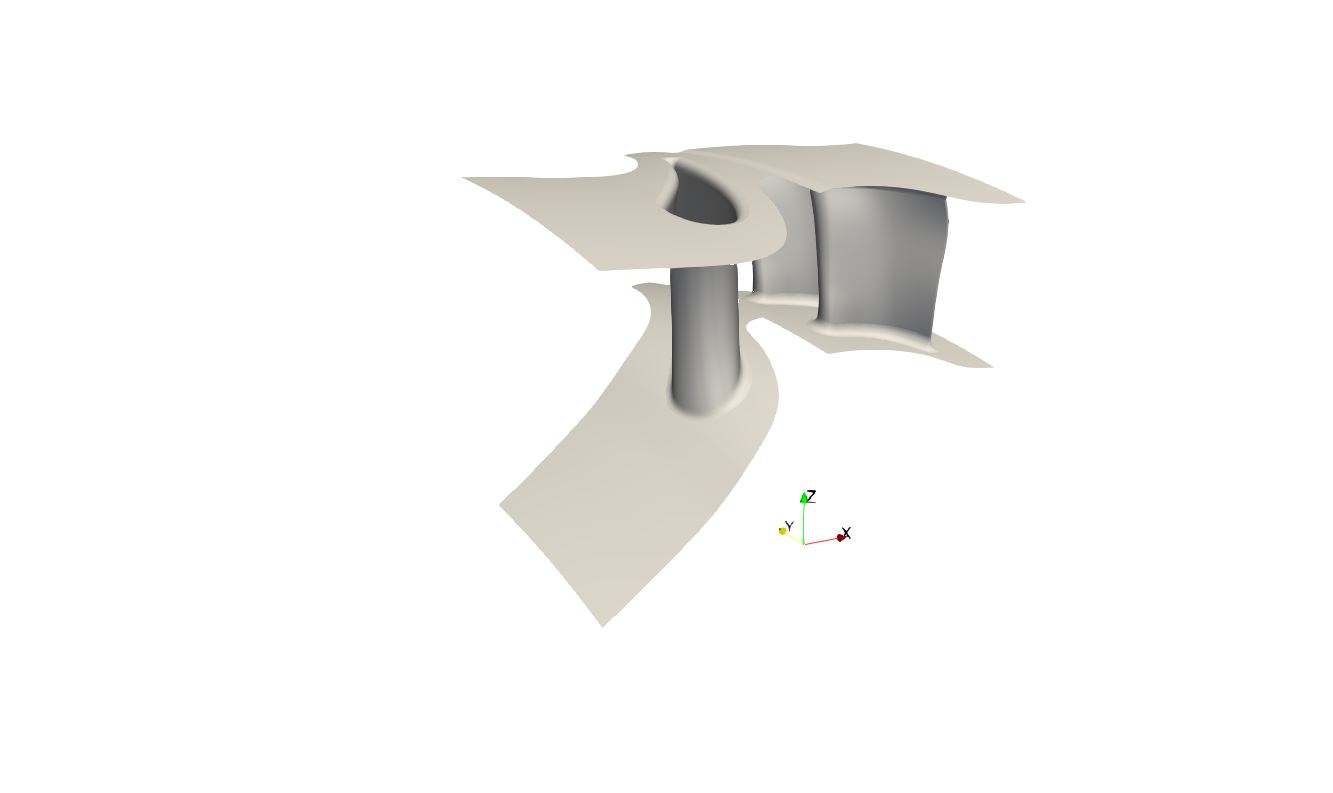}}  
			\hfill
			\subfigure[\label{fig:H01S_EvalPlanes}]{\includegraphics[trim={14cm 3cm 12cm 0},clip, height=15\baselineskip ,keepaspectratio]{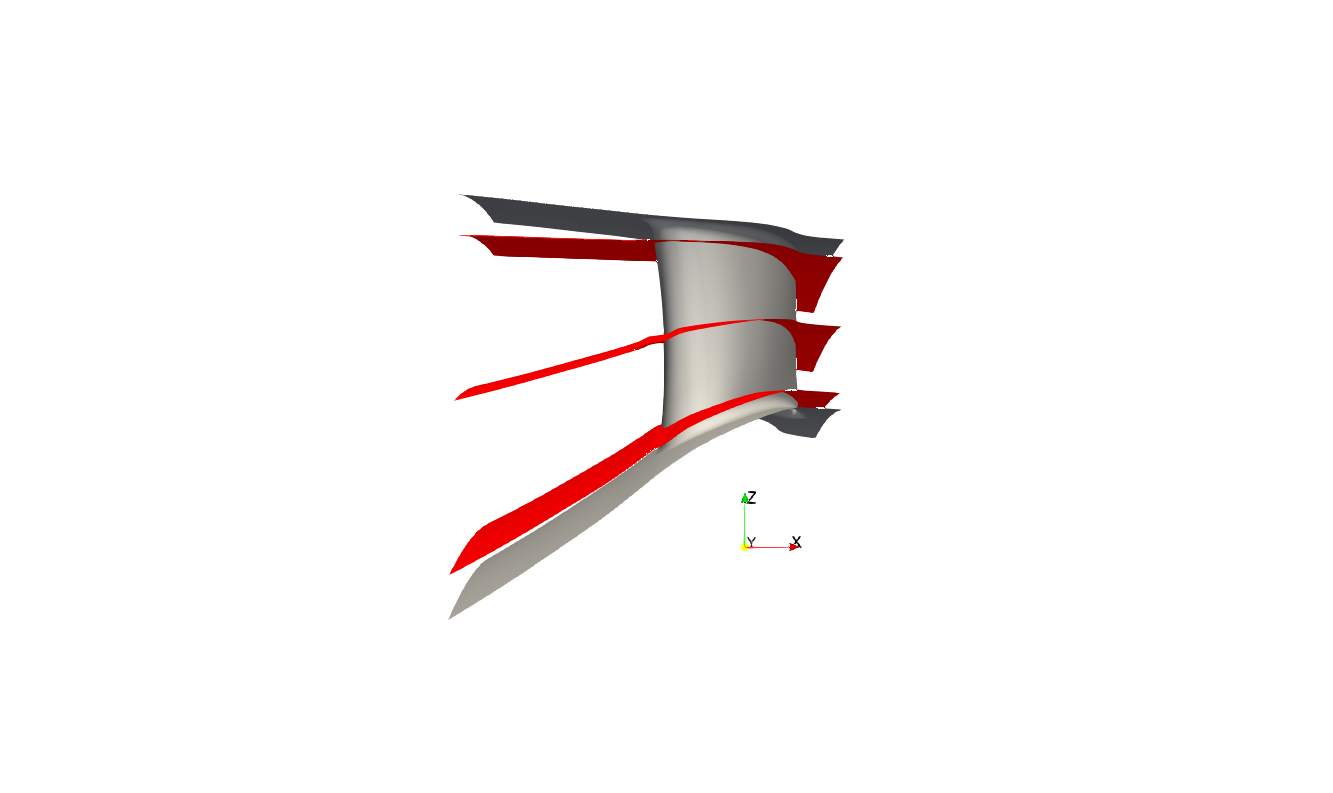}}
			\put(-124pt,90pt){\footnotesize{90\%}}
			\put(-124pt,60pt){\footnotesize{50\%}}
			\put(-124pt,30pt){\footnotesize{10\%}}
			\hfill
			\subfigure[\label{fig:H01S_Partitions}]{\includegraphics[trim={12cm 0 10cm 0},clip, height=15\baselineskip ,keepaspectratio]{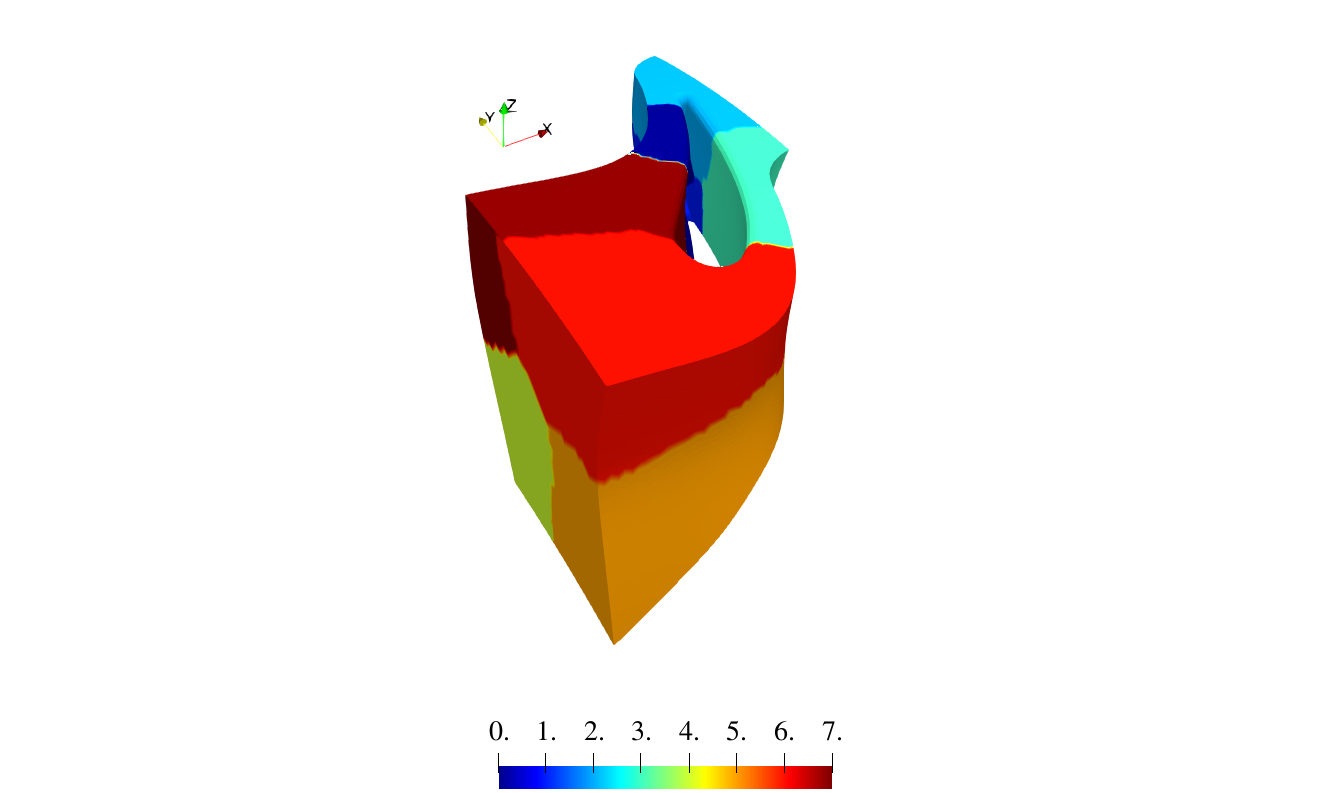}}
			\put(-0.25\linewidth,14){Partition Index} 
		\end{subfigmatrix}
	\end{adjustwidth}
	
	\caption{(a) Isometric view on numerical setup of turbine stage. (b) Three evaluation span levels (10\%, 50\%, 90\%) in turbine stator domain. (c) Partitioned stator domain with partition indices.}
	\label{fig:compH01S_Setup}
\end{figure}

A second dataset of a representative state-of-the-art turbine stage of an aircraft engine is now used to apply the proposed approach to a flow field with about $10^6$ points.
This turbine dataset consists of 100 geometry variations of a turbine stator from simulations of a complete turbine stage.
An isometric view on the numerical setup is presented in Fig.~\ref{fig:H01S_Isoview}.
The dataset consists of turbine stator domains with $0.75\times10^6$ points in average and $0.97\times10^6$ points maximal.
Rolls-Royce's HYDRA \cite{HYDRA} was employed as a \ac{cfd} solver to solve the \ac{urans} equations.
The one equation turbulence model from Spalart-Allmaras~\cite{spalart} is used for resolving eddy viscosity.
A revolution or period is resolved with 50 timesteps and in average 5 to 6 periods were needed to reach a converged time varying cyclic solution.
The numerical grid and all timesteps of the last period are saved and then analogue to \cite{pfaff2021learning} preprocessed into a \ac{ml} dataset.
A small subset (5 trajectories) of the dataset will be used as validation set and is not provided during training.
In Fig.~\ref{fig:H01S_EvalPlanes} the evaluation planes that are used here for visual inspection are pictured. As these cuts on a fixed relative span do not regard mesh cells, cutting artifacts may occur in the pictures.
According to the approach presented in the previous section \ref{ch:method} the flow domain is partitioned in eight parts. 
Fig.~\ref{fig:H01S_Partitions} presents an isometric view colored by partitions.

All previously introduced input node features $\mathbf{f}_\text{i,t}$ are included in the turbine dataset input.
Additionally, the velocity in $z$ direction $w$ is used, as a \ac{3d} flow field is considered now.
Output node features $\mathbf{p}_\text{i,t}$ are analogously defined with an additional $\Delta w$ accounting for the difference in the third velocity component and $\Delta \rho$ which regards the difference in the density field of the compressible flow.
Training noise levels were set according to \cite{pfaff2021learning} and not varied throughout this work.

In contrast to the previous section, no single \ac{gpu} application is possible as this would exceed the device memory using the current implementation of \ac{mgn}.
In order to be able to train the model on samples of this size, adjustments regarding the model size had to be made.
While exploring the proposed distribution approach, one overall intention of this work is to limit the training to a single node with eight devices.
On the one hand this limits the amount of resources that each model configuration requires and on the other hand multi-node training with hard synchronisation between devices is not efficient in parallel. 
Section \ref{subsec:perf} further elaborates on this topic.
Hence, the model size is reduced to either $8$ or $10$ GraphNet blocks and a lowered latent vector size of $96$ or $64$.

\begin{table}[b!]
	\begin{center}
		\caption{Comparison of of mean and standard error of \ac{rmse} over the turbine validation set after $5\times10^6$ optimizer steps}
		\label{tab:CompH01S} 
		\begin{tabular}{lccP{2.1cm}P{2.1cm}P{2.1cm}}
			\toprule
			\textbf{Model} & $K$ & $l$ &  \textbf{RMSE \hspace{10pt} 1-step}\hspace{10pt} $\times 10^{-3}$ & \textbf{RMSE 50-step}\hspace{10pt} $\times 10^{-3}$ & \textbf{RMSE Nextstep} $\times 10^{-3}$\\
			\midrule
			\textsc{MGN-Halo} & 8 &	96 &$271.80\pm14.34$&$456.38\pm12.52$&$270.24\pm1.93$\\
			\textsc{MGN-Halo} & 10 & 64 &$244.79\pm11.08$&$424.95\pm11.56$&$243.08\pm1.53$\\
			\textsc{MGN-NoComm} & 8 & 96 &$100.27\pm4.93$&$101.04\pm4.92$&$98.22\pm0.64$\\
			\textsc{MGN-NoComm} & 10 & 64 &$96.76\pm4.28$&$99.51\pm3.9$&$94.10\pm0.53$\\
			%			\bottomrule
		\end{tabular}	
	\end{center}
\end{table}
Table~\ref{tab:CompH01S} presents all model configurations that were trained on the turbine dataset. Two different training approaches are compared to each other. 
One is denoted with \textsc{MGN-Halo} and employs the proposed halo exchanging multi-device learning method. 
Through gradient accumulation a batchsize of two is achieved with two consecutive samples.
The other is denoted \textsc{MGN-NoComm} and trains on the same number of devices but in traditional fashion, e.g. without halo exchange. 
Additionally, the second approach does not use gradient accumulation so that one sample per batch per device is used.
The gradients of each model replica are synchronized after a loss backward pass in both approaches.
To draw a direct comparison, the same parameter variation for the message passing steps and latent vector size is used.
For inference, a single \ac{gpu} memory is large enough to hold a complete turbine stator domain with up to and over $10^6$ points.
\begin{figure}[b!]
	\begin{adjustwidth}{1.8cm}{1.8cm}
		% trim={<left> <lower> <right> <upper>}
		\begin{subfigmatrix}{3}
			\subfigure{\includegraphics[trim={18cm 3cm 18cm 0},clip, width=0.32\linewidth, height=0.17\paperheight,keepaspectratio=false ]{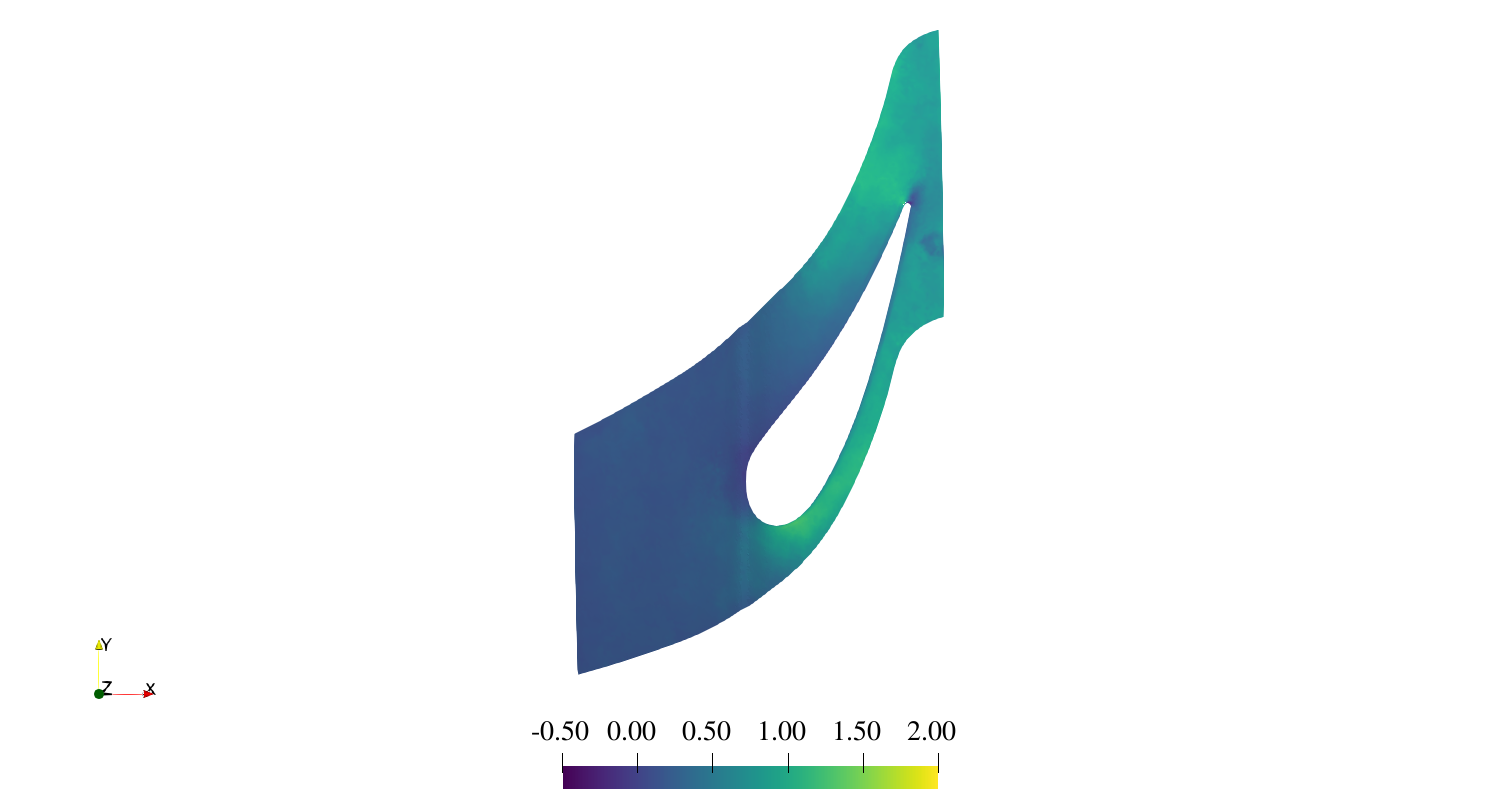}}
			\put(-101pt,76pt){\rotatebox{90}{10\% Span}}   
			\hfill
			\subfigure{\includegraphics[trim={18cm 3cm 18cm 0},clip, width=0.32\linewidth, height=0.17\paperheight,keepaspectratio=false ]{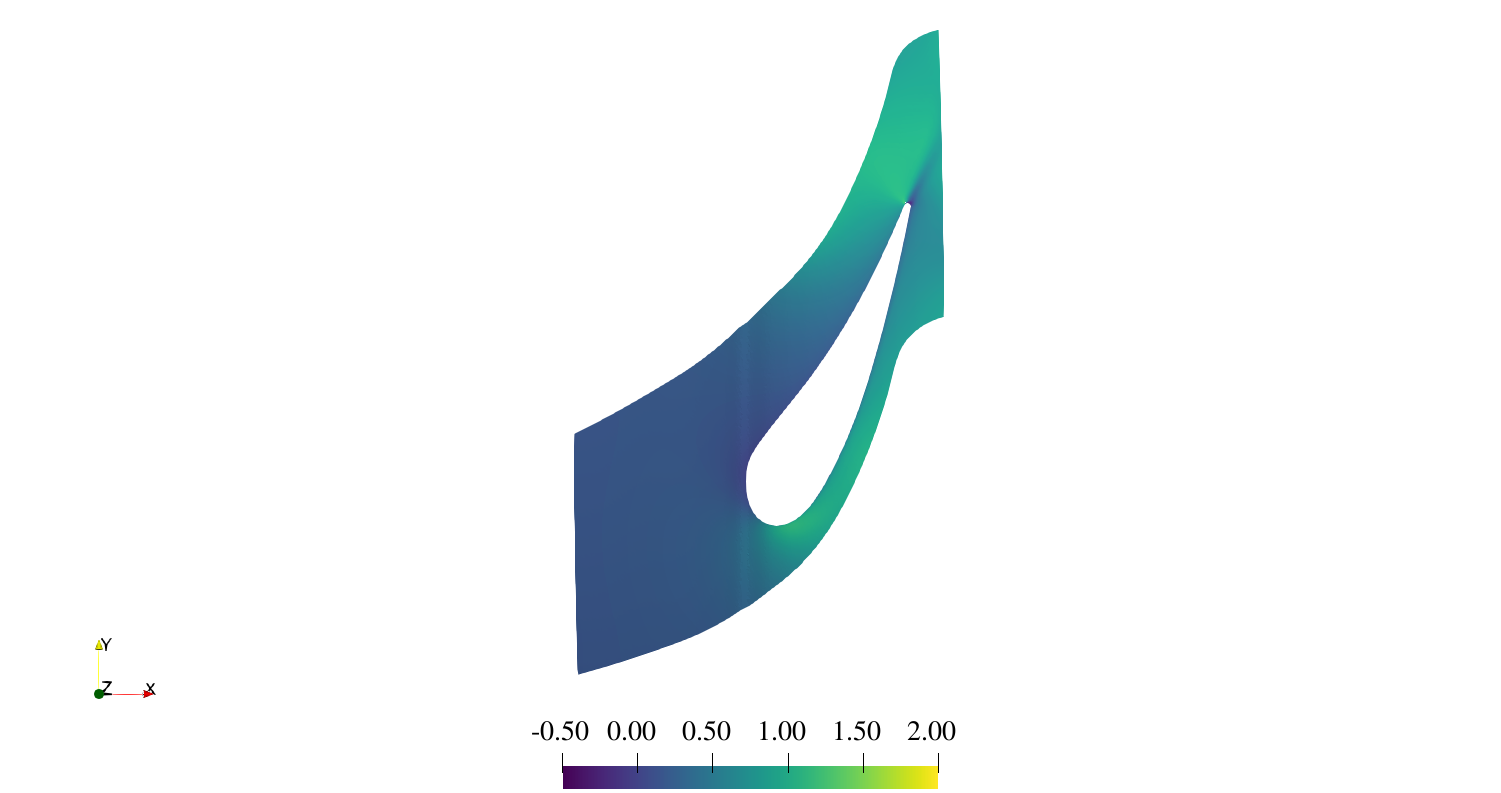}}
			\hfill
			\subfigure{\includegraphics[trim={18cm 3cm 18cm 0},clip, width=0.32\linewidth, height=0.17\paperheight,keepaspectratio=false ]{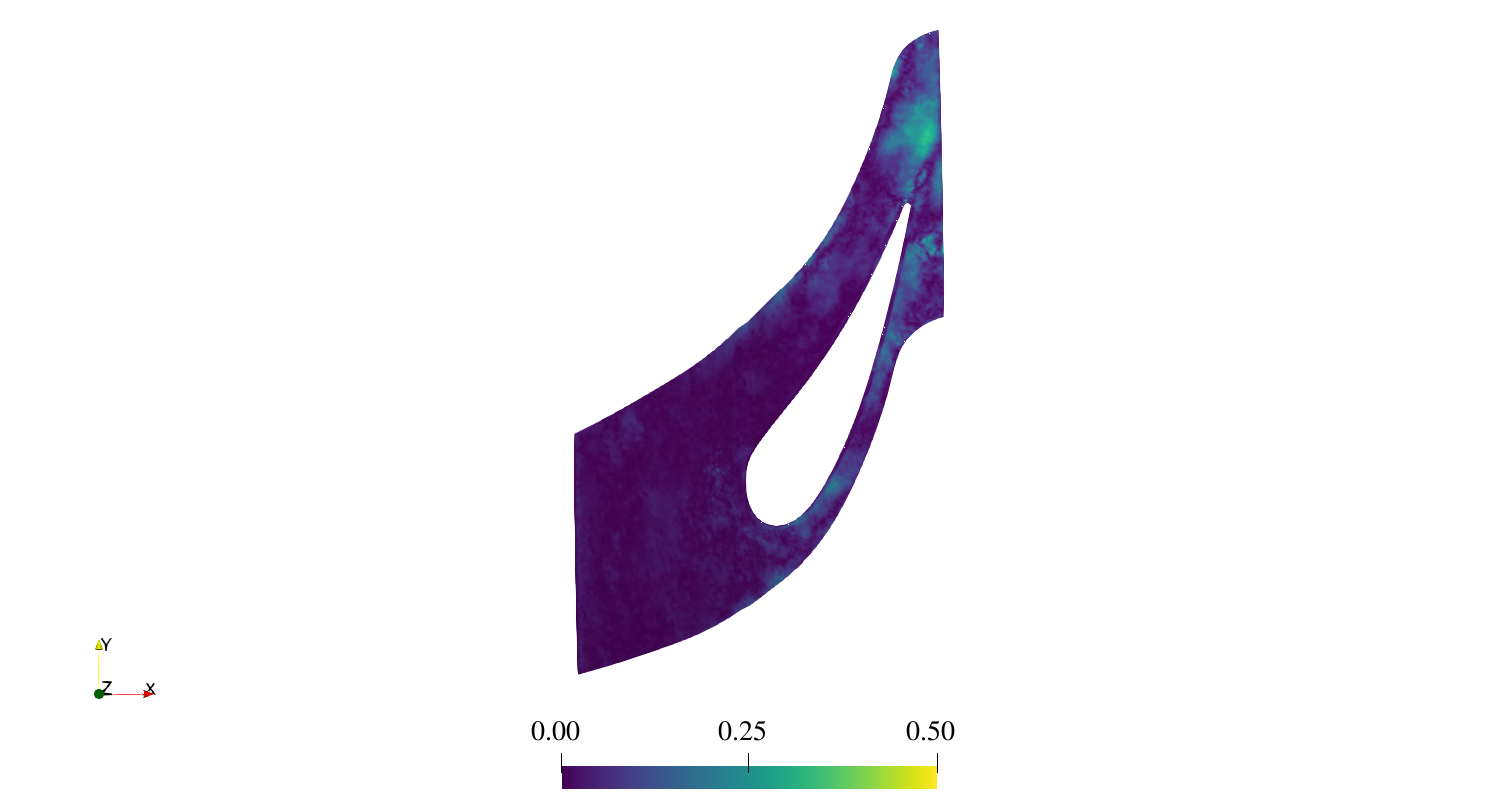}}  
			%
			%% New Line
			%
			\subfigure{\includegraphics[trim={18cm 3cm 18cm 0},clip, width=0.32\linewidth, height=0.18\paperheight,keepaspectratio=false  ]{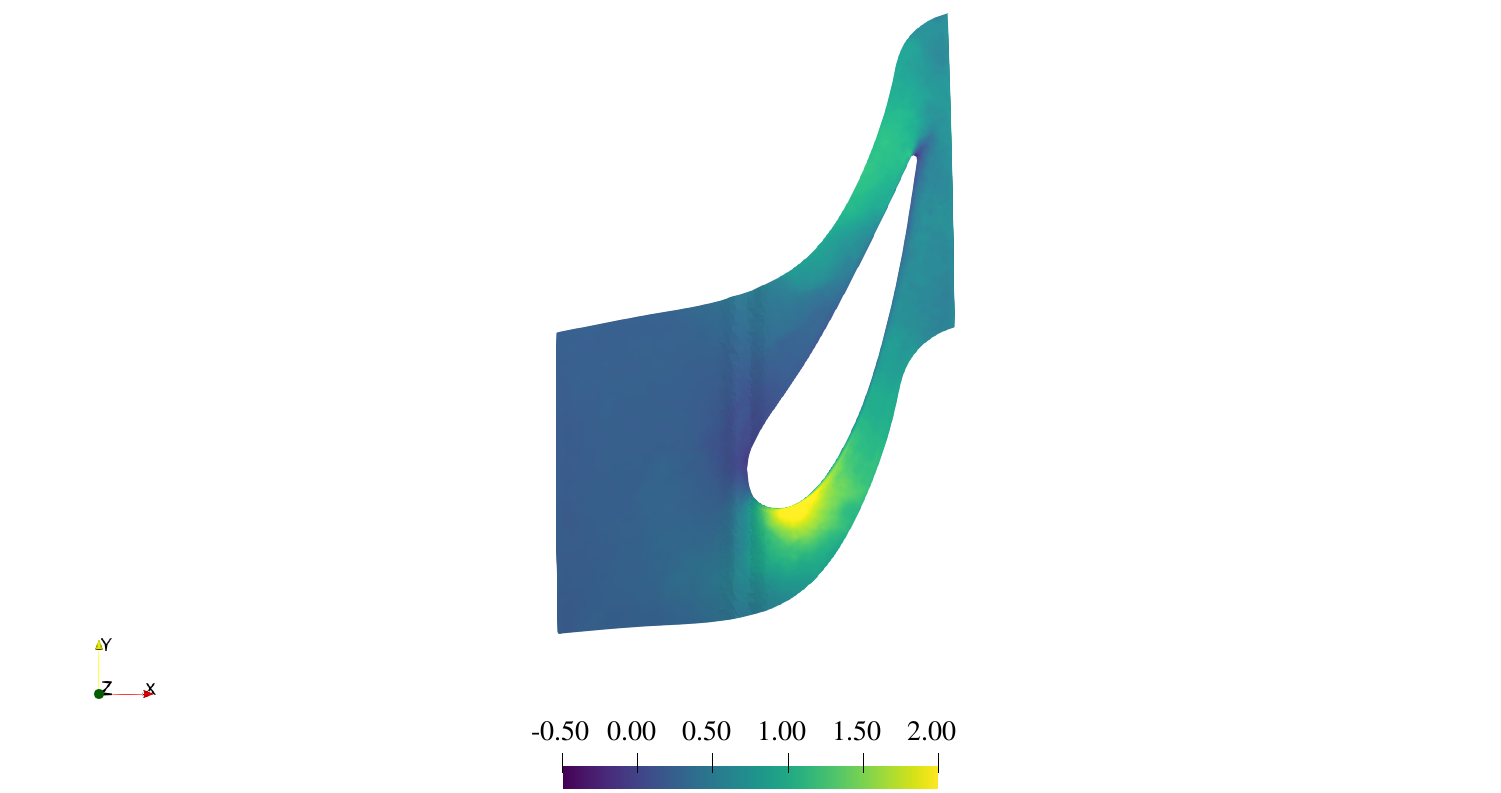}}
			\put(-104pt,95pt){\rotatebox{90}{50\% Span}}   
			\hfill
			\subfigure{\includegraphics[trim={18cm 3cm 18cm 0},clip, width=0.32\linewidth, height=0.18\paperheight,keepaspectratio=false  ]{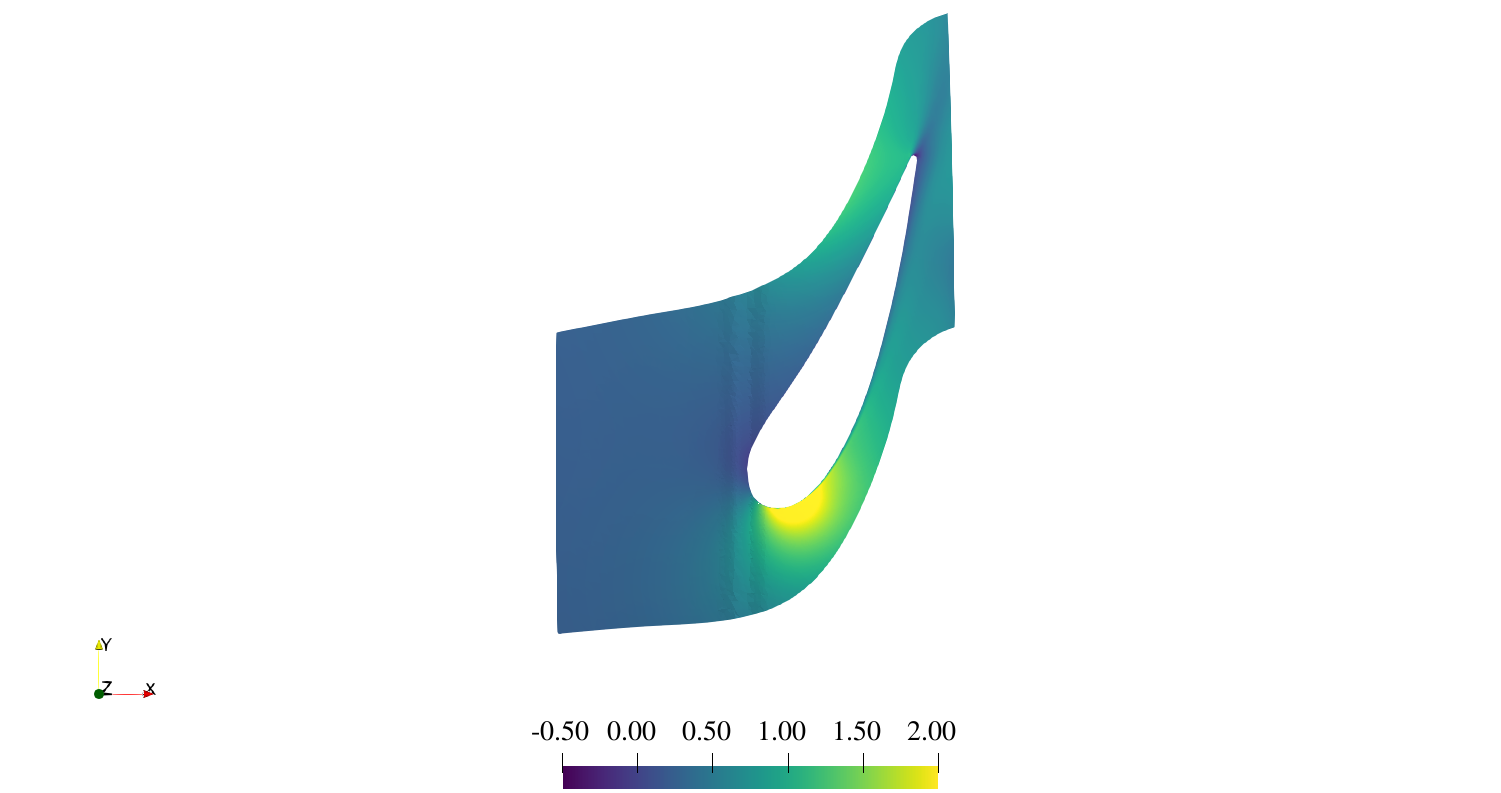}}
			\hfill
			\subfigure{\includegraphics[trim={18cm 3cm 18cm 0},clip, width=0.32\linewidth, height=0.18\paperheight,keepaspectratio=false  ]{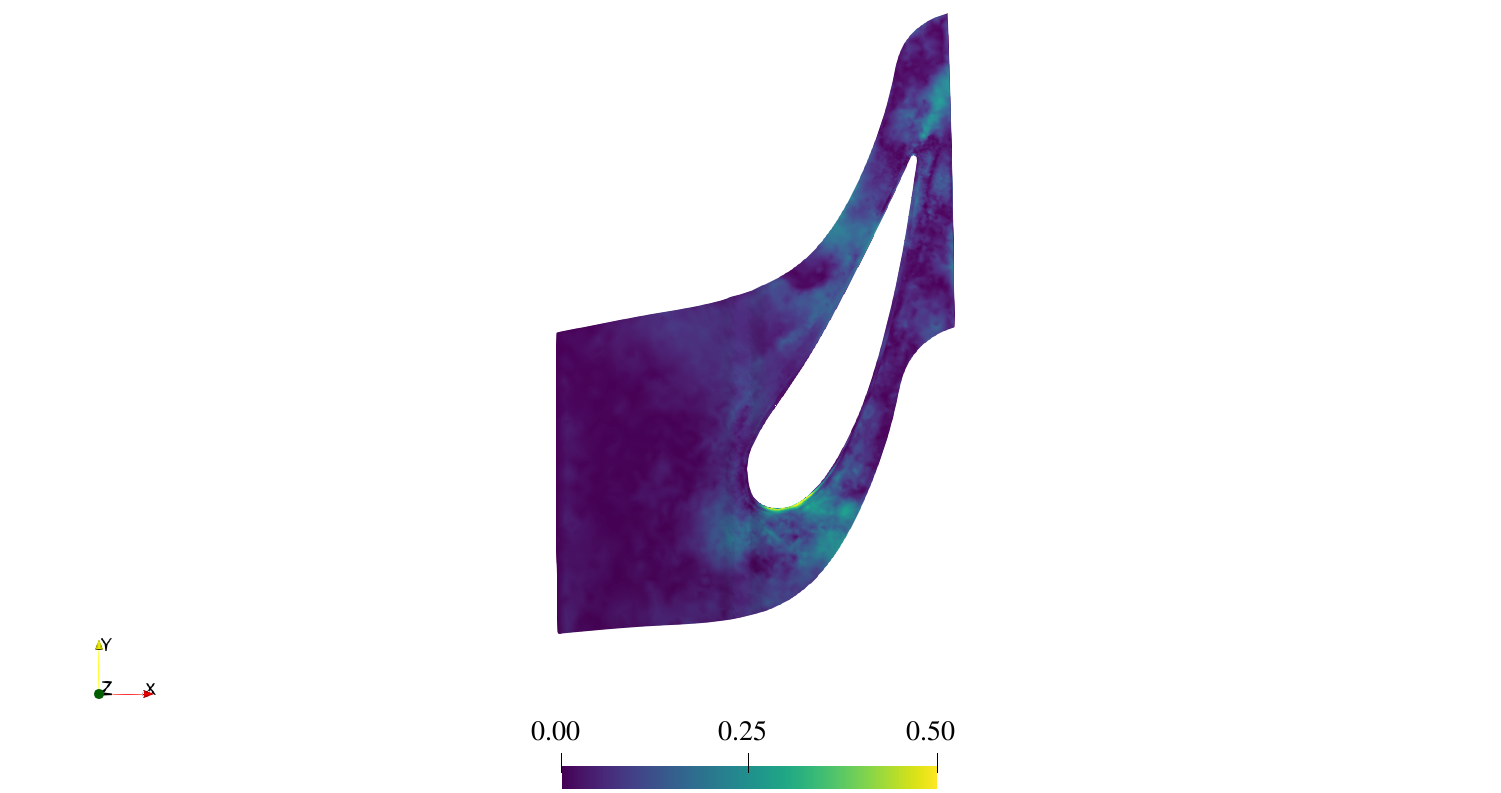}}  
			%  
			%% New Line 
			%
			\subfigure{\includegraphics[trim={18cm 3cm 18cm 0},clip, width=0.32\linewidth, height=0.19\paperheight,keepaspectratio=false  ]{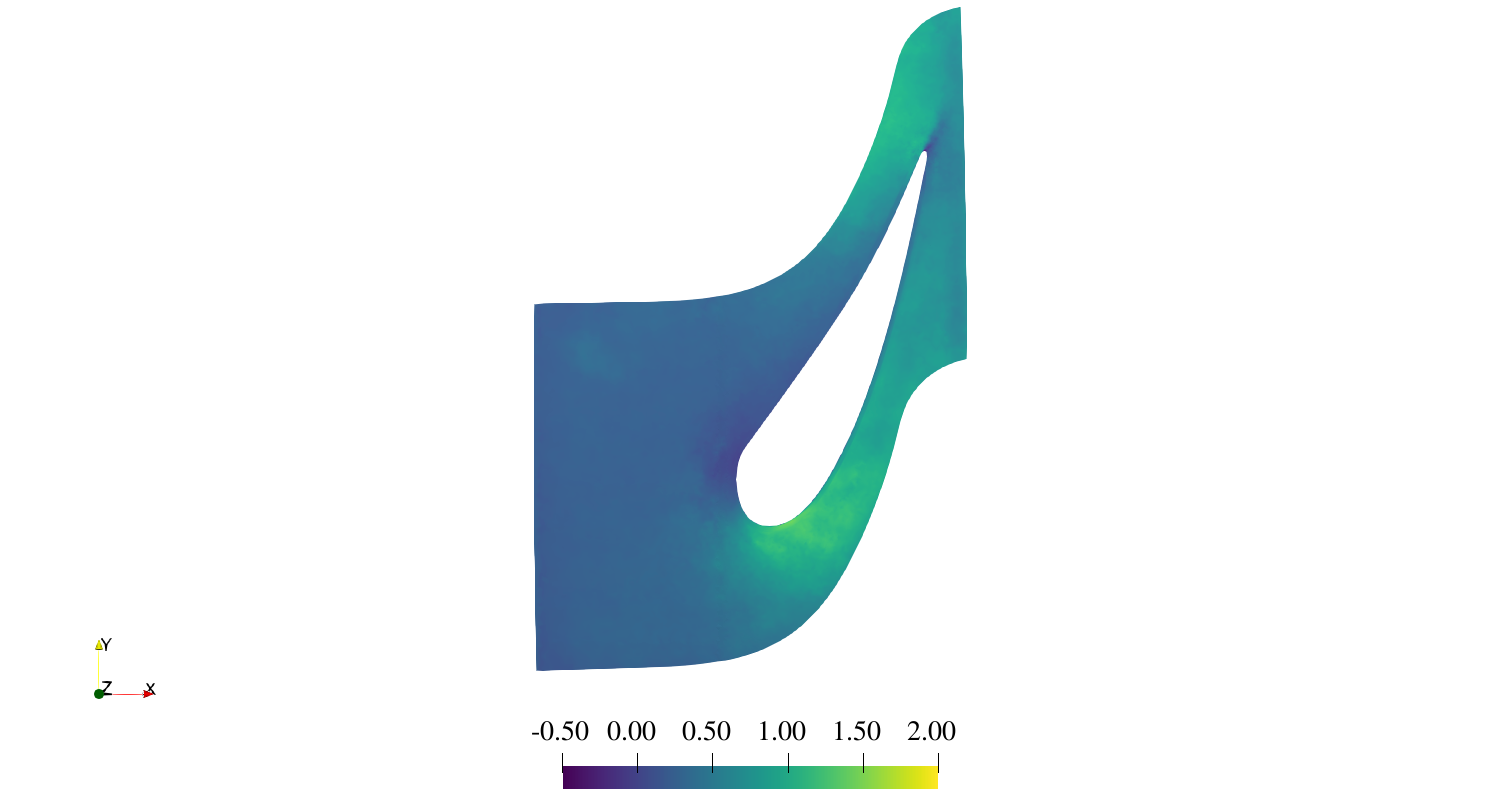}}
			\put(-109pt,104pt){\rotatebox{90}{90\% Span}}   
			\hfill
			\subfigure{\includegraphics[trim={18cm 3cm 18cm 0},clip, width=0.32\linewidth, height=0.19\paperheight,keepaspectratio=false  ]{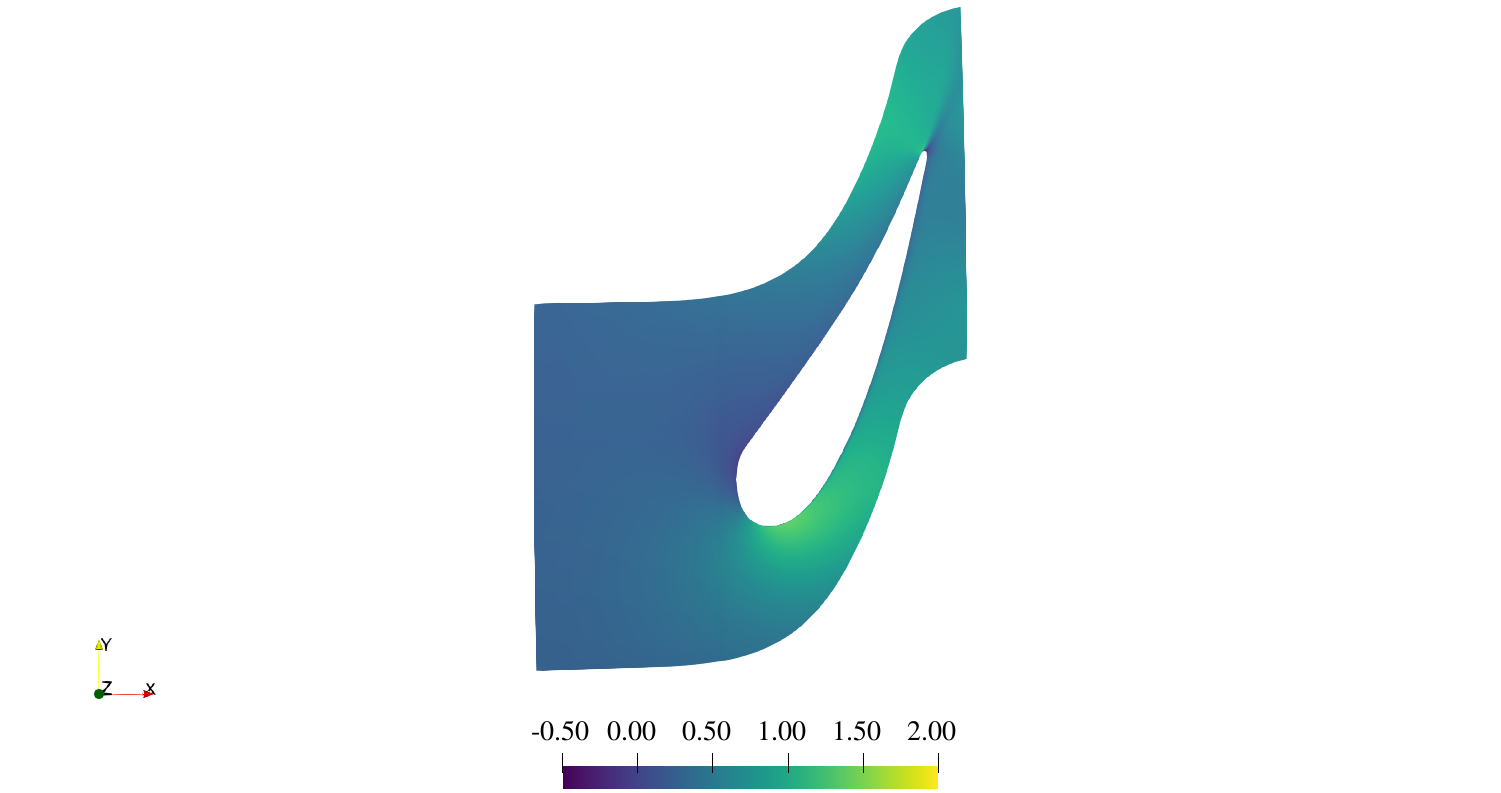}}
			\hfill
			\subfigure{\includegraphics[trim={18cm 3cm 18cm 0},clip, width=0.32\linewidth, height=0.19\paperheight,keepaspectratio=false  ]{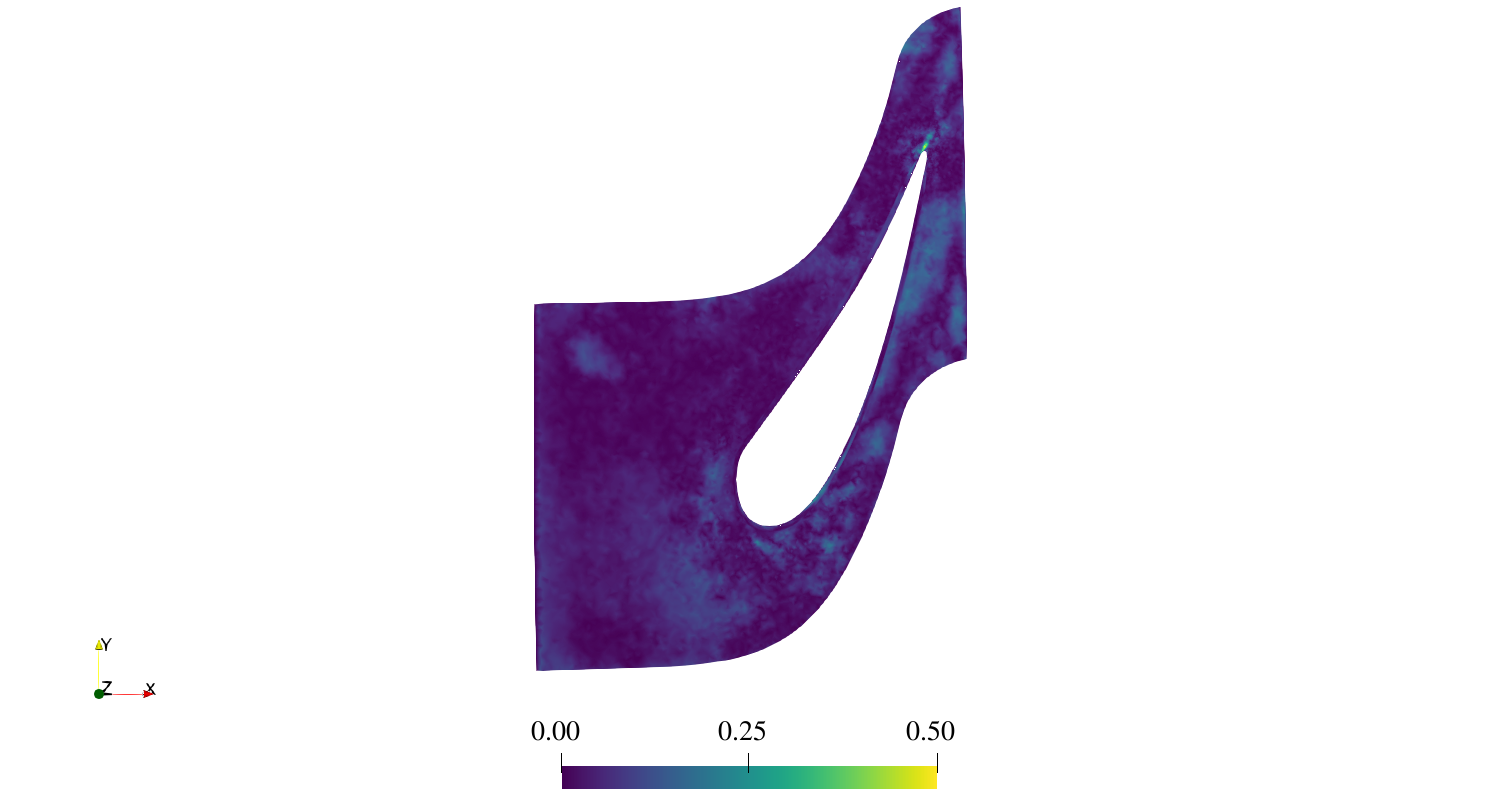}}  
			%
			% New Line (trim for color legend)
			%
			\subfigure{\includegraphics[trim={18cm 0 18cm 25cm},clip, height=0.19\paperheight ,keepaspectratio]{./figures/turbine/Hvd8_10-64_90percSpan_VeloOutXTarget.png}}
			\put(-0.29\linewidth,22){Predicted Velocity $\tilde{u}$, \SI{}{\metre\per\second}} 
			\hfill
			\subfigure{\includegraphics[trim={18cm 0 18cm 25cm},clip, height=0.19\paperheight ,keepaspectratio]{./figures/turbine/Hvd8_10-64_90percSpan_VeloOutXPred.png}}
			\put(-0.29\linewidth,22){Target Velocity $u$, \SI{}{\metre\per\second}}      
			\hfill
			\subfigure{\includegraphics[trim={18cm 0 18cm 25cm},clip, height=0.19\paperheight ,keepaspectratio]{./figures/turbine/Hvd8_10-64_90percSpan_RelError_VeloX.png}}
			\put(-0.29\linewidth,22){Normalized Error $E(\tilde{u})$}     
		\end{subfigmatrix}
	\end{adjustwidth}
	\caption{Side-by-side comparison of predicted velocity $u$ (left), corresponding target velocity (mid) and normalized error (right) for a single-step rollout over three different spans (row-wise) from \textsc{MGN-NoComm-10-64} of the turbine validation sample with the lowest \ac{rmse}.}
	\label{fig:compH01S}
\end{figure}

The comparison between 1-step and 50-step rollout \ac{rmse} already shows that both \textsc{MGN-NoComm} models are superior to the \textsc{MGN-Halo} versions as the error is nearly a factor of 3 lower.
Because the overall goal is to employ this model in a numerical solver, a \textit{Nextstep} error, which is a 1-step rollout starting from each step $t$ in the trajectory is evaluated. The results are in all four versions close to the 1-step rollout starting from the first timestep.
A temporal resolved distribution of the error is not presented as the error does not vary significantly throughout the trajectory for all models.

Figure~\ref{fig:compH01S} presents a 1-step rollout prediction of the sample with the lowest \ac{rmse} in the validation set for the superior \textsc{MGN-NoComm-10-64} model.
The three previously introduced evaluation planes (cf. Fig.~\ref{fig:H01S_EvalPlanes}) are used to visualize the flow field.
The corresponding version with halo exchange \textsc{MGN-Halo-10-64} is pictured in the appendix in Fig.~\ref{fig:compH01S_Halo}.
The prediction of the velocity in $x$-direction is compared to the corresponding target distribution and the resulting normalized error between both.
The normalized error for a prediction $\tilde{\varphi}$ is determined from $E(\varphi) = \mathrm{abs}(\tilde{\varphi} - \varphi) / (\mathrm{max}(\varphi) - \mathrm{min}(\varphi))$.
It becomes apparent from the right column of Fig.~\ref{fig:compH01S} that the model is able to predict the overall flowfield of the next timestep with an error of mostly below $10\%$. 
For the flow close to the casing or hub region in the first and third row of the figure the error is mostly around the trailing edge of the airfoil. 
For the mid section at $50\%$ span the error concentrates around the suction side in towards the leading edge.
Here, areas where the error is up to $25\%$ are visible as well as behind the trailing edge on all spans.
Instead the prediction of \textsc{MGN-Halo-10-64} in Fig.~\ref{fig:compH01S_Halo} shows large artifacts and prediction noise as well as significantly more areas with high error values.

\begin{figure}[b!]
	\begin{adjustwidth}{2.5cm}{2.5cm}
		\subfigure[\label{fig:tempmean}]{\includegraphics{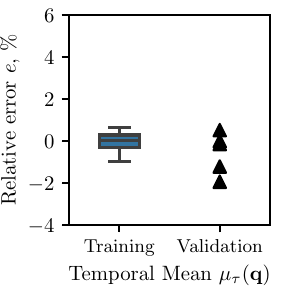}}
		\hfill
		\subfigure[\label{fig:tempstd}]{\includegraphics{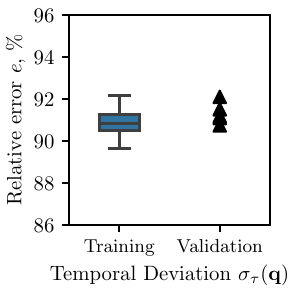}}
	\end{adjustwidth}
	\caption{Relative error $e$ of averaged temporal mean and standard deviation for prediction of trainingset and the 5 validation trajectories from \textsc{MGN-NoComm-10-64}}
	\label{fig:meanstd}
\end{figure}
From validation set statistics in Table~\ref{tab:CompH01S} and visual inspection of the validation set trajectory predicted best in Fig.~\ref{fig:compH01S} and \ref{fig:compH01S_Halo} it becomes apparent that model \textsc{MGN-NoComm} is superior to the current implementation of the proposed method.
Therefore, the following part focuses on limitations of \textsc{MGN-NoComm}.

In order to evaluate the capability of the model to learn temporal variability of the target, the mean and standard deviation over all timesteps $T$ of a trajectory are computed.
Then, the values are averaged over the $O$ output features and afterwards over all $N$ points of the domain.
For easier notation, the averaged temporal mean is denoted $\mu_{\tau}$ and the averaged temporal standard deviation $\sigma_{\tau}$.
For the target flow field $\mathbf{q}$ they are calculated as follows
\begin{align}
	\mu_{\tau}(\mathbf{q}) &= \frac{1}{N\cdot O \cdot T} \sum_{i=1}^{N} \sum_{g=1}^{O} \sum_{t=1}^{T} \mathbf{q}_\mathrm{i,g,t} & &\mathrm{, and} \\
	\sigma_{\tau}(\mathbf{q}) &= \frac{1}{N\cdot O } \sum_{i=1}^{N} \sum_{g=1}^{O}  \sqrt{\frac{1}{T} \sum_{t=1}^{T} \left( \mathbf{q}_\mathrm{i,g,t} - \mu \right)^2} & &\mathrm{, with}~ \mu =  \sum_{t=1}^{T} \mathbf{q}_\mathrm{i,g,t} \mathrm{~.}
\end{align}
These calculations were executed separately for training and validation set and repeated for the predicted flow fields $\tilde{\mathbf{q}}$.
To compare temporal mean of prediction and target, the relative error $e$ of $\mu_{\tau}$ is calculated as follows
\begin{align}
	e(\mu_{\tau}) = \frac{\mu_{\tau}(\tilde{\mathbf{q}}) - \mu_{\tau}(\mathbf{q})}{\mu_{\tau}(\mathbf{q})}.
\end{align}
The relative error for the temporal deviation $\sigma_{\tau}$ is calculated analogously.

In Fig.~\ref{fig:meanstd} the distribution of the relative error $e$ of the temporal mean and temporal deviation for the training and validation set is presented.
It becomes apparent from Fig.~\ref{fig:tempmean} that the model learned the mean flow field with a minimal and maximal relative error of below $1\%$.
The mean relative error as well as the $25\%$ and $75\%$ quantiles are very close to zero.
The distribution of the relative error for the test set is because of the low sample size presented as a scatter plot in the same figure.
For the validation set the spread of the values is higher than for the training set but $e(\mu_{\tau}(\mathbf{q})) \in [-1.94,0.52]$ so that one can argue that quantitatively the mean flow field is predicted correctly even for the testset.
However, Fig.~\ref{fig:tempstd} reveales the inability of \textsc{MGN-NoComm-10-64} to reflect the temporal variability. The relative error of the temporal deviation $\sigma_{\tau}$ is around $91\%$ in average for the training set which corresponds to a nearly stationary prediction without any temporal variation.
The model fails similarly for the validation set.

\begin{figure}[b!]
	\newcommand\x{10} %	0.33
	\begin{adjustwidth}{1cm}{1cm}
		% trim={<left> <lower> <right> <upper>}
		\begin{subfigmatrix}{4}
			\subfigure{\includegraphics[trim={18cm 3cm 18cm 0},clip, width=0.3\linewidth, height=0.17\paperheight,keepaspectratio=false  ]{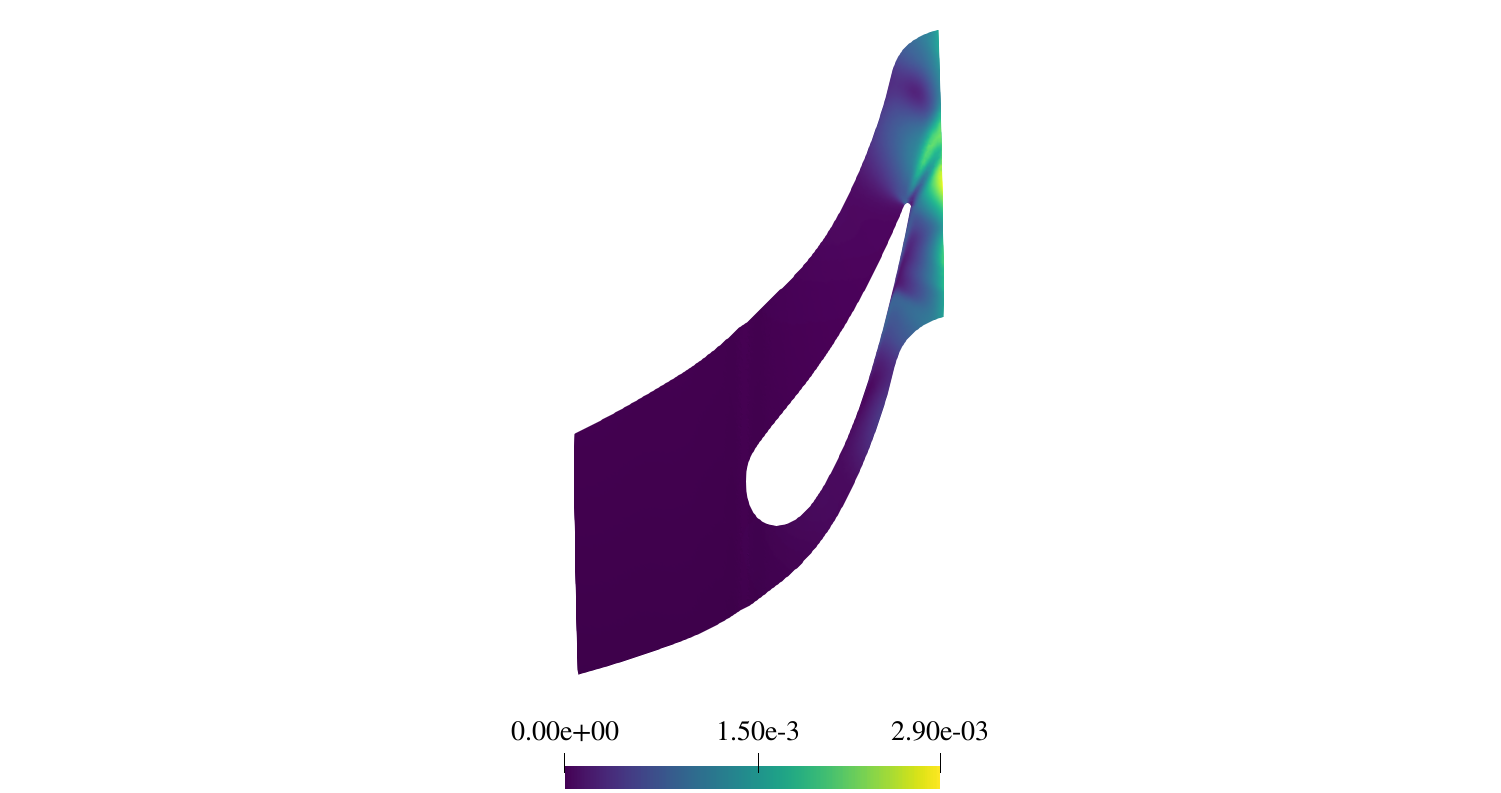}}
			\put(-105pt,70pt){\rotatebox{90}{10\% Span}}   
			\hfill
			\setcounter{subfigure}{0}
			\subfigure[\label{fig:compH01S_StdVeloX_a}]{\includegraphics[trim={18cm 3cm 18cm 0},clip, width=0.31\linewidth, height=0.17\paperheight,keepaspectratio=false  ]{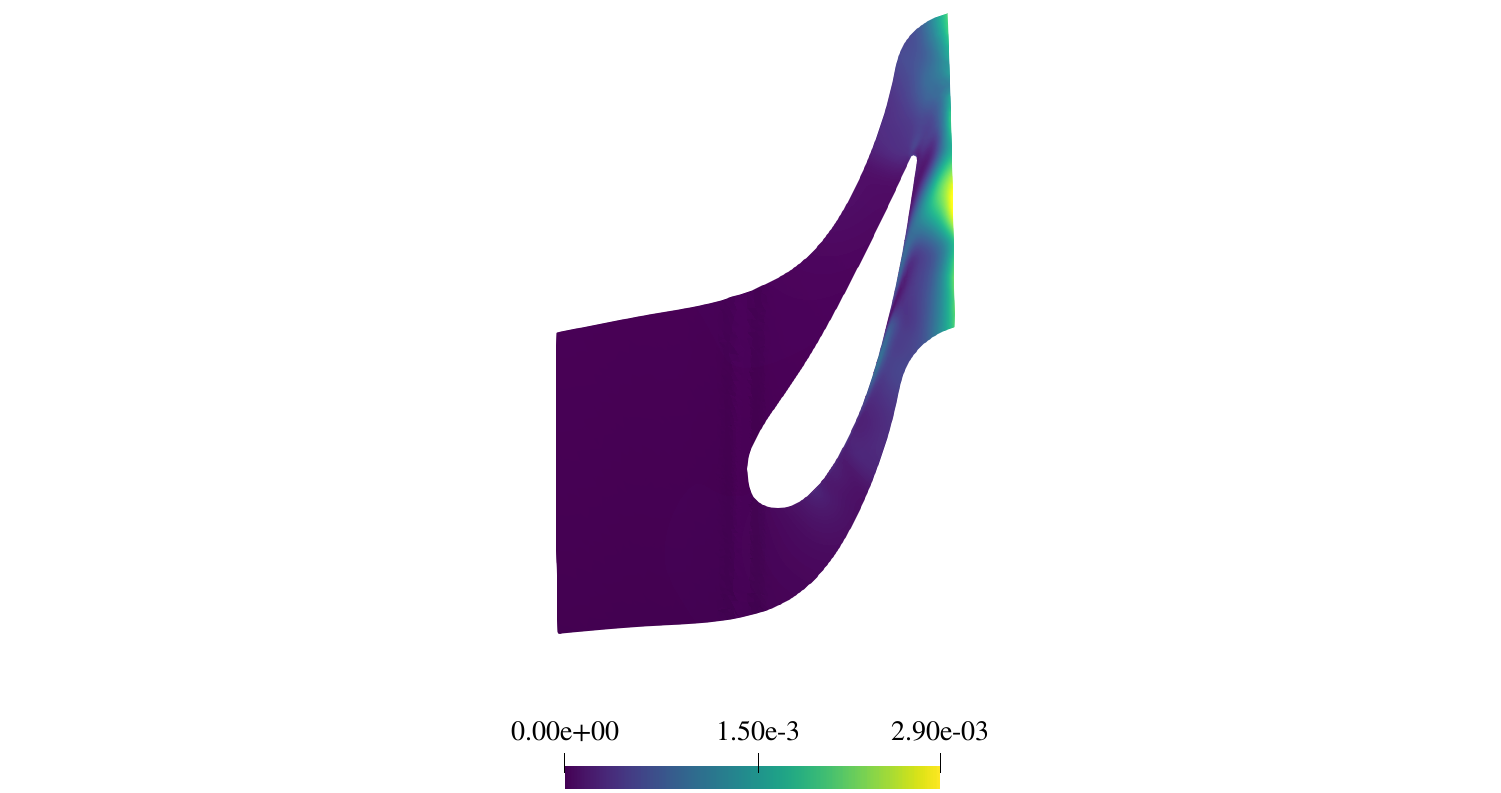}}
			\put(-112pt,85pt){\rotatebox{90}{50\% Span}}   
			\hfill
			\subfigure{\includegraphics[trim={18cm 3cm 18cm 0},clip, width=0.26\linewidth, height=0.17\paperheight,keepaspectratio=false  ]{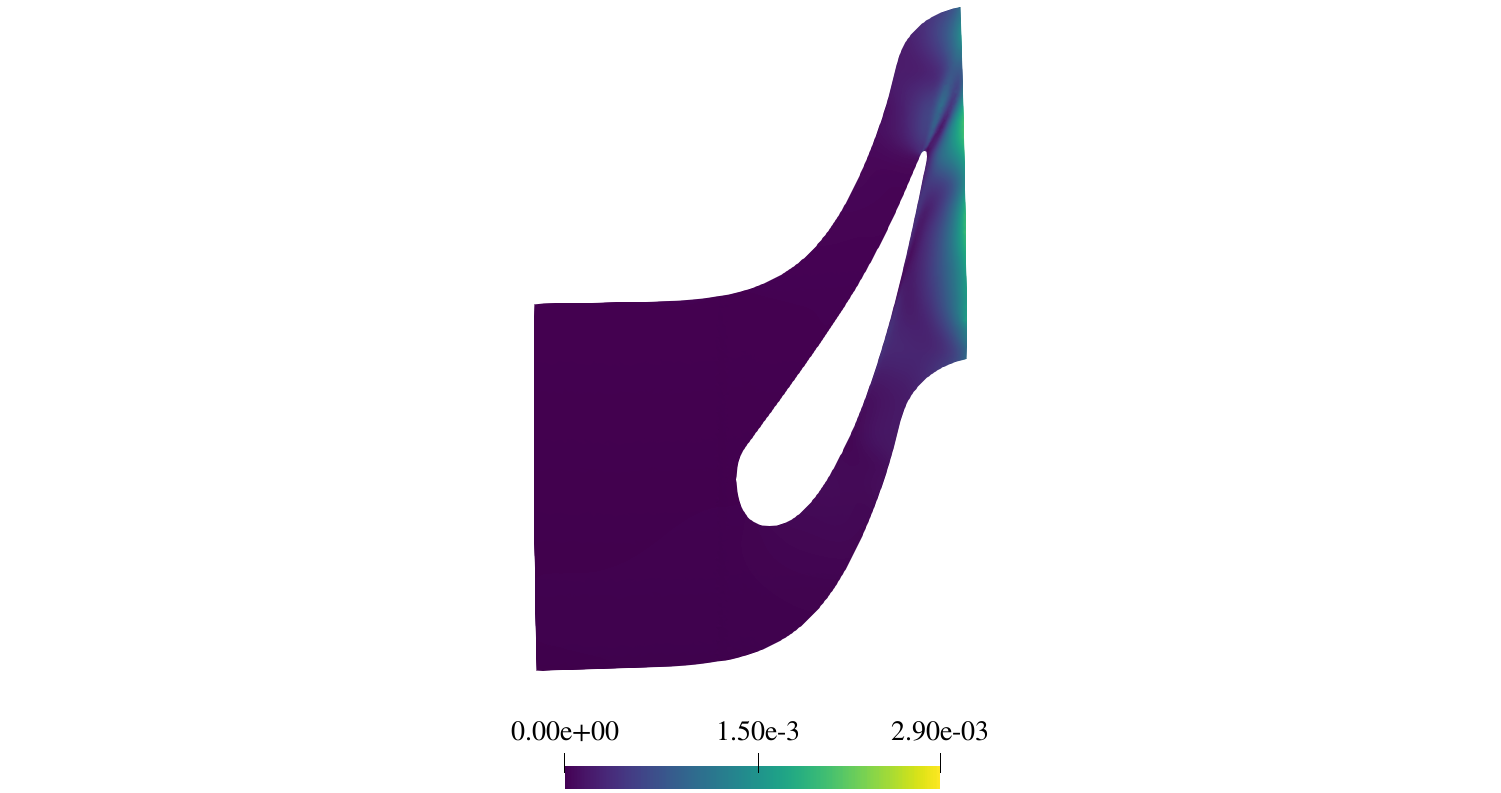}}  
			\put(-101pt,90pt){\rotatebox{90}{90\% Span}}   
			%
			% New Line (trim for color legend)
			%
			\subfigure{\includegraphics[trim={0 15cm 9cm 0},clip, width=0.1\linewidth ]{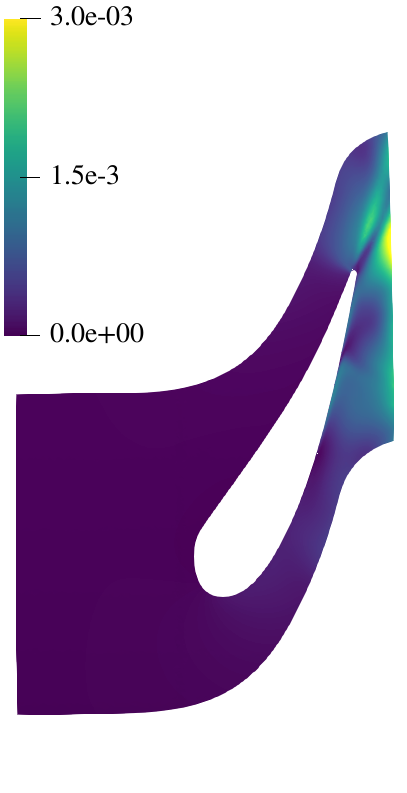}}
			\put(0.01\linewidth,1){\rotatebox{90}{\small Standard Deviation of Velocity $u$, \SI{}{\metre\per\second}}}     
		\end{subfigmatrix}
		\begin{subfigmatrix}{4}
			%
			% blocks
			%
			\subfigure{\includegraphics[trim={18cm 3cm 18cm 0},clip, width=0.3\linewidth, height=0.17\paperheight,keepaspectratio=false ]{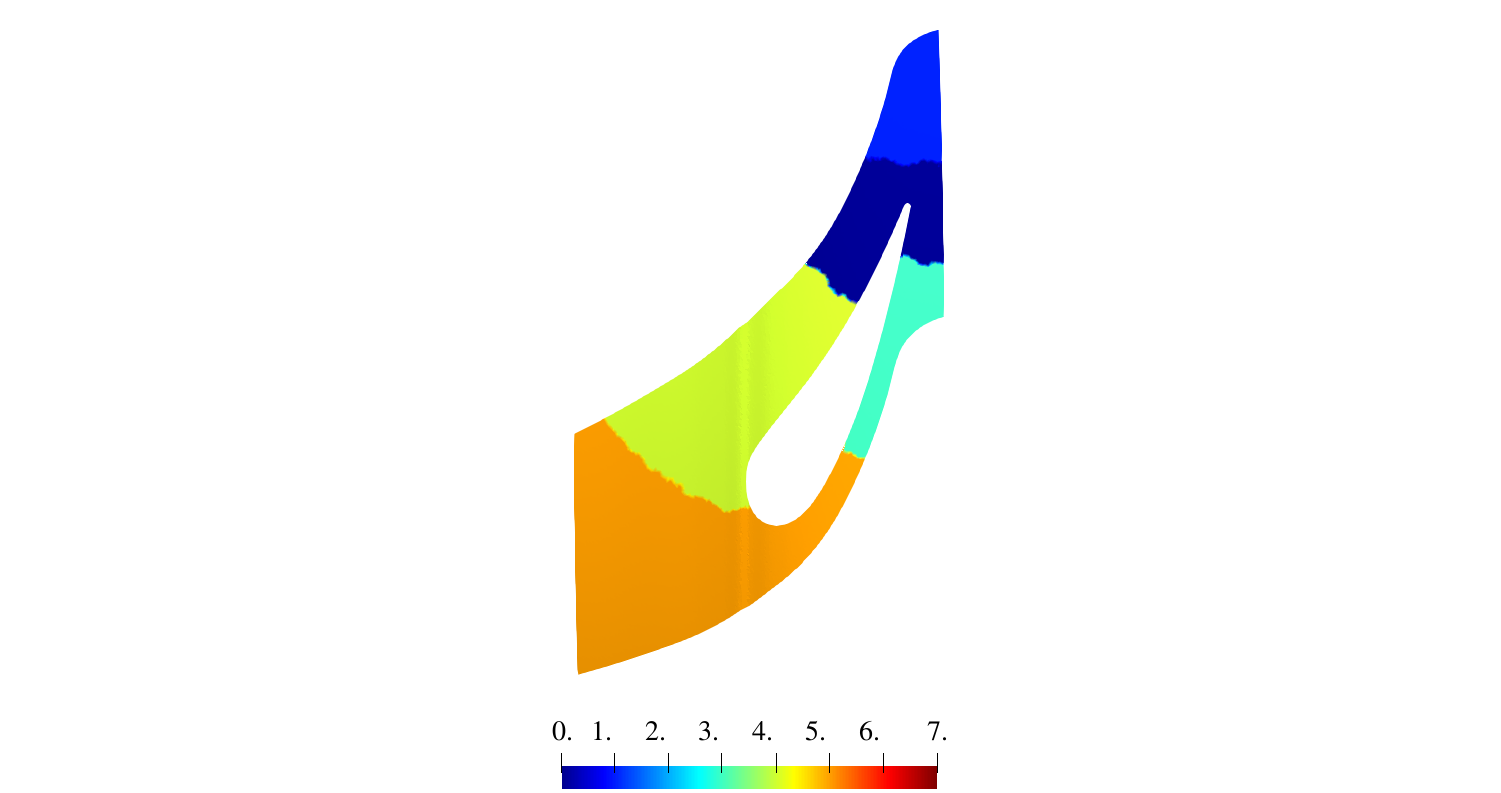}}
			\put(-105pt,70pt){\rotatebox{90}{10\% Span}}   
			\hfill
			\setcounter{subfigure}{1} %reset counter
			\subfigure[\label{fig:compH01S_StdVeloX_b}]{\includegraphics[trim={18cm 3cm 18cm 0},clip, width=0.31\linewidth, height=0.17\paperheight,keepaspectratio=false ]{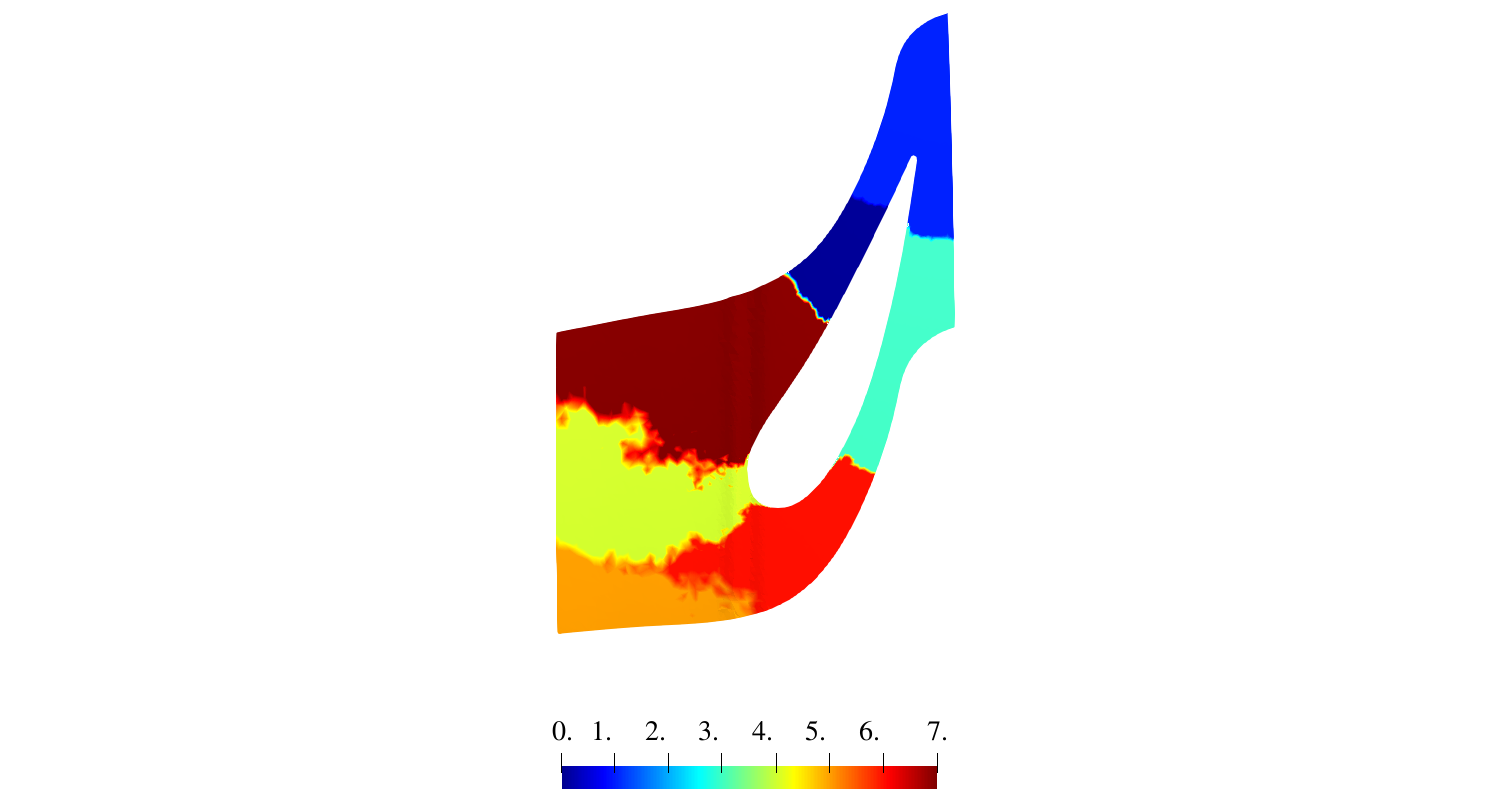}}
			\put(-112pt,85pt){\rotatebox{90}{50\% Span}}   
			\hfill
			\subfigure{\includegraphics[trim={18cm 3cm 18cm 0},clip, width=0.26\linewidth, height=0.17\paperheight,keepaspectratio=false ]{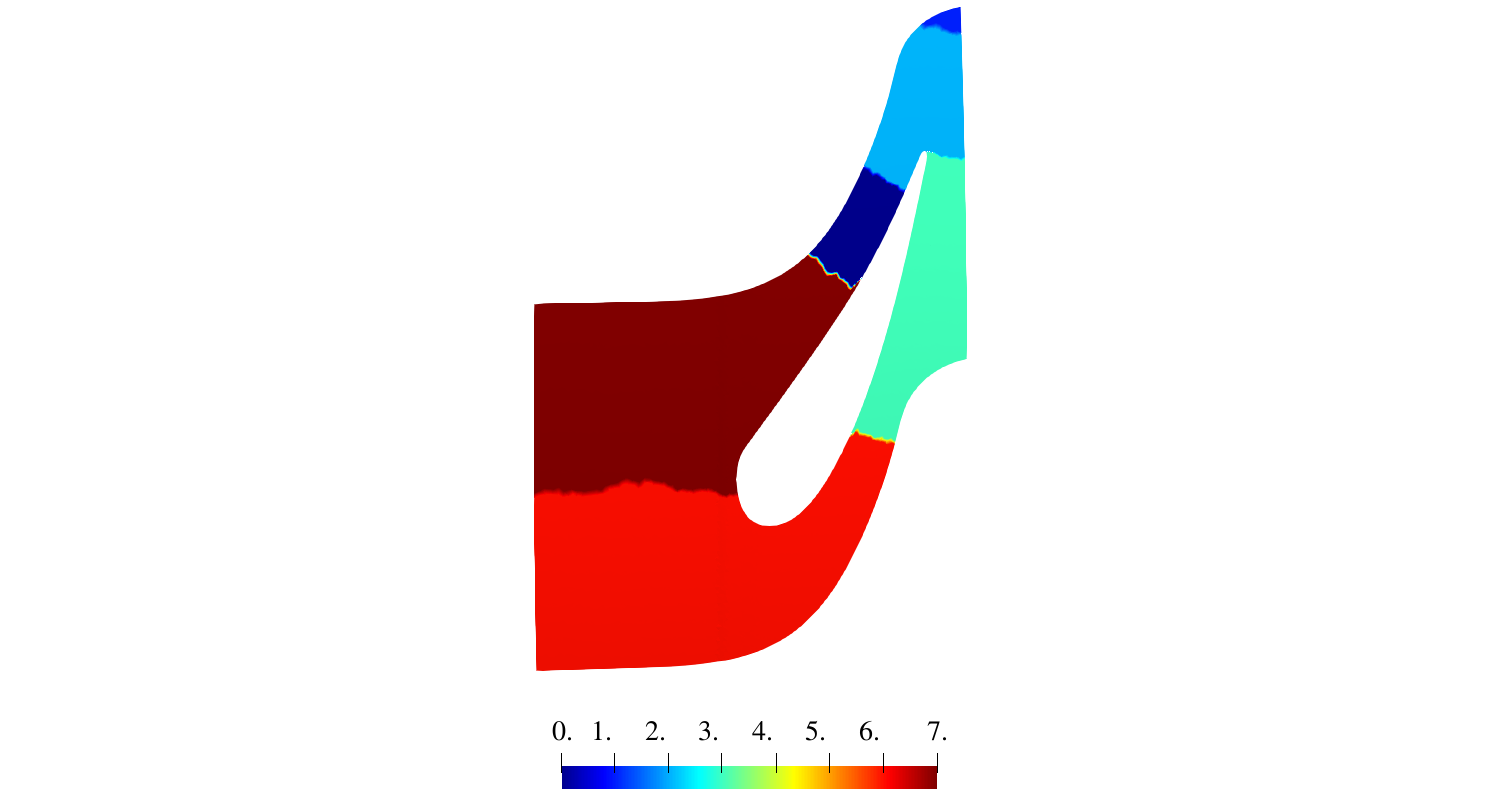}}  
			\put(-101pt,90pt){\rotatebox{90}{90\% Span}}   
			%
			% (trim for color legend)
			%
			\subfigure{\includegraphics[trim={0 15cm 9cm 0},clip, width=0.1\linewidth, ]{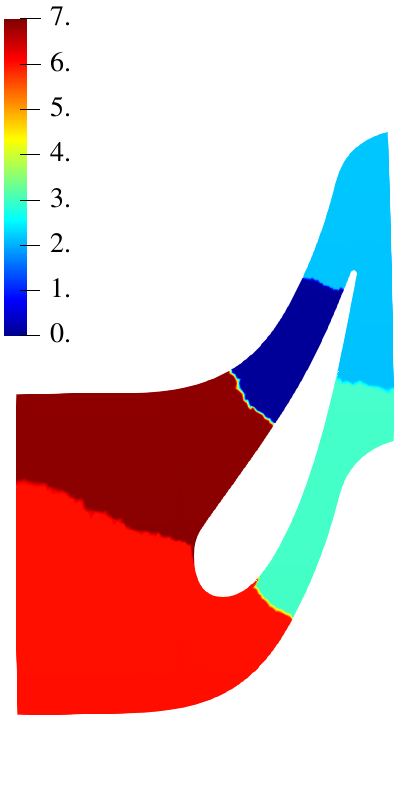}}
			\put(-0.03\linewidth,22){\rotatebox{90}{\small Partition Index}}      
		\end{subfigmatrix}
	\end{adjustwidth}
	\caption{(a) Side-by-side comparison of standard deviation of velocity $u$ over whole trajectory of sample at hand and (b) of the partition index for the complete flow domain, each for three spans of the nominal turbine sample.}
	\label{fig:compH01S_StdVeloX}
\end{figure}
In the following, a possible explanation why the partitioned flow field of the turbine stator was not learned completely with it's temporal variability is outlined.
Before a conjecture is drawn and in order to evaluate how significant the temporal variability especially in the main flow $u$ is, the temporal standard deviation of $u$ of the target flow field of the nominal stator sample is presented in Fig.~\ref{fig:compH01S_StdVeloX_a}. 
It becomes apparent that on all three spans pictured, the main variations occur in the area of the trailing edge. 
The rest of the domain does not vary at all over the complete trajectory.
For turbine simulations with stationary inflow conditions this is common for the turbine stator which is why this poses a challenge that needs to be addressed in future work.
In Fig.~\ref{fig:compH01S_StdVeloX_b} the partitioning of the stator domain is pictured. 
It is assumed that in gradient synchronization during training, the model replicas that do not see any temporal variations (partitions towards the inlet of the domain) dominate the weight updates so that the variability that happens in only three of the eight partitions is not learned.

\subsection{Performance Study}\label{subsec:perf}
Distributed training of \acp{dnn} and the outlined distribution strategy in particular require a deeper look at the runtime and communication intensive parts of the code. 
All results in this work were produced using nodes with a NVIDIA HGX A100 8-GPU platform with \SI{400}{\watt} GPU TDP. A single node was used if not stated otherwise. 
Nvidia NCCL~\cite{NvidiaNCCL} was chosen as communication backend in the distributed PyTorch setup~\cite{DDPDesignNote}.

The initial \textsc{MGN-Halo} implementation needs $~0.4s/\mathrm{step}$ to process a turbine stator sample.
Leveraging TorchScript~\cite{torchscript} and \ac{amp}~\cite{torchAMP} leads to a $35\%$ runtime reduction and a runtime of $~0.26s/\mathrm{step}$.
We observed diverging behavior when choosing high learning rate values which is why the usage of \ac{amp}~\cite{torchAMP} for model parameter studies might be considered carefully.

The performance measurements are conducted on the accelerated \textsc{MGN-Halo} model.
The CUDA streams are synchronized with the host to enable host site time measurement of the forward pass.
This introduces an overhead of $1.4\%$.
The PyTorch model is traced with Score-P \cite{Knuepfer2012} and the Score-P Python Bindings \cite{Gocht2020}.
Tracing introduces additional $9.2\%$ overhead resulting in a total measurement overhead of $10.7\%$.
Vampir \cite{vampir} is used afterwards to visualize and explore the trace.

The parallel execution of the \ac{dnn} introduces two types of parallelization overhead.
First, the potential load imbalance introduced by unbalanced computational work due to mesh partitions of different sizes per GPU.
Second, communication overhead which includes the cost of packing and unpacking the send and receive buffer respectively and sending the data to all neighbors.

The load imbalance is described with the fraction of device idle time $fr_\text{idle}$ which occurs when waiting for other processes to finish.
It can be calculated with the formula
\begin{align}
fr_\text{idle} = 1 - \frac{\sum_{i=1}^N{t_\text{active,i}} } {N \cdot t_\text{active,max}}
\end{align}
with ($N$), the number of GPUs, $t_\text{active,i}$, the individual compute time of each GPU and $t_\text{active,max}$, the longest compute time of all GPUs.

The maximum device idle time is $19\%$.
This correlates with the time needed for send mask and recv mask creation and packing and unpacking of the buffer.
Using TorchScript and \ac{amp} amplifies this problem by disproportionately accelerating the computational tasks in contrast to  the buffer packing and unpacking tasks.

\begin{figure}[b!]
	\centering
	\includegraphics[width=15cm]{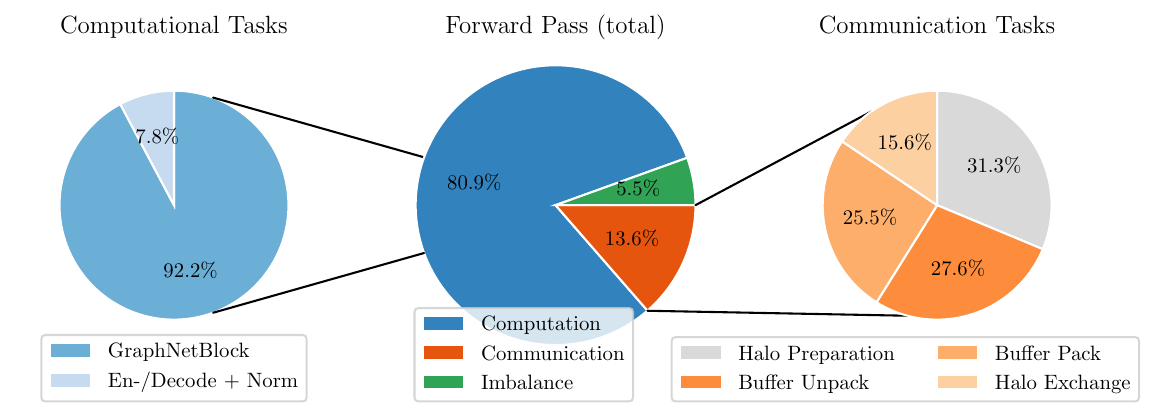}
	\caption{Runtime Distribution of the forward pass of \textsc{MGN-Halo-8-96}}
	\label{fig:MGN_runtime_distribution}
\end{figure}

Figure~\ref{fig:MGN_runtime_distribution} visualizes the temporal composition of 90 forward passes of \textsc{MGN-Halo-8-96}.
$80.9\%$ are spent in computational tasks, $13.6\%$ are spent in communication tasks and $5.5\%$ are spent idling due to load imbalance.
The communication overhead of parallelization is to be divided between the actual communication and its preparation. 
The latter includes the creation of sender and receiver mask, which happens once per timestep, and the buffer pack and unpack tasks, which happens every message passing step (compare with algorithm \ref{alg:distr_meshgn_training}). 
Altogether, this preparation needs $84.4\%$ of the time spend in communication tasks.
The halo exchange itself is implemented with a collective \textit{all\_to\_all} operation provided by PyTorch's Distributed Data Parallel class and accounts for only $15.6\%$ of the communication overhead.
Furthermore it becomes apparent from Fig.~\ref{fig:MGN_runtime_distribution} that $92.2\%$ of computational tasks within the forward pass are spent within the GraphNetBlock. 
The remaining $7.8\%$ are used for normalization, encoding and decoding tasks.

\begin{figure}[h!]
	\centering
	\includegraphics{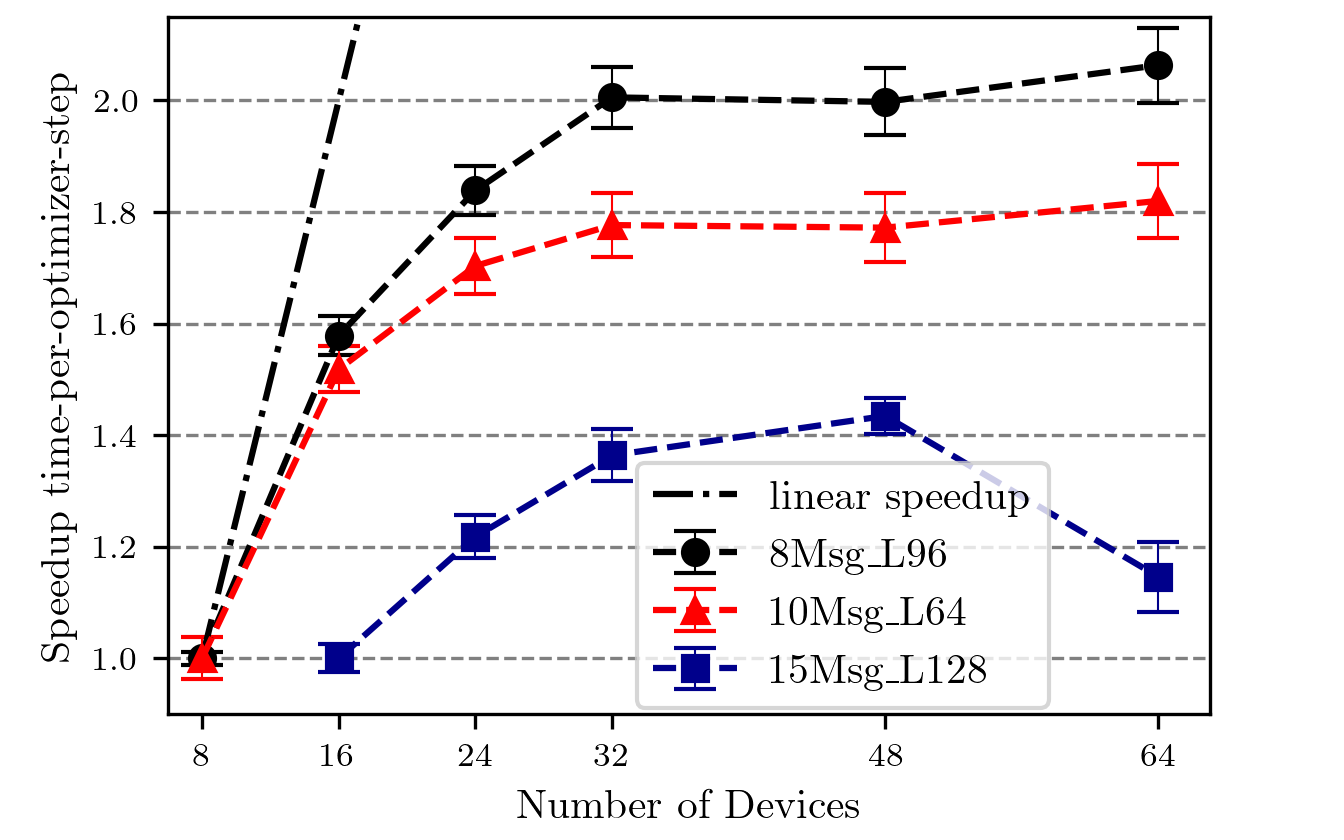}
	\caption{Strong scaling of three different models using proposed distributed training approach. Largest model (15Msg\_L128) does not run on single node (eight devices).}
	\label{fig:Scaling}
\end{figure}

Finally, a strong scaling analysis is performed to assess whether the proposed halo approach benefits from parallelization to more than the devices required in terms of memory.
Figure~\ref{fig:Scaling} presents the time-per-optimizer step normalized by the smallest possible configuration for each model.

This can be denoted as speedup which ideally scales linearly with the hardware available.
For the smaller models with $8$ and $10$ message-passing steps this ideal speedup line is shown as a dashed line.
However, using 16 \acp{gpu} which is a duplication of the hardware, to train a single domain (constant problem size) leads to a speedup of only $50\%-60\%$.
With two nodes or 16 \acp{gpu} one can train additionally the original model configuration with 15 message-passing steps and a latent size of 128.

For all model configurations the relative speedup to each duplication of hardware decreases with more \acp{gpu} used.
For the large model it even declines from $48$ to $64$ \acp{gpu}.
As a quintessence one can argue that the smallest number of \acp{gpu} that are able to train a certain domain size should be used as every larger device count decreases the runtime but lavishes precious resources.

This analysis outlines possible future improvements.
This includes the preparation of send and receive mask beforehand or asynchronously to reduce the overhead and the overall device idle time.
Furthermore, a more efficient method of buffer packing and unpacking may be employed as well as overlapping communication with computation. %using another algorithm

\section{Conclusions and Future Work}\label{ch:conclusion}
Recent implementations of \ac{ml} architectures to predict fluid flow achieve already outstanding results on academic examples. 
Building upon a state-of-the-art \ac{dnn} this work outlines and implements an approach for graph-based \ac{ml} models to predict fluid flow on industrially-relevant mesh sizes.
As halo exchanges between devices and adjacent message-passing steps is introduced, this approach is considered a logical extension of the single \ac{gpu} implementation of \ac{mgn}.
In order to validate the proposed approach a pre-study on a two-dimensional cylinderflow dataset from \cite{pfaff2021learning} is carried out and their results presented in section \ref{subsec:CYL}.
The proposed approach \textsc{MGN-Halo} performs slightly worse for single and multi-step rollouts than the single \ac{gpu} implementation.
Using gradient accumulation as a batching strategy for the single device implementation as well, the single step rollout results are comparable, which is why the pre-study was declared successful.
However, the multi step rollout results of \textsc{MGN-Halo} are still worse.

In the subsequent section \textsc{MGN-Halo} is applied to a dataset of a representative state-of-the-art turbine stator with $0.75\times10^6$ points in average.
A classic distributed learning strategy using \textsc{Horovod}~\cite{sergeev2018horovod} on partitioned samples without additional communication is used as a comparison and denoted \textsc{MGN-NoComm} in the following.
Both training approaches are compared after half of the number of optimizer steps used in~\cite{pfaff2021learning}. 
For single and multi-step rollouts as well as for 1-step rollouts starting from any timestep (\textit{Nextstep} prediction) \textsc{MGN-NoComm} is superior.
Visual inspections of 1-step rollouts of a validation sample confirmed the low prediction quality of \textsc{MGN-Halo} as areas of large error values as well as prediction artifacts are visible.
A possible explanation is that here, a classical halo exchange was performed, transferring only the node features.
But \ac{mgn} learns on node and edge features. An extension to exchange edge features additionally is therefore the mandatory next step to achieve sufficient results. 
%A more sophisticated and \ac{cfd} specific error measure is part of future improvements

Then, the capabilities of the superior \textsc{MGN-NoComm} model are examined and further prediction shortcomings were discovered.
The examined model configuration is not able to predict temporal variability that is present in the training. 
Instead, the model learned mean flow fields of the different geometries provided.
The relative error $e$ of the temporal deviation $\sigma_{\tau}$ is around $91\%$ for prediction on the training set. 
Prediction on the validation set are similarly stationary.
First evaluation of the amount of temporal deviation that was present in the dataset lead to the assumption that the model might suffer from vanishing gradients during gradient synchronization.
When the majority of the model replicas see quasi stationary flow, their gradients are close to zero and will dominate the gradient average that is computed.
Another possible explanation is that through the increased number of points in the domain the number of message passing steps is too low to exchange relevant flow information beyond the local neighborhood.
In \cite{pfaff2021learning} the correlation between lower message passing steps and accuracy deterioration is already presented.
Here, memory and resource constraints forced a lower message passing step number than in the original implementation which probably further lowered the accuracy of the proposed approach.
Recent advances in multiscale \ac{mgn} approaches \cite{Lino2021,Lino2022,Lino2022a,Fortunato2022} could resolve this issue as on coarse grids the number of message passing steps that is required for flow information to travel to distant parts of the domain is reduced.

Even though the proposed approach fell short of traditional distributed training, valuable experience with the training and inferencing of \acp{gnn} on large \ac{cfd} meshes was gained.
To the best of the authors knowledge, this is the first attempt of applying \ac{mgn} to train on a domain of about $10^6$ points.
Even though the proposed method turns out worse than the traditional training approach with \textsc{Horovod}, the experience of how to train on graph domains of this size is essential for future applications to even larger domains.
Furthermore, multiple mandatory improvements for the detected shortcomings of the models presented here were outlined and presumably lead to the proposed method being a useful extension to \ac{mgn} for the training on large \ac{cfd} domains after all. 
Future work will besides the evaluation of the improved halo exchange method include hybrid approaches using multiscale \ac{mgn} approaches, physics-based loss functions and transfer learning for sustainable usage of the expensively trained model.

Not limited to fluid flow, \ac{gnn} architectures can also be adapted to support other mesh-based numerical methods e.g. FEM.
This opens new possibilities for the integration of \ac{ml} models into design pipelines or into numerical solvers directly in order to speed-up convergence and decrease the required computational resources overall.

\newpage
\section*{Appendix}
\begin{figure}[h!]
	\newcommand\x{15} %	0.33
	\begin{adjustwidth}{1.8cm}{1.8cm}
		% trim={<left> <lower> <right> <upper>}
		\begin{subfigmatrix}{3}
			\subfigure{\includegraphics[trim={18cm 3cm 18cm 0},clip, width=0.33\linewidth, height=0.18\paperheight,keepaspectratio=false ]{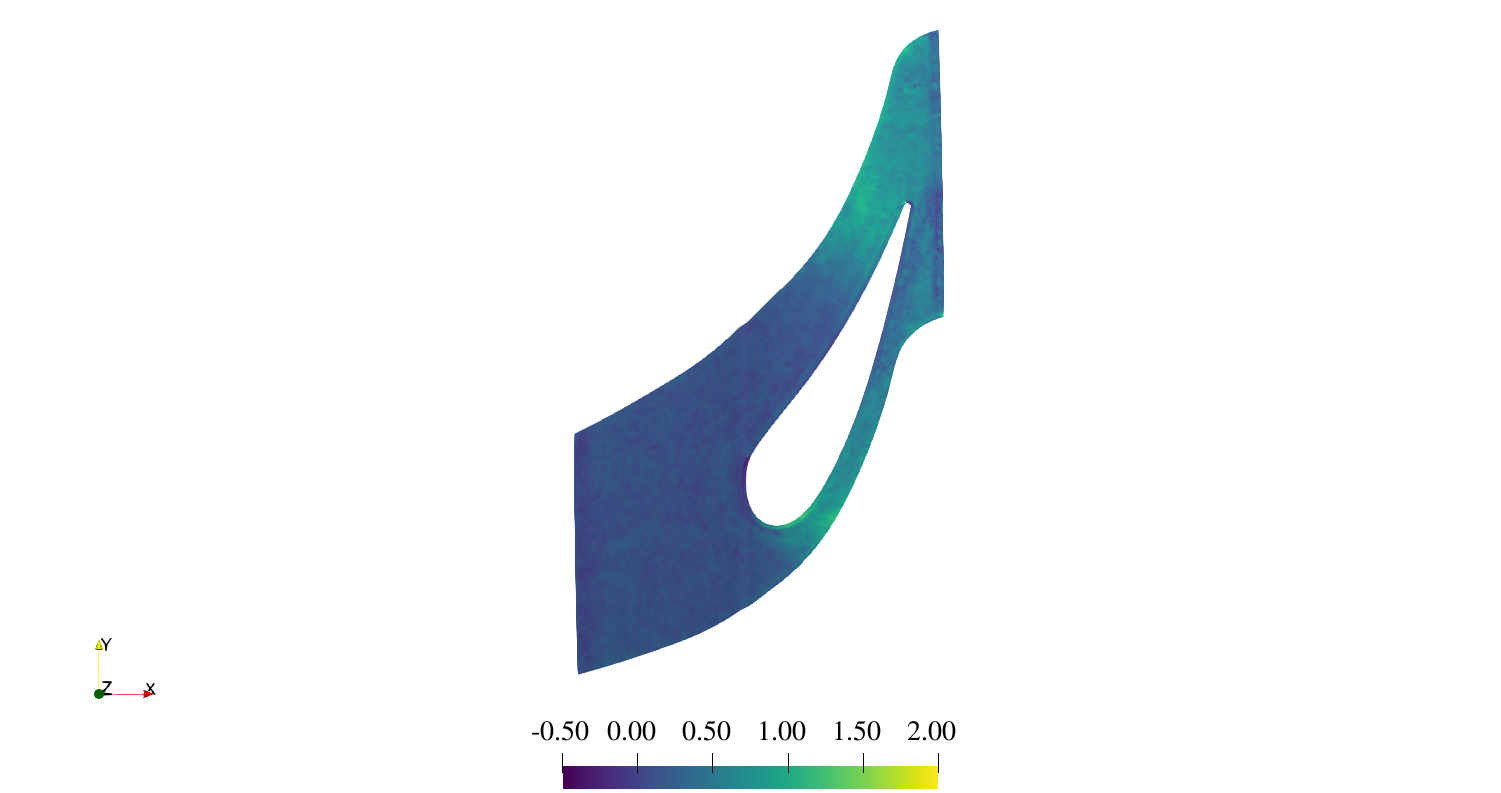}}
			\put(-105pt,75pt){\rotatebox{90}{10\% Span}}   
			\hfill
			\subfigure{\includegraphics[trim={18cm 3cm 18cm 0},clip, width=0.33\linewidth, height=0.18\paperheight,keepaspectratio=false ]{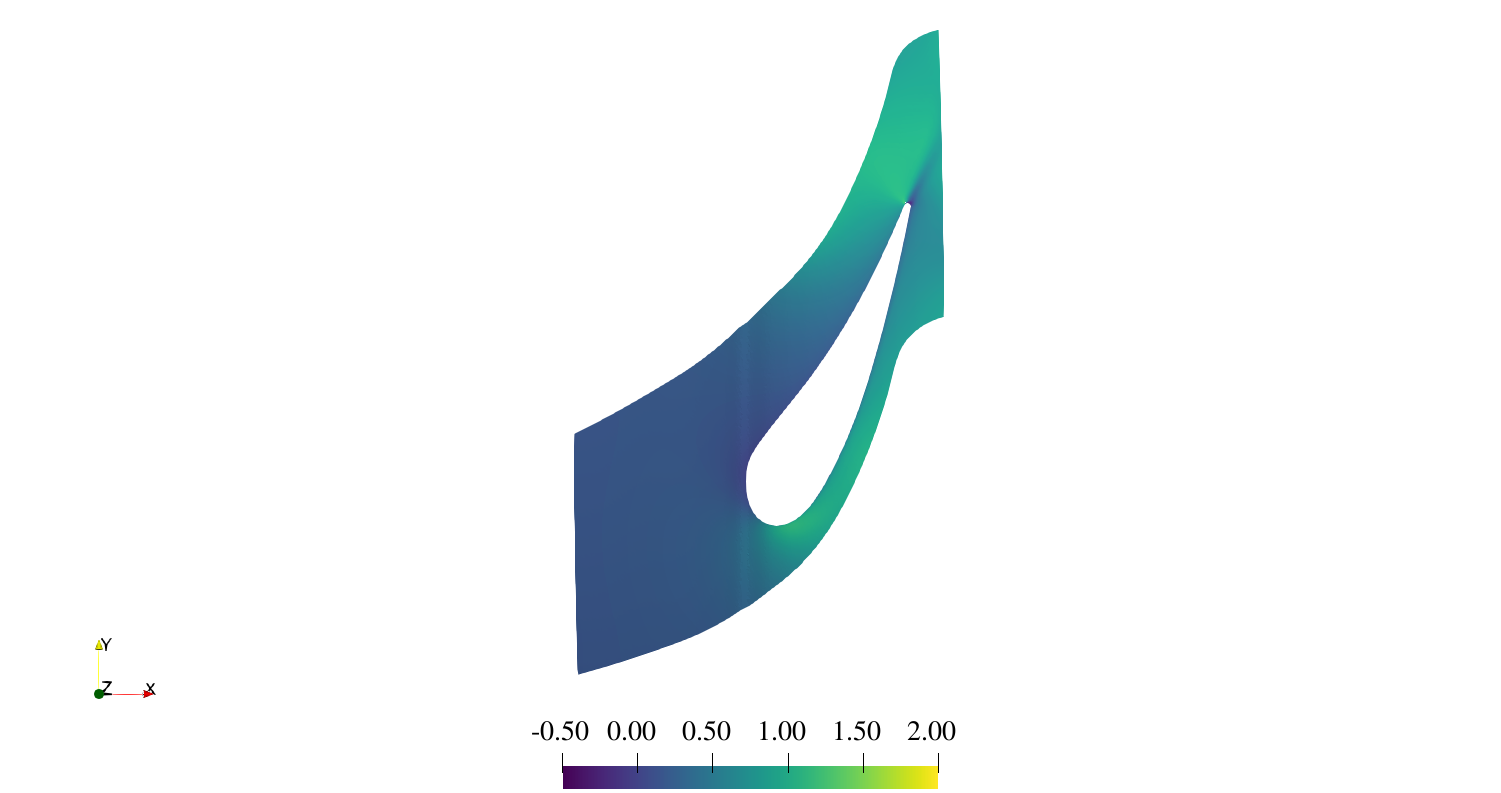}}
			\hfill
			\subfigure{\includegraphics[trim={18cm 3cm 18cm 0},clip, width=0.33\linewidth, height=0.18\paperheight,keepaspectratio=false ]{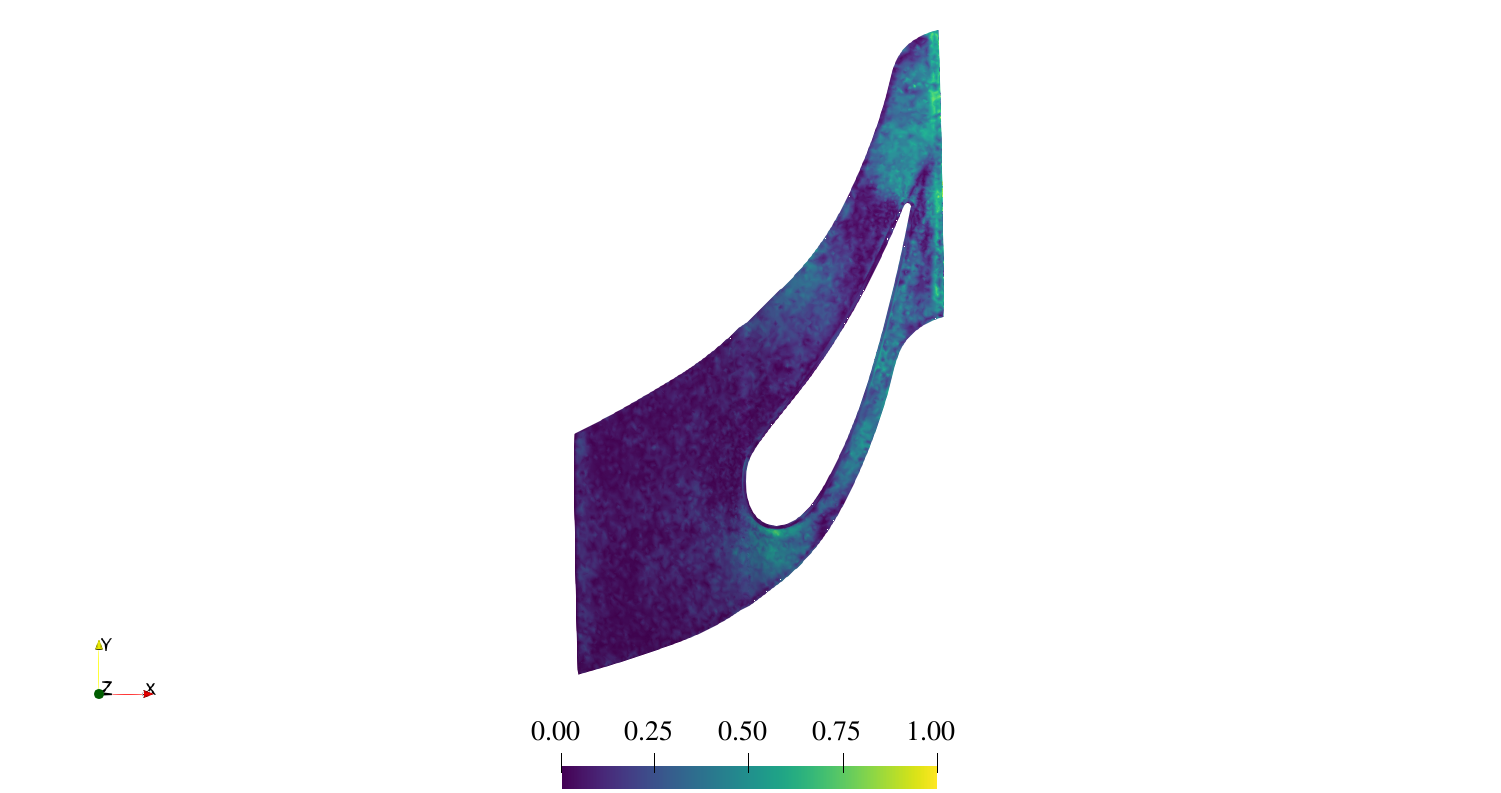}}  
			%
			%% New Line
			%
			\subfigure{\includegraphics[trim={18cm 3cm 18cm 0},clip, width=0.33\linewidth, height=0.17\paperheight,keepaspectratio=false ]{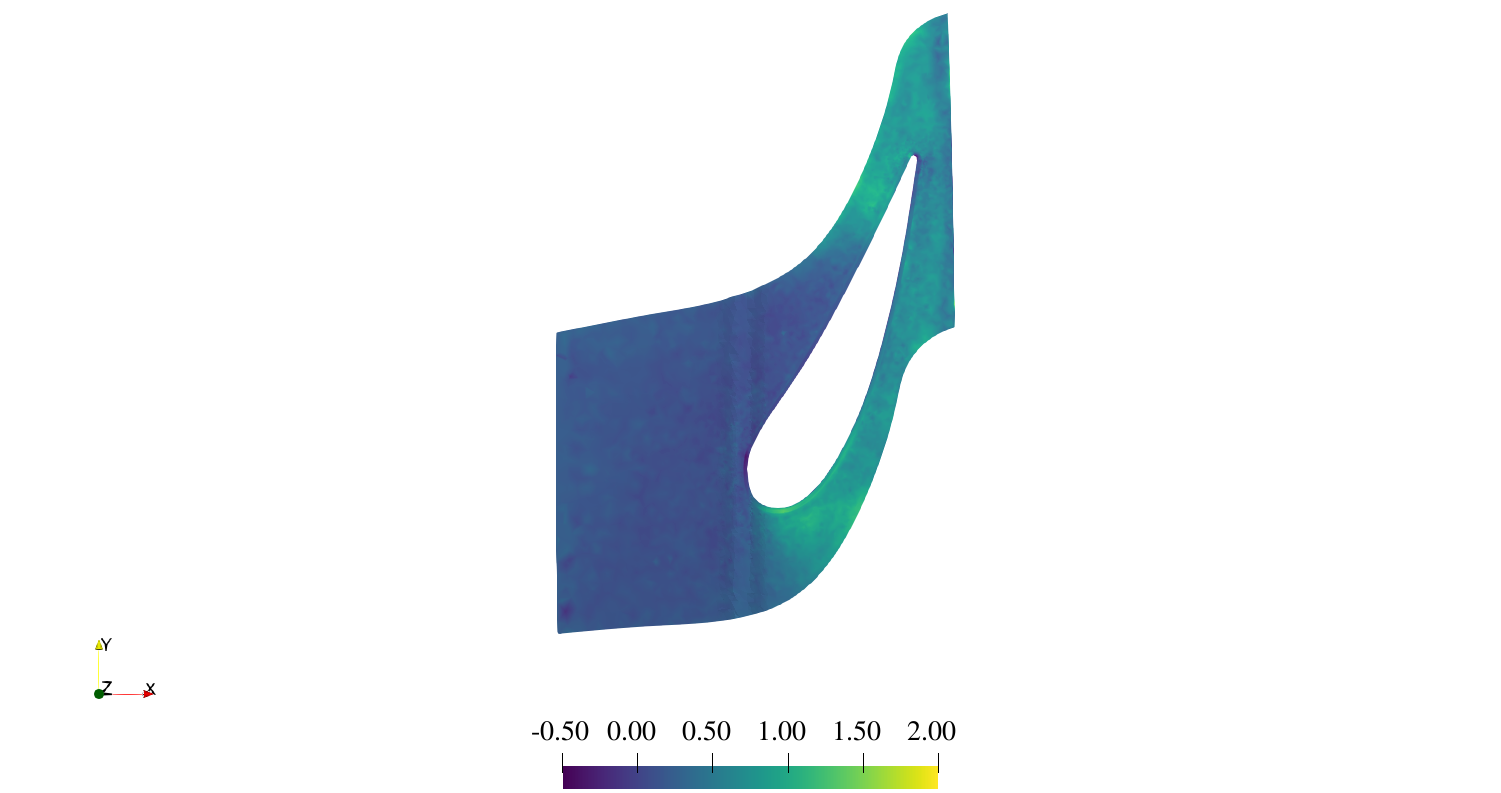}}
			\put(-109pt,85pt){\rotatebox{90}{50\% Span}}   
			\hfill
			\subfigure{\includegraphics[trim={18cm 3cm 18cm 0},clip, width=0.33\linewidth, height=0.17\paperheight,keepaspectratio=false ]{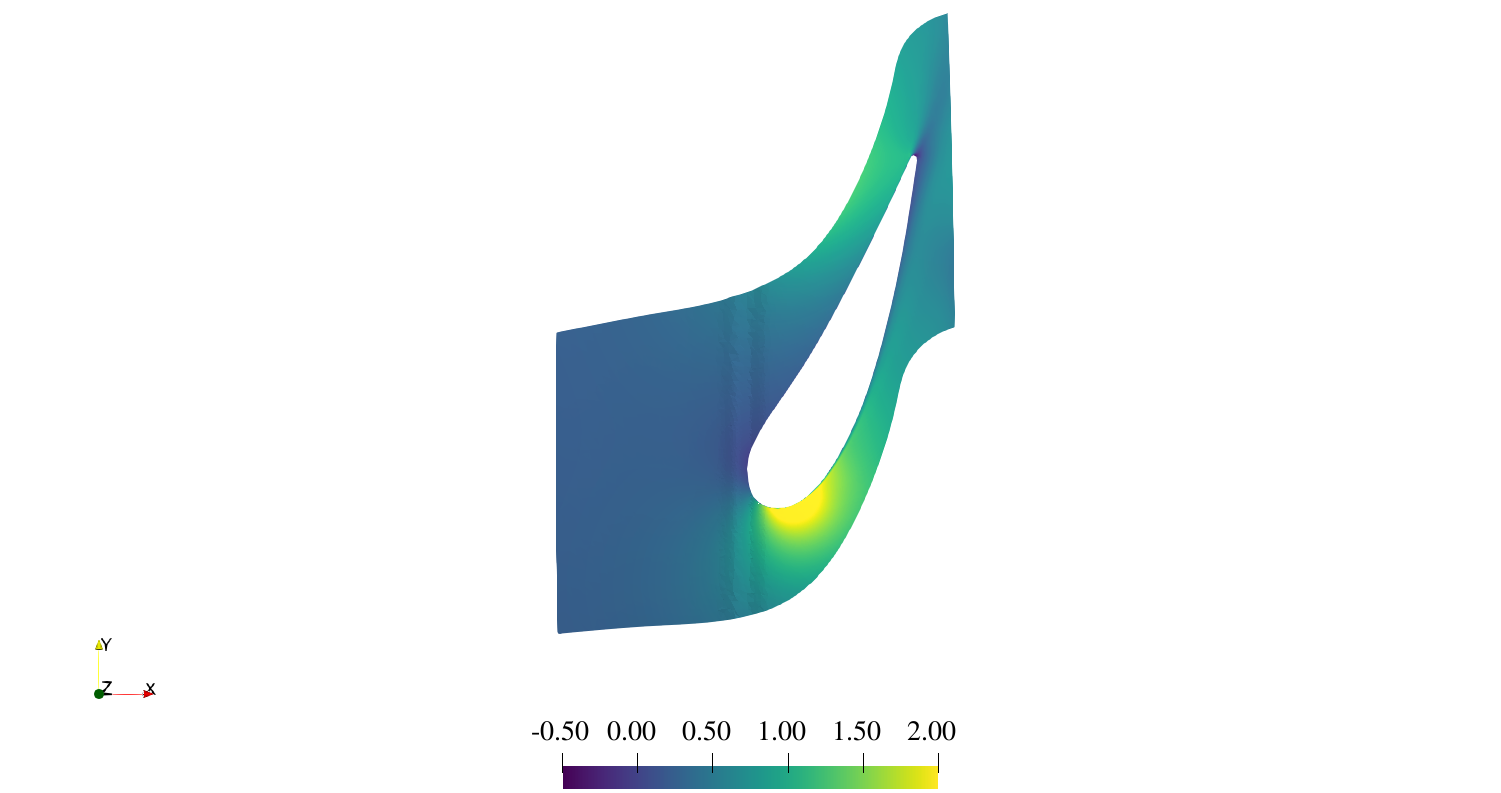}}
			\hfill
			\subfigure{\includegraphics[trim={18cm 3cm 18cm 0},clip, width=0.33\linewidth, height=0.17\paperheight,keepaspectratio=false ]{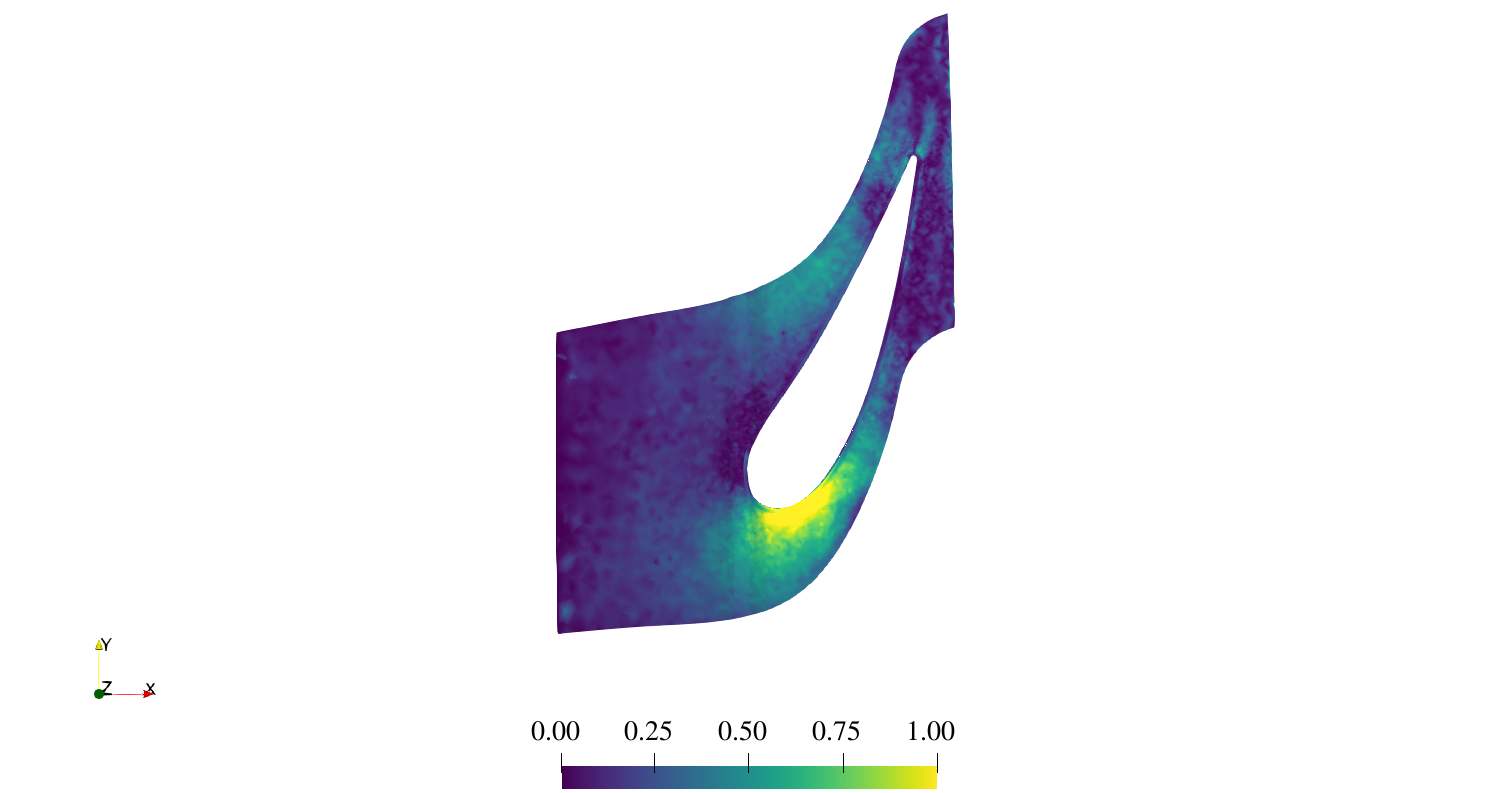}}  
			%  
			%% New Line 
			%
			\subfigure{\includegraphics[trim={18cm 3cm 18cm 0},clip, width=0.33\linewidth, height=0.19\paperheight,keepaspectratio=false ]{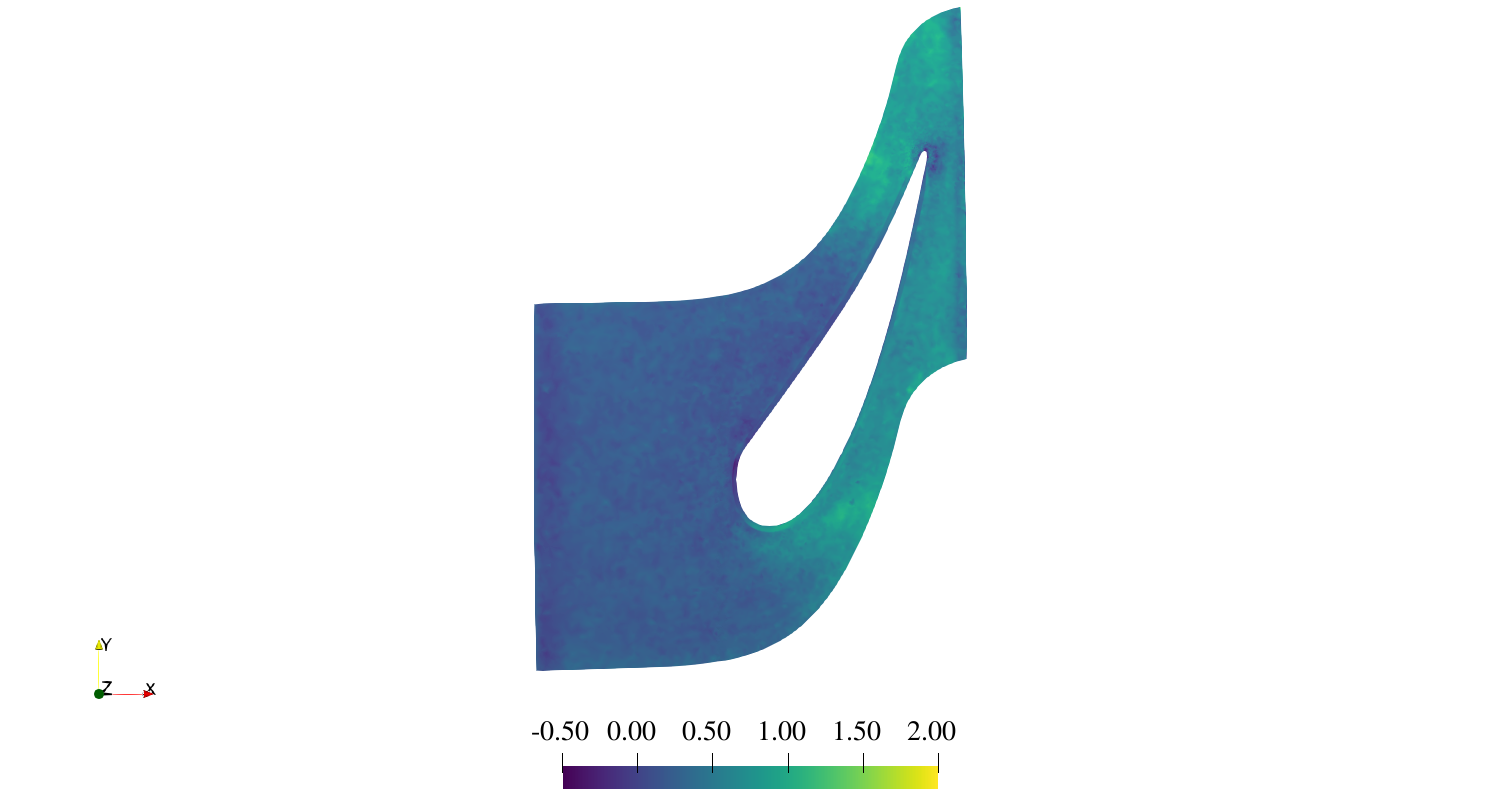}}
			\put(-114pt,100pt){\rotatebox{90}{90\% Span}}   
			\hfill
			\subfigure{\includegraphics[trim={18cm 3cm 18cm 0},clip, width=0.33\linewidth, height=0.19\paperheight,keepaspectratio=false ]{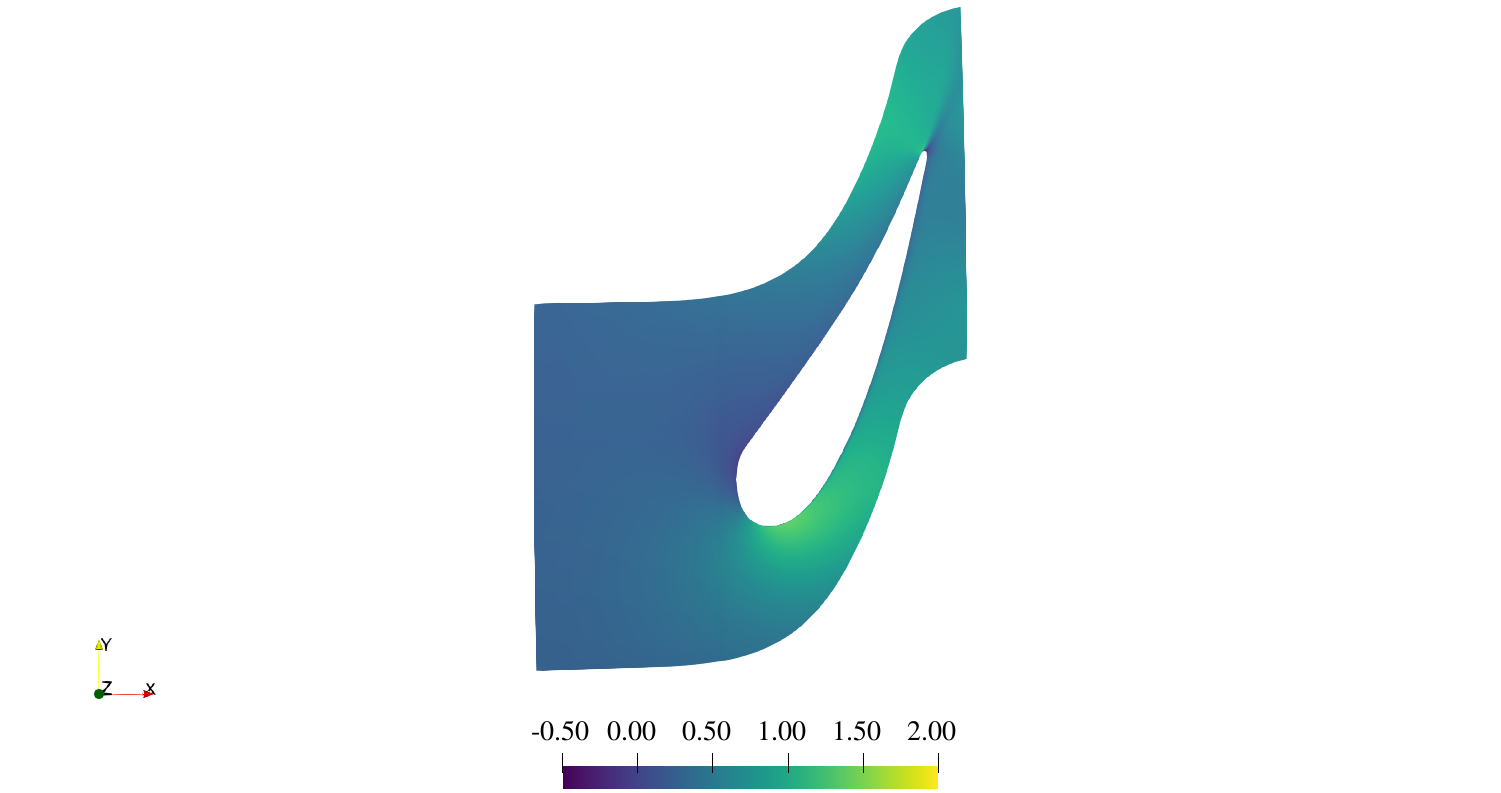}}
			\hfill
			\subfigure{\includegraphics[trim={18cm 3cm 18cm 0},clip, width=0.33\linewidth, height=0.19\paperheight,keepaspectratio=false ]{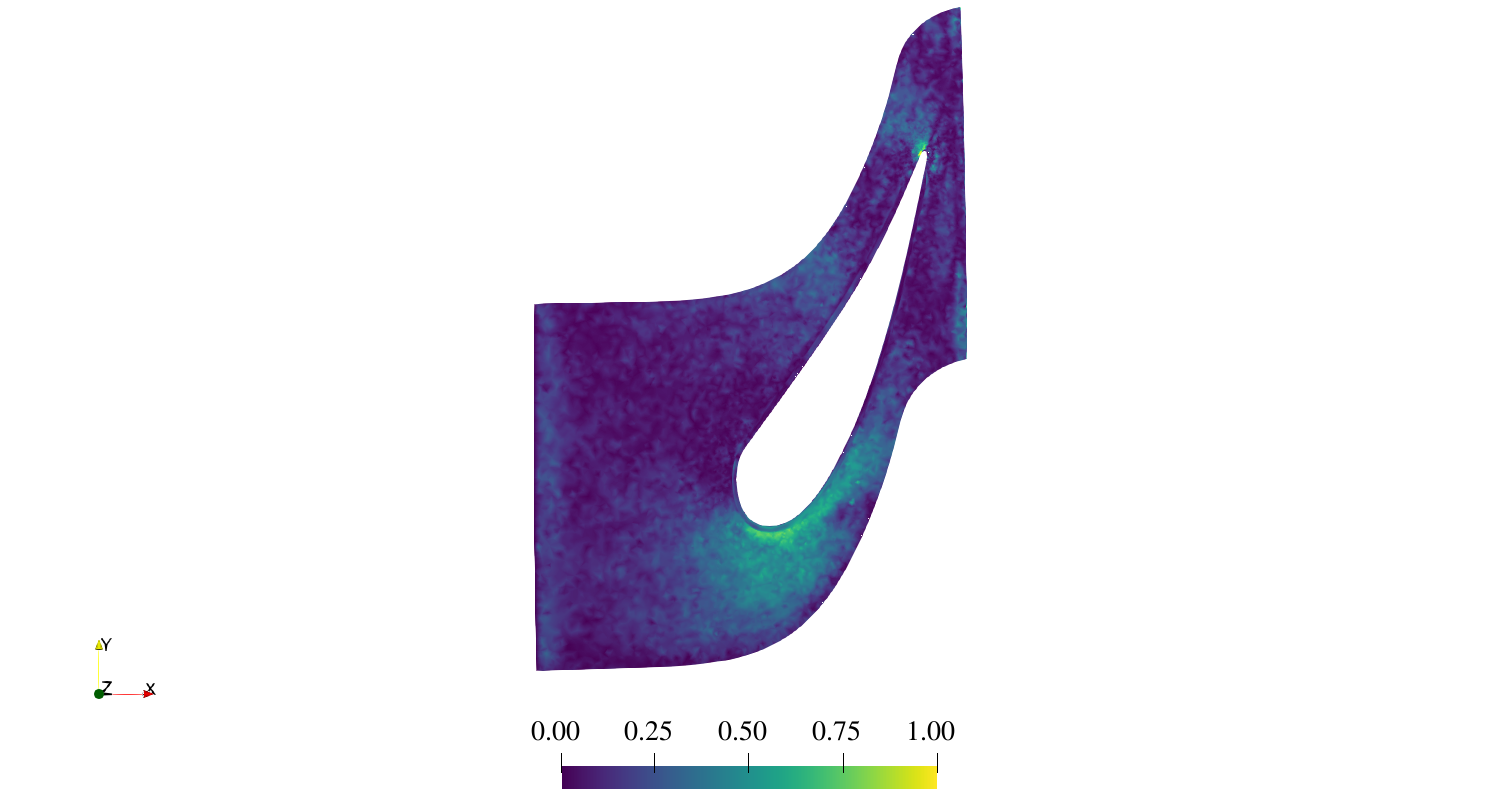}}  
			%
			% New Line (trim for color legend)
			%
			\subfigure{\includegraphics[trim={18cm 0 18cm 25cm},clip, height=\x\baselineskip ,keepaspectratio]{./figures/turbine/DDP8_10-64_90percSpan_VeloOutXTarget.png}}
			\put(-0.29\linewidth,22){Predicted Velocity $\tilde{u}$, \SI{}{\metre\per\second}} 
			\hfill
			\subfigure{\includegraphics[trim={18cm 0 18cm 25cm},clip, height=\x\baselineskip ,keepaspectratio]{./figures/turbine/DDP8_10-64_90percSpan_VeloOutXPred.png}}
			\put(-0.29\linewidth,22){Target Velocity $u$, \SI{}{\metre\per\second}}      
			\hfill
			\subfigure{\includegraphics[trim={18cm 0 18cm 25cm},clip, height=\x\baselineskip ,keepaspectratio]{./figures/turbine/DDP8_10-64_90percSpan_RelError_VeloX.png}}
			\put(-0.29\linewidth,22){Normalized Error $E(\tilde{u})$}     
		\end{subfigmatrix}
	\end{adjustwidth}
	\caption{Side-by-side comparison of predicted velocity $u$ (left), corresponding target velocity (mid) and normalized error (right) for a single-step rollout from \textsc{MGN-Halo-10-64} of the turbine testset sample with the lowest \ac{rmse}.}
	\label{fig:compH01S_Halo}
\end{figure}

\begin{algorithm}
	\caption{Training of distributed MeshGraphNets}
	\label{alg:distr_meshgn_training}
	\textbf{Input:} Model $\textit{f}_{\boldsymbol{\theta}} =$ $\{\textit{EncNode}_{\boldsymbol{\theta}}$,
	 $\textit{EncEdge}_{\boldsymbol{\theta}}$,
	 $\textit{GraphNetBlock}_{\boldsymbol{\theta}}$,
	 $\textit{DecNode}_{\boldsymbol{\theta}}\}$,
	 time series data on the nodes $\{\boldsymbol{X},\boldsymbol{Q}\} = \{\boldsymbol{x}^n,\Delta\boldsymbol{q}^{n}\}_{n=1}^{N}$,
	 edges and edge features describing the geometry $ \{\boldsymbol{\mathcal{E}}^n,\mathbf{e}^n\}_{n=1}^{N}$,
	 blocks and neighbors describing the mesh partitions $ \{\boldsymbol{b}^n,\boldsymbol{nb}^n\}_{n=1}^{N}$ and
 	 loss mask excluding boundary conditions  $\{\boldsymbol{\textit{mask}}^n_\textit{loss}\}_{n=1}^{N}$ of length $N$;
 	 Number of epochs: $M$;
 	 Gradient accumulation step: \textit{acc};
 	 Learning rate: $\eta$;
 	 Number of Processes:~\textit{CS};
     Own rank: \textit{pid}; \\

	\For{\text{epoch} $= 1$ \KwTo $M$}
	{
        $\nabla \boldsymbol{\theta} \gets 0$\;
		\For{$n = 1$ \KwTo $N$}
		{	
			$\Delta \boldsymbol{q}^n_{\textit{norm}},\boldsymbol{x}^n_{norm},\mathbf{e}^n_{norm} \gets \text{Normalize}(\Delta \boldsymbol{q}^n,\boldsymbol{x}^n,\mathbf{e}^n)$\;
			$\boldsymbol{x}^n_{enc} \gets \textit{EncNode}_{\boldsymbol{\theta}}(\boldsymbol{x}^n_{norm})$\;
			$\mathbf{e}^n_{enc} \gets \textit{EncEdge}_{\boldsymbol{\theta}}(\mathbf{e}^n_{norm})$\;
			$\textit{buff}_\textit{send},\textit{buff}_\textit{recv} \gets \textit{CS}\times[]$\;
			$\textit{mask}_\textit{send}, \textit{mask}_\textit{recv} \gets \textit{CS}\times[]$\;
			
			\For{$\textit{com} \gets 1$ \KwTo $\textit{CS}$}
			{
				\If{$\textit{com} \in \boldsymbol{nb}^n$}
				{
					$\textit{mask}_\textit{send}$[com], $\textit{mask}_\textit{recv}$[com] $\gets$ create\_masks($\boldsymbol{b}^n$, $\boldsymbol{\mathcal{E}}^n$, \textit{pid}, \textit{com}) \;
				}
			}
			\For{$k \gets 1$ \KwTo $K$}
			{
			    $\boldsymbol{x}^{n,k+1}_{enc}, \mathbf{e}^{k+1}_{enc} \gets \textit{GraphNetBlock}_{\boldsymbol{\theta}}(\boldsymbol{x}^{n,k}_{enc}, \mathbf{e}^{k}_{enc}, \boldsymbol{\mathcal{E}}^n)$\;
				\For{$\textit{com} \in \boldsymbol{nb}^n$}
				{
					$\textit{buff}_\textit{send}[\textit{com}] \gets \boldsymbol{x}^{n,k+1}_{enc}[\textit{mask}_\textit{send}[\textit{com}]]$\;
				}
				$\textit{buff}_\textit{recv} \gets \text{all\_to\_all}(\textit{buff}_\textit{send})$\;
				\For{$\textit{com} \in \boldsymbol{nb}^n$}
				{
					$\boldsymbol{x}^{n,k+1}_{enc}[\textit{mask}_\textit{recv}[\textit{com}]] \gets \textit{buff}_\textit{recv}[\textit{com}]$\;
				}
			}
			
			$\mathbf{p}^n \gets \textit{DecNode}_{\boldsymbol{\theta}}(\boldsymbol{x}^{n,k+1}_{enc})$\;
			$\mathcal{L}_{local} \gets SSE(\mathbf{p}^n,\Delta \boldsymbol{q}^n_{norm})$\;

			$\mathcal{L} \gets \textit{loss\_allreduce}(\mathcal{L}_{local},\textit{mask}^{n}_{\textit{loss}})$\;
			$\nabla \boldsymbol{\theta} \gets \nabla \boldsymbol{\theta} + \text{Backprop}(\mathcal{L})$\;
			\If{$Mod(n, \mathrm{\textit{acc}})=0$}
			{
				$\boldsymbol{\theta} \gets \boldsymbol{\theta} - \eta \nabla \boldsymbol{\theta}$\;
				$\nabla \boldsymbol{\theta} \gets 0$\;
			}
		}
	}
\textbf{Output:} Trained model $\textit{f}_{\boldsymbol{\theta}} =$ $\{\textit{EncNode}_{\boldsymbol{\theta}}$,
$\textit{EncEdge}_{\boldsymbol{\theta}}$,
$\textit{GraphNetBlock}_{\boldsymbol{\theta}}$,
$\textit{DecNode}_{\boldsymbol{\theta}}\}$
\end{algorithm}

\newpage
\section*{Acknowledgments}
The work presented in this paper was conducted within the framework of the DARWIN research project (20D1911C), funded by Rolls-Royce Deutschland Ltd \& Co KG and the Bundesministerium für Wirtschaft und Klimaschutz. 
Rolls-Royce Deutschland’s permission to publish this work is greatly acknowledged.
The authors gratefully acknowledge the GWK's support for funding this project by providing computing time through the Center for Information Services and HPC (ZIH) at TU Dresden.
\bibliography{literature}

\end{document}